%% file: AAStemplatev2_0_6.tex
\documentclass[letterpaper, preprint, paper,11pt]{AAS}	

\usepackage{bm}
\usepackage{amsmath}
\usepackage{amsfonts}
\usepackage{subcaption}
\usepackage[colorlinks=true, pdfstartview=FitV, linkcolor=black, citecolor= black]{hyperref}
\usepackage{overcite}
\usepackage{footnpag}			      	
\usepackage[export]{adjustbox}
\usepackage{multirow}

\usepackage[table]{xcolor}
\usepackage{colortbl}
\usepackage{booktabs} 
\newcommand{\ra}[1]{\renewcommand{\arraystretch}{#1}} 
\definecolor{yellow}{cmyk}{0, 0, 1.0, 0}

\DeclareMathOperator*{\argmin}{arg\,min}

\newcommand{\vvec}[1]{\bm{\mathrm{#1}}}

\usepackage{enumitem,amssymb}
\newlist{todolist}{itemize}{2}
\setlist[todolist]{label=$\square$}
\usepackage{pifont}

\PaperNumber{23-114}

\begin{document}

\title{Efficient Feature Description for Small Body Relative Navigation using Binary Convolutional Neural Networks}

\author{Travis Driver\thanks{PhD Student, Institute for Robotics and Intelligent Machines, Guggenheim School of Aerospace Engineering, Georgia Institute of Technology, Atlanta, GA 30332, USA.} 
\ and Panagiotis Tsiotras\thanks{David and Lewis Chair and Professor, Institute for Robotics and Intelligent Machines, Guggenheim School of Aerospace Engineering, Georgia Institute of Technology, Atlanta, GA 30332, USA.}
}

\maketitle{}

\begin{abstract}
Missions to small celestial bodies rely heavily on optical feature tracking for characterization of and relative navigation around the target body. 
While techniques for feature tracking based on deep learning are a promising alternative to current human-in-the-loop processes, designing deep architectures that can operate onboard spacecraft is challenging due to onboard computational and memory constraints. 
This paper introduces a novel deep local feature description architecture that leverages binary convolutional neural network layers to significantly reduce computational and memory requirements. 
We train and test our models on real images of small bodies from legacy and ongoing missions and demonstrate increased performance relative to traditional handcrafted methods. 
Moreover, we implement our models onboard a surrogate for the next-generation spacecraft processor and demonstrate feasible runtimes for online feature tracking.
\end{abstract}





\section{Introduction}
\input{Text/introduction}


\section{Related Work}
\input{Text/related_work}


\section{Proposed Approach}
\input{Text/approach}


\section{Results}
\input{Text/results}


\section{Conclusion}

In this paper, we presented a novel deep local feature description architecture, DidymosNet, that leverages BNN layers to significantly reduce the computational requirements of the model. 
The model was validated on \textit{real} remote imagery of small bodies and demonstrated significant performance increases with respect to handcrafted feature extraction methods. 
Moreover, we demonstrated feasible feature extraction runtimes for online feature tracking onboard a surrogate for the next-generation spacecraft processor~\cite{doyle2013hpsc,powell2018hpsc}, i.e., a ROCK Pi 4b.
We believe our network could viably operate onboard future robotic spacecraft, and is a promising alternative to the current human-in-the-loop approach based on DTMs. 
Future work will focus on combining DidymosNet with a learned keypoint detection scheme and integrating our approach into a full SLAM system, e.g., AstroSLAM~\cite{dor2022astroslam}.
Our code, data, and trained models will be made available to the public at \url{https://github.com/travisdriver/didymosnet}.


\section*{Acknowledgements}

This work was supported by a NASA Space Technology Graduate Research Opportunity. 
The authors would like to thank Kenneth Getzandanner and Andrew Liounis from NASA Goddard Space Flight Center for several helpful discussions and comments.


\bibliographystyle{AAS_publication}   
\bibliography{references}   


\appendix
\input{Text/appendix}

\end{document}

%% file: Text/introduction.tex
Feature detection and tracking is an integral component of current small celestial body shape reconstruction and relative navigation approaches. 
The current state-of-the-practice for small body feature detection and tracking leverages high-fidelity digital terrain maps (DTMs) of salient surface regions as local feature representations, which require extensive human involvement and mission operations planning for accurate construction~\cite{barnouin2020,palmer2022practical}. 
Autonomous keypoint detection and feature description methods based on deep convolutional neural networks (CNNs) are a promising alternative to the current human-in-the-loop approach~\cite{driver2022astrovision}, but the inherent computational overhead of deep architectures makes their implementation onboard spacecraft challenging. 

Binary convolutional neural networks (BNNs) are a promising technology for enabling deep learning onboard future robotic spacecraft. 
BNNs represent weights and activations with 1-bit and use bitwise operations to perform cross-correlation, leading to significant computational, memory, and energy savings. 
Indeed, binary convolutions have been shown to be almost 60$\times$ faster while using less than 1\% of the energy compared to 32-bit floating-point convolutions on FPGAs and ASICs~\cite{mishra2018iclr}. 
While BNNs have traditionally led to significant performance degradation, recent work has demonstrated impressive performance on image classification by minimizing quantization error~\cite{rastegari2016eccv,bulat2019bmvc} and maximizing information entropy~\cite{qin2020cvpr} within the BNN layers. 
Therefore, we propose a novel and efficient feature description architecture with BNN layers for deep feature tracking onboard spacecraft. 
We refer to our network as \textit{DidymosNet}, named after the \textit{binary} asteroid 65803 Didymos currently being investigated by the Asteroid Impact \& Deflection Assessment (AIDA) mission~\cite{cheng2018aida}.

The primary contributions of this work are as follows: 
(i) we develop DidymosNet, a novel feature description architecture that leverages BNN layers to significantly increase the computational efficiency of the model, 
(ii) we train and test our model on real images of small bodies from legacy and ongoing missions to small bodies and demonstrate increased performance relative to traditional handcrafted methods, 
(iii) we deploy our model onboard flight-relevant hardware and achieve significant runtime reductions relative to other deep architectures. 
Our code, data, and trained models will be made available to the public at \url{https://github.com/travisdriver/didymosnet}.

%% file: Text/related_work.tex
In this section, we first detail the current state-of-the-practice for small body optical navigation, and then we review some common feature detection and description methods. 


\subsection{Small Body Optical Navigation}

Robust tracking of salient image features is a critical component of current small body relative navigation methods, as the apparent displacement of tracked features between images can be leveraged to estimate the relative pose of the spacecraft as it moves around the body. 
In the context of optical feature tracking, saliency typically refers to the ability to detect and precisely localize the feature under multiple viewing conditions (i.e., \textit{repeatability}) and to the distinctiveness of the feature to ensure accurate matching between images (i.e., \textit{reliability})~\cite{dusmanu2019d2,revaud2019r2d2}. 
The current state-of-the-practice for small body feature tracking leverages high-fidelity DTMs, local topography and albedo maps, of salient surface regions as local feature representations, typically constructed on the ground using a process known as stereophotoclinometry (SPC)~\cite{gaskell2008}. 
DTM construction requires extensive human-in-the-loop verification and mission operations planning to achieve optimal results~\cite{barnouin2020,palmer2022practical}. 
Moreover, criteria for selecting salient features typically undergo multiple iterations through testing and development of the DTMs~\cite{norman2022autonomous,mario2022ground}.
Next, each DTM is combined with \textit{a priori} estimates of the spacecraft's pose and Sun pointing vector, along with a photometric model, to yield a photorealistic rendering of the DTM with respect to the input image. 
Finally, tracking is performed by comparing the rendering against the input image near the expected feature location using normalized cross-correlation, where a match is declared if a significant correlation peak is detected~\cite{lorenz2017,norman2022autonomous}. 
The relative pose of the spacecraft when the image was taken can be computed using the registered matches and the \textit{a priori} DTM position estimates. 
Therefore, this DTM-based method relies on the fidelity of the \textit{a priori} data products and can only be utilized after the target body has been adequately observed and the surface has been reconstructed at the required resolutions~\cite{bhaskaran2011}. 
This process has been used to produce an onboard navigation solution for execution of safety-critical maneuvers, e.g., during the OSIRIS-REx Touch-And-Go (TAG) sample collection event, where it was referred to as Natural Feature Tracking (NFT)~\cite{lorenz2017,norman2022autonomous}. 
While this approach has achieved much success, its reliance on extensive human involvement for extended durations limits mission capabilities and increases operational costs~\cite{quadrelli2015,nesnas2021,getzandanner2022scitech}. 

In this work, we instead rely on \textit{autonomous} keypoint detection and feature description in order to reduce the reliance on high-fidelity, \textit{a priori} data products for spacecraft navigation. 
\textit{Keypoints} localize salient regions in the image, which are typically extracted from a saliency map, where saliency can be predefined (e.g., corners) and localized using image filtering methods or learned from data (see the following section). 
Feature description is the task of forming a latent representation of the local image data at detected keypoints, where the latent representation commonly takes the form of a $d$-dimensional vector referred to as the \textit{descriptor}. 
Feature description seeks to assign a descriptor to each keypoint such that descriptors of corresponding keypoints are closer together than those of other non-corresponding keypoints with respect to some metric. 
Common metrics include the Euclidean distance, or the Hamming distance for binary descriptors~\cite{rublee2011iccv}.
Finally, feature tracking is conducted through detection of keypoints and matching of their corresponding descriptors between images. 
We refer the reader to Reference \citenum{driver2022astrovision} for more details on autonomous keypoint detection and feature description for small body relative navigation. 


\subsection{Feature Detection and Description}

Many computer vision algorithms rely on local image features. 
The seminal work of David Lowe's Scale Invariant Feature Transform (SIFT)~\cite{lowe2004ijcv} laid the foundation for the field, where he outlined a rigorous framework for identifying and describing image features.
SIFT follows a \textit{detect-then-describe} paradigm, whereby a series of predetermined (or \textit{handcrafted}) filters are applied to the image for keypoint localization, followed by pooling and normalization of image gradients to form the descriptor. 
SIFT aims to extract features that are invariant to changes in scale, illumination, and rotation. 
Keypoints are extracted from local extrema of the saliency map derived by convolving the difference of Gaussians (DoG) kernel with the input image, as the DoG function provides a close approximation to the scale-normalized Laplacian of Gaussian function which has been shown to be scale invariant~\cite{lindeberg1994scale}. 
This detection scheme generally results in keypoints centered around large gradients in the image (e.g., edges, corners).
Descriptors are then computed by pooling gradients in a local window of each keypoint into histograms according to their orientation, where a canonical orientation is assigned to each keypoint according to the dominant gradient orientation to provide robustness to rotation.
The oriented histograms are then concatenated and normalized to form the descriptor vector. 
%
%
Oriented FAST and Rotated BRIEF (ORB)~\cite{rublee2011iccv} has become a popular alternative to SIFT, especially for real-time simultaneous localization and mapping (SLAM) applications~\cite{mur2015tr}, and has also been applied to asteroid relative navigation~\cite{dor2022astroslam}. 
ORB is based on Features from Accelerated Segment Test (FAST) detectors~\cite{rosten2006eccv} and Binary Robust Independent Elementary Features (BRIEF) descriptors~\cite{calonder2010eccv} and outputs binary descriptor vectors, enabling more efficient matching. 

More recently, feature detection and description methods that leverage deep \textit{convolutional neural networks} (CNNs) have achieved state-of-the-art performance and have been shown to significantly outperform handcrafted methods on spaceborne imagery, especially in scenarios involving significant illumination and perspective change~\cite{driver2022astrovision}.
The first data-driven methods focused on individual components of the full image processing pipeline, including keypoint detection~\cite{verdie2015cvpr}, orientation estimation~\cite{Yi_2016CVPR}, and feature description~\cite{SimoSerra_2015ICCV}. 
Yi et al.~\cite{Yi_2016ECCV} developed the first complete learning-based pipeline, Learned Invariant Feature Transform (LIFT). 
LIFT uses a patch-based Siamese training architecture and implements each component of the traditional feature detector and descriptor scheme sequentially using CNNs. 
The approach relies on an incremental training procedure to pretrain each subnetwork component individually, with a final training phase that optimizes over the entire network end-to-end. 
SuperPoint~\cite{Detone_2018CVPRW} developed a network composed of separate interest point and descriptor decoders that operate on a spatially reduced representation of the input image from a shared encoder network. 
Simulated data of simple geometric shapes is used to pretrain the interest point detector, which is then combined with a random homographic warping procedure to train the network end-to-end in a \textit{self-supervised} fashion. 
Operating on a spatially reduced feature map (i.e., $1/8$ the spatial resolution of the input image) from the encoder allows SuperPoint to operate in real-time using a Titan X GPU. 
However, as we will demonstrate, the runtime performance of SuperPoint suffers on more computationally constrained hardware. 

Towards joint detection and description, the seminal work of D2-Net~\cite{dusmanu2019d2} proposed a \textit{detect-and-describe} approach that trains a single deep CNN to detect and describe salient image features. 
\textit{Reliability} (or \textit{distinctiveness}) of descriptors is enforced through a triplet margin ranking loss term which is weighted according to soft detection scores to jointly enforce \textit{repeatability} of detections. 
R2D2~\cite{revaud2019r2d2} leverages the detect-\textit{and}-describe paradigm to perform simultaneous feature detection and description, but repeatability and reliability are enforced in separate terms in the loss function. 
ASLFeat~\cite{luo2020cvpr} builds upon the success of D2-Net and proposes a multi-level detection scheme to generate detection scores that enable more accurate keypoint localization, and leverages deformable convolutional networks (DCNs)~\cite{zhu2019deformable} to model local geometric variations in the image and learn more transformation invariant features. 
While the recent success of data-driven feature detection and description is an attractive alternative to the current feature tracking approach used for missions to small bodies, implementing this technology onboard spacecraft is challenging due to the highly constrained computational resources. 
Therefore, we propose to leverage binary network quantization to greatly increase the throughput of our deep feature description model and provide a reliable architecture that could feasibly operate onboard future robotic spacecraft. 

The concurrent work of Kanakis and Maurer et al.~\cite{kanakis2022zippypoint} is the most similar to our work, which leverages mixed precision quantization within a detect-\textit{and}-describe architecture based on KP2D~\cite{tang2020kp2d}. 
However, the authors of Reference \citenum{kanakis2022zippypoint} found that binary convolutions lead to significant performance degradation and instead leverage 8-bit integer layers and 1-bit binary layers with high-precision residual operations, as well as full-precision layers in the decoder. 
Conversely, we rely exclusively on binary operations in our quantized layers and utilize a detect-\textit{then}-describe scheme in order to reduce unecessary computations by only performing description on regions centered around detected keypoints. 
Moreover, we evaluate our model runtimes on flight-relevant hardware. 


\subsection{Efficient Machine Learning for Space Applications}

Many previous works have demonstrated the utility of machine learning to improve the state-of-the-art for image processing tasks onboard future spacecraft, such as semantic segmentation~\cite{claudet2022benchmark,tomita2022jsr,drivertomita2023dmhd}, noncooperative spacecraft rendezvous~\cite{park2022spnv2}, and feature tracking~\cite{fuchs2015,driver2022astrovision}. 
However, fewer works have validated their proposed solutions onboard computationally constrained devices. 
Claudet et al.~\cite{claudet2022benchmark} investigated semantic segmentation approaches for safe landing site selection and demonstrate runtimes of up to 10 frames per second (FPS) on a Raspberry Pi 4B, albeit on downsampled inputs (i.e., 128$\times$128 pixels). 
Park and D'Amico~\cite{park2022spnv2} developed a deep CNN architecture for pose estimation of a noncooperative spacecraft from monocular imagery and report inference times of 0.63 seconds per frame on a high-end desktop CPU, i.e., an Intel® Core™ i9-9900K CPU @ 3.60 GHz. 
In terms of feature tracking, Fuchs et al.~\cite{fuchs2015} train a random forest classifier on patches extracted from 119 images of the comets Hartley 2 and Tempel 1 and estimate runtimes of 20 seconds per frame on a RAD750. 
In this work, we evaluate the efficiency of our models on a single board computer, i.e., a ROCKPi 4B, and demonstrate feasible runtimes for online feature tracking on real, full-resolution (1024$\times$1024 pixels) imagery of small bodies.

%% file: Text/approach.tex
We propose a novel local feature description architecture that leverages BNN layers to efficiently and reliably describe extracted image patches. 
Our model is trained on real imagery acquired during legacy and ongoing missions to small bodies. 


\subsection{Binary Convolutional Neural Networks}

\begin{figure}
    \centering
    \includegraphics[width=\linewidth]{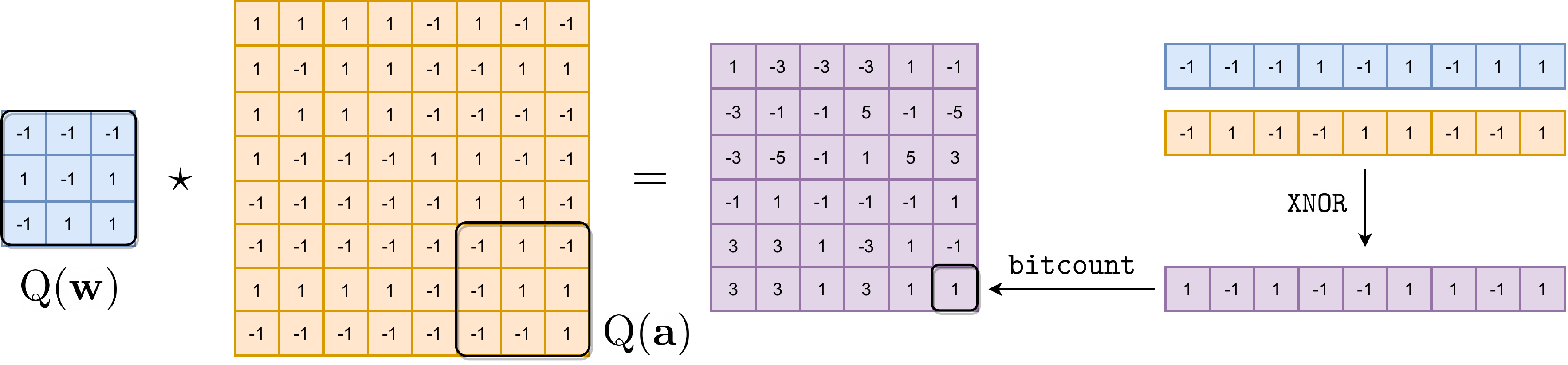}
    \caption{\textbf{Convolution with binary weights and activations.}}
    \label{fig:binary-conv}
\end{figure}

The main operation in CNNs can be formulated as
\begin{equation}
    z = \vvec{w}^\top \vvec{a},
\end{equation}
where $\vvec{w}, \vvec{a} \in \mathbb{R}^{s\cdot s\cdot c_{\text{in}}}$ are the (vectorized) weight tensor with kernel size $s$ and input activation tensor from the previous layer with $c_{\text{in}}$ channels, respectively. 
BNNs represent the weights and activations as 1-bit binary values for more efficient computation and storage. 
In general, binary network quantization can be formulated as 
\begin{equation} \label{eq:binary-quant}
    z = \gamma \mathrm{Q}(\vvec{w})^\top \mathrm{Q}(\vvec{a}) =  \gamma (\vvec{b}_{\mathrm{w}} \odot \vvec{b}_{\mathrm{a}}),
\end{equation}
where $\vvec{b}_{\mathrm{w}}, \vvec{b}_{\mathrm{a}} \in \{-1, +1\}^d$ are the quantized 1-bit weights and activations, respectively, $\gamma \in \mathbb{R}$ is a scaling factor, and $\odot$ denotes the inner product with bitwise operations \texttt{XNOR} and \texttt{bitcount} (see Figure \ref{fig:binary-conv}). 
The quantization function $\mathrm{Q}$ is typically taken to be the $\texttt{sign}$ function:
\begin{equation}
    \mathrm{Q}(\vvec{x}) =
    \begin{cases}
    +1, & \text{if } \vvec{x} \geq 0, \\
    -1, & \text{if } \vvec{x} < 0,\\
    \end{cases},
\end{equation}
which has been shown to be optimal with respect to the quantization error~\cite{rastegari2016eccv}. 
The scaling factor $\gamma$ can be set to an approximation of the optimal solution~\cite{rastegari2016eccv,qin2020cvpr} or learned via back-propagation~\cite{bulat2019bmvc}, where we chose the latter. 
However, instead of learning the factors explicitly in each convolutional layer~\cite{bulat2019bmvc}, we found that relying on the learned affine parameters in the normalization layers offered virtually identical performance and avoided redundant computations (see the Appendix). 

Moreover, following the work of Qin et al.~\cite{qin2020cvpr}, we implement a stochastic quantization method to reduce information loss and quantization error from the forward $\texttt{sign}$ operator and leverage an incremental function approximation for the backward gradient. 
Specifically, the stochastic binarization method models each element $b$ in the quantized weights $\vvec{b}_{\mathrm{w}}$ as a Bernoulli random variable with parameter $p$:
\begin{equation}
    f(b;p) =
    \begin{cases}
    p, & \text{if } b = +1, \\
    1-p, & \text{if } b = -1. \\
    \end{cases}.
\end{equation}
Therefore, the information entropy of the quantized elements is maximized with $p = 0.5$, which is approximately enforced via weight standardization~\cite{qin2020cvpr}:
\begin{equation}
    \Hat{\vvec{w}} = \frac{\vvec{w} - \mu}{\sigma},
\end{equation}
where $\mu$ and $\sigma$ are the mean and standard deviation of the filter weights $\vvec{w}$. 
This results in activations $z$ of the form
\begin{equation}
    z = \gamma\left(\texttt{sign}(\Hat{\vvec{w}}) \odot \texttt{sign}(\vvec{a})\right).
\end{equation}
%
%
%
Next, the discontinuity due to the \texttt{sign} operation in the forward step is approximated during backpropagation by~\cite{qin2020cvpr}
\begin{equation} \label{eq:tanh}
    g_{t,k}(x) = k\tanh(tx),
\end{equation}
%
where $k$ and $t$ are control variables that are incrementally adjusted during the training process:
\begin{equation}
    t := T_{\min} 10^{\frac{i}{n}\log(T_{\max}/T_{\min})},\quad k := \max\left(1/t, 1\right),
\end{equation}
where $i$ is the current epoch, $n$ is the total number of epochs, and $T_{\min}$ and $T_{\max}$ are taken to be 0.1 and 10, respectively.


\subsection{Network Architecture}

\begin{figure}[tb!]
\centering
\begin{subfigure}[t]{.44\linewidth}
  \centering
  \includegraphics[width=\linewidth]{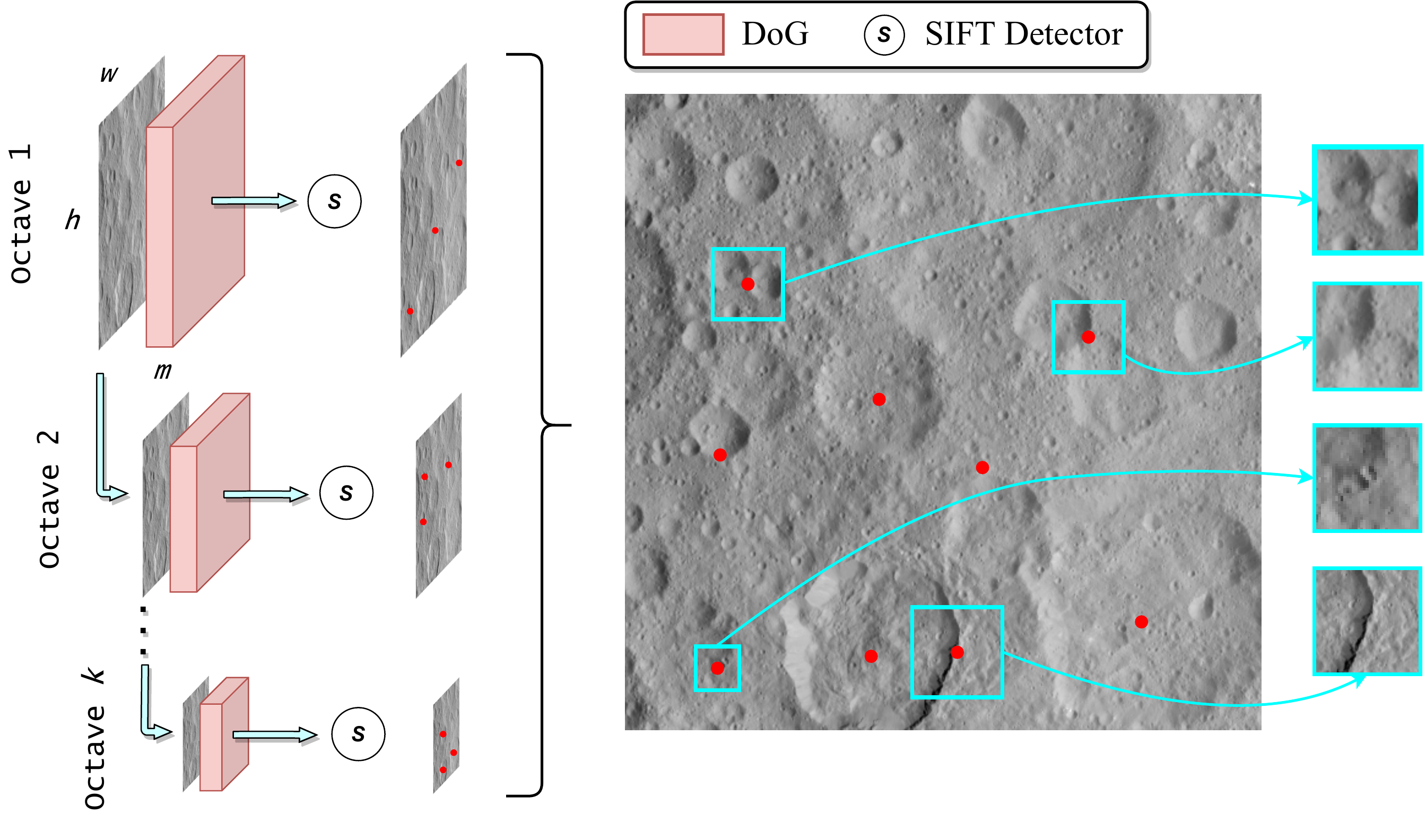}
  \caption{Keypoint detection}
  \label{fig:patch-arch-detect}
\end{subfigure}%
\hfill
\begin{subfigure}[t]{.52\linewidth}
  \centering
  \includegraphics[width=\linewidth]{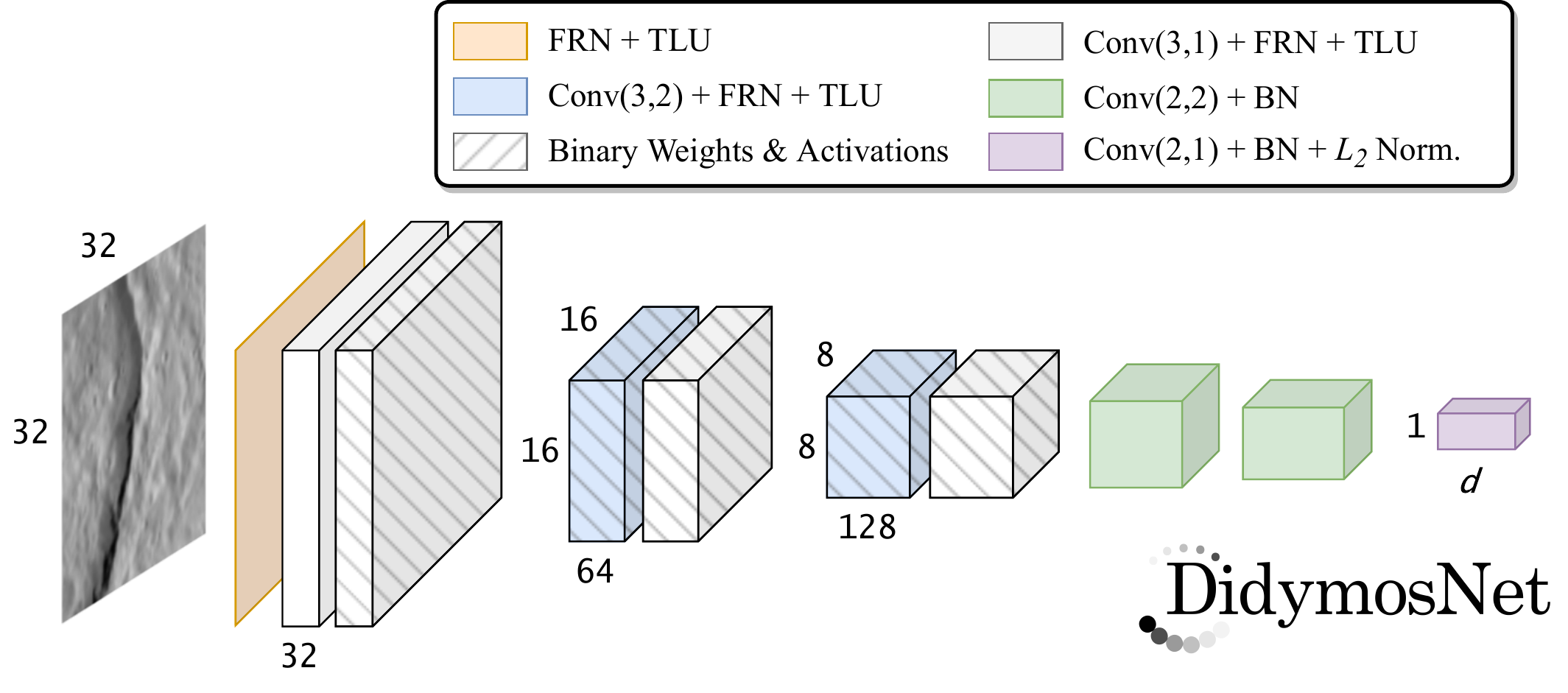}
  \caption{Feature description}
  \label{fig:patch-arch-describe}
\end{subfigure}%
\caption{\textbf{Feature extraction architecture.} Conv($a$, $b$) denotes convolution with a kernel size of $a$ and stride $b$.}
\label{fig:patch-arch}
\end{figure}

Keypoint detection and feature description methods operate by computing \textit{keypoints} which localize salient regions in the image and \textit{descriptors} which form a latent representation of the local image data at detected keypoints, where the descriptor typically takes the form of a $d$-dimensional vector. 
Feature tracking is conducted by matching features between frames with respect to some distance metric between their corresponding descriptors. 
We leverage the detect-\textit{then}-describe paradigm whereby local image patches are first extracted using a handcrafted keypoint detection scheme (Figure \ref{fig:patch-arch-detect}) and then described by passing the patches through the description network (Figure \ref{fig:patch-arch-describe}). 
This is in contrast to the detect-\textit{and}-describe paradigm whereby the network takes in the full image and outputs dense saliency and descriptor maps that cover the entire image~\cite{driver2022astrovision}. 
The detect-\textit{then}-describe paradigm is leveraged for two main reasons: firstly, description is only conducted on regions centered around detected keypoints, thus reducing unnecessary computations, and, secondly, \textit{multiscale} detection can be conducted without performing the expensive description step, as multiscale detection is usually conducted by feeding a ``pyramid" of successively downsampled images through the network one-at-a-time~\cite{dusmanu2019d2,revaud2019r2d2,luo2020cvpr}. 

We leverage the detection scheme of SIFT~\cite{lowe2004ijcv} for patch extraction. 
SIFT keypoints are extracted from local extrema of the saliency map derived by convolving the  difference of Gaussians (DoG) kernel with the input image, as the DoG function provides a close approximation to the scale-normalized Laplacian of Gaussian function which has been shown to be scale invariant~\cite{lindeberg1994scale}. 
Patches centered around each keypoint are extracted and resized to $32\times 32$, and are then passed through the description network to compute a descriptor for each patch. 
We use HyNet~\cite{tian2020nips} as our base description architecture, which consists of seven convolutional layers and outputs 128-dimensional descriptors. 
However, we replace the final convolutional layer, which features an $8\times 8$ kernel, with three successive layers with $2\times 2$ kernels, as we found that it offers comparable performance while significantly reducing the number of computations and parameters (see Appendix). 
Each convolutional layer except the last one before the final Batch Normalisation (BN) layer is followed by Filter Response Normalization (FRN) and a Thresholded Linear Unit (TLU)~\cite{singh2020cvpr} activation layer. 
We replace all full-precision CNN layers except the first and last with BNN layers.  
The full architecture is illustrated in Figure \ref{fig:patch-arch}. 

Moreover, the hybrid loss proposed by Tian et al.~\cite{tian2020nips} is used to learn discriminative local feature descriptors. 
Consider correspondences $\mathcal{M} := \{(j, \tau(j))\ |\ \tau: \mathcal{J} \leftrightarrow \mathcal{J}'\}$ between descriptors $\{\mathbf{d}_j\}_{j\in \mathcal{J}}$ and $\{\mathbf{d}_{j'}'\}_{j'\in \mathcal{J}'}$. 
First, a regularization term, denoted by $\mathcal{R}_{L_2}$, is used to account for varying descriptor magnitudes before the final $L_2$ normalization step:
\begin{equation}
    \mathcal{R}_{L_2} = \frac{1}{|\mathcal{M}|} \sum_{(k, k') \in \mathcal{M}} \left(\|\vvec{x}_k\| - \|\vvec{x}_{k'}'\|\right)^2,
\end{equation}
where $\vvec{x}_k$ and $\vvec{x}_{k'}'$ are the corresponding descriptors \textit{before} the final normalization step, and $\|\cdot\| = \|\cdot\|_2$.
Next, a triplet margin ranking loss that considers both the inner product and the $L_2$ distance between descriptors is used, which has been shown to allow for more balanced gradient updates between positive and negative pairs~\cite{tian2020nips}:
\begin{align} \label{eq:trip-loss}
    \mathcal{L}_{\text{triplet}} &= \frac{1}{|\mathcal{M}|} \sum_{(k, k') \in \mathcal{M}} \max(0, m + s_H(\vvec{d}_k, \vvec{d}_{k'}') - s_H(\vvec{d}_k^-, \vvec{d}_{k'}'^-)), \\
    s_H(\vvec{d}, \vvec{d}') &= \frac{1}{Z}(\alpha (1 - \vvec{d}^\top \vvec{d}') + \|\vvec{d} - \vvec{d}'\|), 
\end{align}
where $\vvec{d}_k^-, \vvec{d}_{k'}'^-$ is the hardest negative sample for the pair $\vvec{d}_k, \vvec{d}_{k'}'$, i.e., 
\begin{equation}
    \vvec{d}_k^-, \vvec{d}_{k'}'^- =
    \begin{cases}
    \vvec{d}_k^\vee, \vvec{d}_{k'}' & \text{if } \|\vvec{d}_k^\vee - \vvec{d}_{k'}'\| < \|\vvec{d}_k - \vvec{d}_{k'}'^\vee\| \\
    \vvec{d}_k, \vvec{d}_{k'}'^\vee & \text{otherwise} \\
    \end{cases},
\end{equation}
\begin{equation}
    \vvec{d}_k^\vee := \argmin_{\vvec{d}_l \in \mathcal{B} \setminus \vvec{d}_{k'}'}\|\vvec{d}_l - \vvec{d}_{k'}'\|,\ \vvec{d}_{k'}'^\vee := \argmin_{\vvec{d}_{l} \in \mathcal{B} \setminus \vvec{d}_{k}}\|\vvec{d}_k - \vvec{d}_{l}\|,
\end{equation}
$\mathcal{B} = \{\mathbf{d}_j\}\cup\{\mathbf{d}_{j'}'\}$, $m$ is the margin between the positive and negative samples, $\alpha \in \mathbb{R}^+$ adjusts the ratio between $s$ and $d$, and $Z$ ensures the gradient has a maximum magnitude of 1.
The hyperparameters $m$ and $\alpha$ are empirically set to 1.2 and 2, respectively, as in Reference \citenum{tian2020nips}. 
Moreover, we also incorporate the second-order similarity (SOS) regularization term, denoted by $\mathcal{R}_{\text{SOS}}$, proposed by Tian et al~\cite{tian2019sosnet} as follows
\begin{equation} \label{eq:sos}
    \mathcal{R}_{\text{SOS}} = \frac{1}{|\mathcal{M}|} \sum_{(k, k') \in \mathcal{M}} d^{(2)}(\vvec{d}_k, \vvec{d}_{k'}'),
\end{equation}
\begin{equation} \label{eq:d2}
    d^{(2)}(\vvec{d}_k, \vvec{d}_{k'}') = \left(\sum_{(l,l')\in \mathcal{Z}_{k, k,'}} (\|\vvec{d}_k - \vvec{d}_l\| - \|\vvec{d}_{k'}' - \vvec{d}_{l'}'\|)^2\right)^{1/2},
\end{equation}
where $\mathcal{Z}_{k,k'} = \{(l, l') \in \mathcal{M} \setminus (k, k')\ |\ \vvec{d}_l\in k\text{NN}(\vvec{d}_k) \vee \vvec{d}_{l'}'\in k\text{NN}(\vvec{d}_{k'}')\}$ and $k\text{NN}(\vvec{d})$ denotes the $k=8$ nearest neighbors (with respect to the Euclidean distance) of the descriptor $\vvec{d}$ over all the descriptors in the batch (excluding itself). 
Finally, the overall loss function is defined as 
\begin{equation}
    \mathcal{L} = \mathcal{L}_{\text{triplet}} + \mathcal{R}_{\text{SOS}} + \gamma \mathcal{R}_{L_2},
\end{equation}
where $\gamma$ is a regularization parameter which we set to $\gamma = 0.1$.

Furthermore, we also modified the network to optionally output binary descriptors. 
As before, let $\vvec{x}\in \mathbb{R}^d$ represent the descriptor before the final $L_2$ normalization step. 
Then, we compute binary descriptors via $\vvec{b} = \texttt{sign}(\vvec{x})\in \{+1,-1\}^d$, which is again approximated during back-propogation by Equation \eqref{eq:tanh}. 
Next, the loss is reformulated by letting $\vvec{d} = \vvec{b}/\|\vvec{b}\|$, which replaces the Euclidean distance in Equations \eqref{eq:trip-loss}--\eqref{eq:d2} with the \textit{normalized Hamming distance}:
\begin{equation}
    \|\vvec{d} - \vvec{d}'\| = \left\|\frac{\vvec{b}}{\|\vvec{b}\|} - \frac{\vvec{b}'}{\|\vvec{b}'\|}\right\|_2 = \frac{2 \sqrt{\text{Hamming}(\vvec{b}, \vvec{b}')}}{\sqrt{d}}.
\end{equation}
Note that $\mathcal{R}_{L_2} = 0$ since $\|\vvec{b}\| = \textrm{const.} = \sqrt{d}$. 
We refer to this network variant as DidymosNet$^\pm$.


\subsection{Dataset Generation}

\begin{figure}[htb!]
\centering
\begin{subfigure}[t]{0.48\linewidth}
  \centering
  \includegraphics[width=\linewidth]{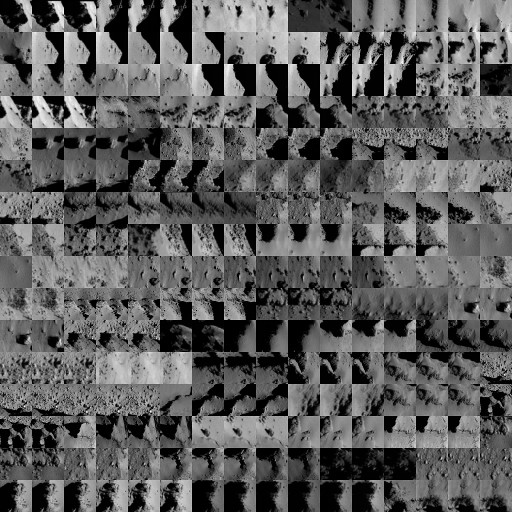}
  \caption*{\texttt{Rosetta @ 67P}}
\end{subfigure}%
\hfill
\begin{subfigure}[t]{0.48\linewidth}
  \centering
  \includegraphics[width=\linewidth]{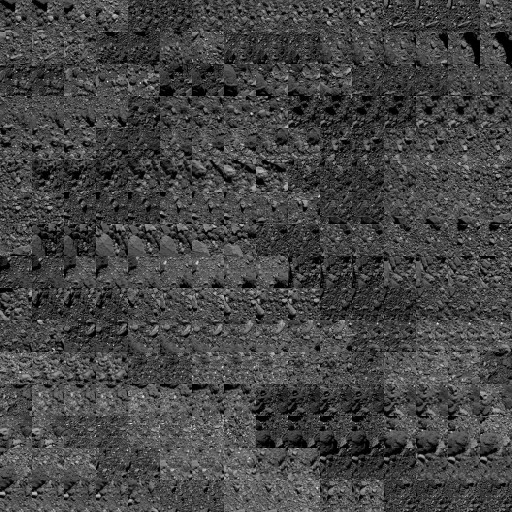}
  \caption*{\texttt{OSIRIS-REx @ 101955 Bennu}}
\end{subfigure}\\
\smallskip
\begin{subfigure}[t]{0.48\linewidth}
  \centering
  \includegraphics[width=\linewidth]{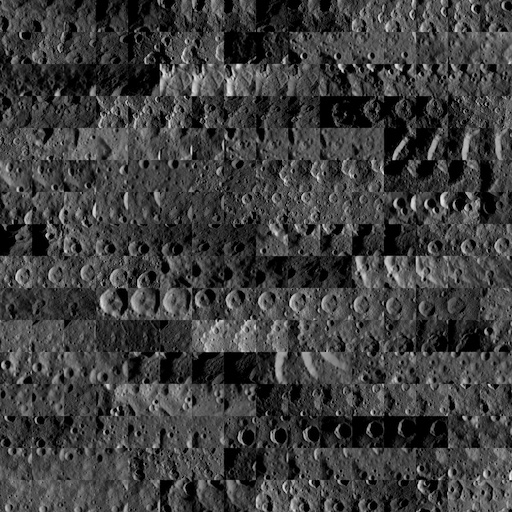}
  \caption*{\texttt{Dawn @ 1 Ceres}}
\end{subfigure}%
\hfill
\begin{subfigure}[t]{0.48\linewidth}
  \centering
  \includegraphics[width=\linewidth]{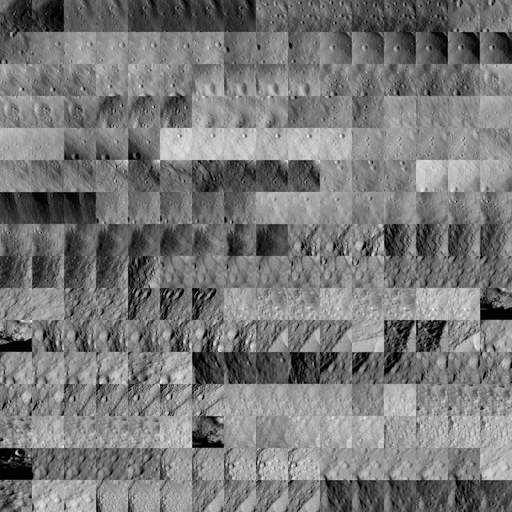}
  \caption*{\texttt{Dawn @ 4 Vesta}}
\end{subfigure}%
\caption{\textbf{Example patch mosaics.}}
\label{fig:patch-mosaics}
\end{figure}

We leverage the AstroVision dataset~\cite{driver2022astrovision} to build a patch dataset for training and testing. 
Specifically, SIFT~\cite{lowe2004ijcv} keypoints are extracted from randomly sampled image pairs and ground truth correspondences are estimated by projecting each keypoint from one image to the other using the ground truth poses and depth maps. 
A correspondence is taken to be valid if the projected keypoint falls within 3 pixels of the matched keypoint in the other image, and a bijective check is performed to ensure one-to-one correspondences, as in Ebel et al.~\cite{ebel2019iccv}. 
Finally, patches are extracted by sampling a square region of length $16\eta$ around each keypoint, where $\eta$ is the SIFT scale estimate. 
Given a set of corresponding patches, one patch was randomly chosen as the reference patch and the ground truth relative scale of the patches was taken to be the ratio of the ground sample distances with respect to the reference patch. 
Next, the ratio of the SIFT scale estimates was computed, and patches whose relative scale estimate differed by more than 25\% with respect to the ground truth value were treated as false positives and rejected, similar to Reference \citenum{brown2010tpami}. 
This process was conducted for the Dawn @ 1 Ceres, Dawn @ 4 Vesta, OSIRIS-REx @ 101955 Bennu, and Rosetta @ 67P AstroVision datasets to build corresponding patch datasets. 
Patches from each of the datasets are provided in Figure \ref{fig:patch-mosaics}. 
The total patch counts for each dataset are as follows: 572,929 patches for Dawn @ 1 Ceres, 572,672 patches for Dawn @ 4 Vesta, 573,184 patches for OSIRIS-REx @ 101955 Bennu, and 571,392 patches for Rosetta @ 67P.

%% file: Text/results.tex
We evaluate the performance of DidymosNet with respect to two different tasks, i.e., patch matching and stereo matching. 
Then, we demonstrate the runtime efficiency of our model on flight-relevant hardware. 


\subsection{Training Details}

We follow the training procedure outlined by Tian et al.~\cite{tian2020nips} 
The network is trained from scratch for 200 epochs with a batch size of 1024 (i.e., 512 corresponding patch pairs) and the Adam optimizer~\cite{kingma2015adam} with a learning rate of 0.01. Moreover, The relative perspective change between a patch pair was limited during training, where the angle of rotation between the orientation quaternions of the respective images with respect to a body-fixed frame was used as a metric for the relative perspective change between two images. 
We ignored patch pairs with a relative perspective change greater than $60^{\circ}$, as in Reference \citenum{driver2022astrovision}.


\subsection{Patch Matching}

As in previous works on feature description~\cite{brown2010tpami,ebel2019iccv,tian2020nips}, we evaluate the performance of our network on patch matching using the popular false positive rate at 95\% true positive recall (FPR95) metric, i.e., the ratio of negative descriptor pairs whose distance is within the distance threshold for 95\% recall of the true positive pairs. 
Therefore, a low FPR95 value indicates that the network is able to reliably discriminate between positive and negative descriptor pairs, which is essential for robust matching between images. 
We also provide the number of millions of floating point operations (MFLOPs) for computing a single patch descriptor as a metric for computational efficiency\footnote{The ratio between binary operations (BOPs) and FLOPs is taken to be $1\ \text{BOP} = \frac{1}{64}$\ \text{FLOP}.~\cite{liu2018eccv,rastegari2016eccv}}.
The network is trained on one dataset and tested on the remaining three, similar to the seminal UBC benchmark~\cite{brown2010tpami}. 
The results provided in Table \ref{tab:patch-match-results} demonstrate that DidymosNet is able to significantly outperform the handcrafted descriptors computed by SIFT, as the DidymosNet descriptors achieve a significantly lower FPR95 value. 
Moreover, DidymosNet achieves comparable performance to HyNet while requiring $\sim$20$\times$ fewer FLOPs. 

We also modified the network to output a \textit{binary} descriptor of length $d = 256$ (i.e., 256 bits), identical to the output length of the binary descriptors of ORB~\cite{rublee2011iccv}, which allows for efficient matching via the Hamming distance. 
We provide two different architectures for computing binary descriptors: DidymosNet$^\pm$ uses the DidymosNet architecture depicted in Figure \ref{fig:patch-arch-describe} and leverages the methodology outlined in the previous section to learn reliable binary descriptors, and DidymosNet$^{\pm'}$ uses the same architecture except that all CNN layers are quantized except for the first layer. 
As demonstrated in Table \ref{tab:patch-match-results}, our learned binary descriptors outperform the full-precision descriptors of SIFT with respect to the FPR95 metric while having a significantly smaller memory footprint. 

\begin{table*}[tp!]
\footnotesize
\scshape
\centering
\ra{1.5}
\caption{\textbf{Patch matching performance on the AstroVision dataset.} Performance is measured with respect to the false positive rate at 95\% true positive recall (FPR95). We also provide the number of millions of floating point operations (MFLOPs) for computing a single patch descriptor as a metric for computational efficiency.}
\begin{adjustbox}{width=\linewidth}
\input{Figures/patch-benchmarks}
\end{adjustbox}\\
\label{tab:patch-match-results}
\end{table*}


\subsection{Stereo Matching} 

\begin{table*}[htp!]
\footnotesize
\scshape
\centering
\ra{1.5}
\caption{\textbf{Feature matching performance on the AstroVision benchmarks.} Performance is with respect to matching precision (P), recall (R), accuracy (A), and pose AUC in percentages for 500 (1000) features. Models are trained on the Dawn @ 4 Vesta patch dataset. The \textbf{first} and \underline{second} best results are bolded and underlined, respectively.}
\begin{adjustbox}{width=\linewidth}
\input{Figures/stereo-benchmarks}
\end{adjustbox}\\
\label{tab:stereo-match-results}
\end{table*}

\begin{figure*}[hp!]
\centering
\input{Figures/eval-qual-compare-fp}
\caption{\textbf{Qualitative comparison of feature matching for the \textit{full-precision} descriptors.} Correct matches are drawn in green, and the keypoints of incorrect matches are drawn in red. Results are shown for the 1000 feature case.}
\label{fig:eval-qual-compare-fp}
\end{figure*}

\begin{figure*}[hp!]
\centering
\input{Figures/eval-qual-compare-bin}
\caption{\textbf{Qualitative comparison of feature matching for the \textit{binary} descriptors.} Correct matches are drawn in green, and the keypoints of incorrect matches are drawn in red. Results are shown for the 1000 feature case.}
\label{fig:eval-qual-compare-bin}
\end{figure*}

We evaluated our DidymosNet architecture using the AstroVision feature matching benchmark developed by Driver et al.~\cite{driver2022astrovision}. 
The AstroVision benchmark evaluates the performance of computed features on a per image pair basis using the standard metrics precision, recall, and accuracy:
\begin{equation}
    \text{precision} = \frac{\#\ \text{correct matches}}{\#\ \text{putative matches}},
\end{equation}
\begin{equation}
    \text{recall} = \frac{\#\ \text{correct matches}}{\#\ \text{ground truth matches}},
\end{equation}
\begin{equation}
    \text{accuracy} = \frac{\#\ \text{correct matches}\ \&\ \text{nonmatches}}{\#\ \text{features}}.
\end{equation}
Moreover, the quality of the relative pose computed from the putative matches is assessed using the area under the curve (AUC) of the normalized cumulative pose error curve, where the pose error is taken to be the maximum of the angular error between the estimated and ground truth pose orientation and (unit) translation in degrees. 
We refer the reader to Reference \citenum{driver2022astrovision} for more details about the AstroVision benchmarking scheme. 

We leverage the keypoint detection scheme of SIFT~\cite{lowe2004ijcv}, which is based on the difference of Gaussians (DoG) filter, to extract image patches for the description step. 
As previously described, patches are extracted by sampling a square region of length $16\eta_{\text{DoG}}$ around each keypoint, where $\eta_{\text{DoG}}$ is the scale estimate derived from the scale-space framework proposed by Lowe~\cite{lowe2004ijcv}. 
For our binary descriptor networks, DidymosNet$^\pm$ and DidymosNet$^{\pm'}$, we leveraged the efficient FAST~\cite{rosten2006eccv} keypoint detector to extract feature patches, similar to the detection method of ORB~\cite{rublee2011iccv}, in order to complement the matching efficiency of the binary descriptors. 
When using FAST keypoints, we extract patches of length $31\eta_{\text{FAST}}$, where $\eta_{\text{FAST}}$ corresponds to the scale of the octave in which the keypoint was detected. 
We compare our approach to SIFT~\cite{lowe2004ijcv}, ORB~\cite{rublee2011iccv}, HyNet~\cite{tian2020nips}, and SuperPoint~\cite{Detone_2018CVPRW}. 
We use the models trained on the Dawn @ 4 Vesta patches for HyNet, DidymosNet, DidymosNet$^\pm$, and DidymosNet$^{\pm'}$, the OpenCV implementations of ORB and SIFT, and the pre-trained model of SuperPoint provided by the authors. 
Note that the Dawn @ 4 Vesta images used to train our models are different than the tesing images, as described in Reference \citenum{driver2022astrovision}.

Results on the AstroVision benchmark computed from 500 and 1000 features are provided in Table \ref{tab:stereo-match-results}, and qualitative examples are provided in Figure \ref{fig:eval-qual-compare-fp} and Figure \ref{fig:eval-qual-compare-bin}. 
Our DidymosNet architecture achieves higher matching precision, recall, and accuracy and enables superior relative pose estimates, as indicated by the larger AUC value, as compared to both SIFT and SuperPoint for all test cases, while demonstrating comparable or better performance to its full-precision counterpart HyNet. 
DidymosNet is especially effective on the more difficult OSIRIS-REx @ 101955 Bennu and Rosetta @ 67P datatsets, which feature large perspective changes and challenging illumination conditions (see Figure \ref{fig:eval-qual-compare-fp}), outperforming both SIFT and SuperPoint by a large margin. 
Moreover, the architectures that output binary descriptors, i.e., DidymosNet$^\pm$ and DidymosNet$^{\pm'}$, outperform ORB with respect to all metrics. 


\subsection{Runtime Analysis}

We leveraged a ROCK Pi 4b, which features a dual Cortex-A72 @ 1.8 Ghz and a quad Cortex-A53 @ 1.4 Ghz, as a surrogate for the next-generation spacecraft processor~\cite{doyle2013hpsc,powell2018hpsc} and measured the runtimes for DidymosNet, HyNet~\cite{tian2020nips}, and SuperPoint~\cite{Detone_2018CVPRW} onboard the device using the daBNN inference framework~\cite{zhang2019dabnn}. 
Feature extraction is performed on full-resolution inputs, i.e., 1024$\times$1024 grayscale images. 
These results are shown in Table \ref{tab:runtimes}. 
Note that SuperPoint operates under the detect-\textit{and}-describe paradigm, i.e., features are extracted from a dense detection score and descriptor map output by the network with fixed spatial dimensions relative to the input image, so feature extraction runtimes remain relatively constant for different numbers of features. 
This is in contrast to the detect-\textit{then}-describe approach, which enables dynamic runtimes by increasing or decreasing the number of detected keypoints where description runtimes generally scale linearly with the number of extracted patches. 
For the detect-\textit{then}-describe methods, keypoint detection and patch extraction are averaged over 100 trials and rounded to the nearest millisecond (ms). 

DoG + DidymosNet is able to operate at up to $\sim$0.5 frames per second, an order of magnitude faster than SuperPoint at 500 and 1000 features and over $5\times$ faster than DoG + HyNet for all cases. 
FAST + DidymosNet$^\pm$ is slightly faster than DoG + DidymosNet due to the efficient detection scheme of FAST and the reduced computational overhead of matching binary descriptors using the Hamming distance as compared to floating-point descriptors. 
However, FAST + DidymosNet$^\pm$ is slightly slower than DoG + DidymosNet at 5000 features, as the feature description step for DidymosNet$^\pm$ is slightly more expensive than DidymosNet due to the final 256-channel CNN layer. 
FAST + DidymosNet$^{\pm'}$ is our most efficient architecture, and can achieve an almost $26\times$ speed-up relative to SuperPoint and almost $8\times$ relative to DoG + HyNet.  
Note that the theoretical speed-ups of $\sim$$20\times$ for DidymosNet and DidymosNet$^\pm$ and $\sim$$40\times$ for DidymosNet$^{\pm'}$ compared to HyNet, shown in Table \ref{tab:patch-match-results}, can be achieved by using specialized hardware comprised of FPGAs, as demonstrated in Reference \citenum{mishra2018iclr}. 
Moreover, the binary descriptors output by DidymosNet$^\pm$ and DidymosNet$^{\pm'}$ allow for efficient matching via the Hamming distance, providing an up to $4\times$ matching speed up with respect to the 128-dimensional floating point descriptors output by HyNet and DidymosNet, and an over $9\times$ speed-up compared to the 256-dimensional floating point descriptors of SuperPoint. 
We believe that these results demonstrate that DidymosNet is a feasible solution for online feature description onboard future robotic spacecraft for missions to small bodies.
Indeed, image acquisition rates typically range from one image per minute~\cite{keller2005deepimpact,shoemaker2022terrain} up to one image per 10 minutes~\cite{bos2018tag}. 
For example, during the Touch-and-Go (TAG) sample collection event of the OSIRIS-Rex mission to Asteroid 101955 Bennu, images processing rates ranged from 10 minutes at initialization down to 2 minutes following the Matchpoint maneuver~\cite{norman2022autonomous}. 

\begin{table*}[tp!]
\footnotesize
\scshape
\centering
\ra{1.5}
\caption{\textbf{Runtime comparison on ROCK Pi 4b.} Feature extraction runtimes include keypoint detection$/$patch extraction$/$feature description, and matching runtimes are for a single image pair.} 
\begin{adjustbox}{width=\linewidth}
\begin{tabular}{lrrrrrrrr}
\toprule
                  & & \multicolumn{6}{c}{Runtime (ms)} & \\
\cmidrule(lr){3-8}
                        &                & \multicolumn{2}{c}{500 features} & \multicolumn{2}{c}{1000 features} & \multicolumn{2}{c}{5000 features} & \multirow[b]{2}{1.5cm}{Model Size (Kb)} \\
\cmidrule(lr){3-4} \cmidrule(lr){5-6} \cmidrule(lr){7-8}
Feature                 & ms$/$desc.     & \multicolumn{1}{c}{Extract} & \multicolumn{1}{c}{Match} & \multicolumn{1}{c}{Extract} & \multicolumn{1}{c}{Match} & \multicolumn{1}{c}{Extract} & \multicolumn{1}{c}{Match} \\
\midrule 
SuperPoint              & --- & 34232 & 22 & 34232 & 116 & 34232 & 5893 & 5086 \\
DoG + HyNet            & 19.36 &                      10394 ($680/34/9680$) & 12 &          20108            ($680/68/19360$) & 51 &                     97816 ($680/336/96800$) & 2634 & 5223 \\
DoG + DidymosNet       &  3.55 &                       2489 ($680/34/1775$) & 12 &           4298\hspace{5pt} ($680/68/3550$) & 51 &            18766 ($680/336/17750$) & 2634 &  820 \\
\midrule 
\rowcolor[gray]{0.9}
FAST + DidymosNet$^\pm$  &  4.02 &  2110\hspace{5pt}  ($66/34/2010$) &  \textbf{5} &  4154\hspace{10pt} ($67/68/4019$) & \textbf{21} &         20516 \hspace{2.5pt} ($80/336/20100$) &  \textbf{630} &  1077 \\
\rowcolor[gray]{0.9}
FAST + DidymosNet$^{\pm'}$  &  2.45 &  \textbf{1325}\hspace{5pt}  ($66/34/1225$) &  \textbf{5} &  \textbf{2585}\hspace{10pt} ($67/68/2450$) & \textbf{21} &         \textbf{12666} \hspace{2.5pt} ($80/336/12250$) &  \textbf{630} &  \textbf{85} \\
\bottomrule
\end{tabular}
\end{adjustbox}\\
\label{tab:runtimes}
\end{table*}



%% file: Figures/patch-benchmarks.tex
\begin{tabular}{lrrrrrrrrrrrrrrrr}
\toprule
Train    & Bennu & Ceres & Vesta   & & 67P & Ceres & Vesta      & & 67P & Bennu & Vesta      & & 67P & Bennu & Ceres & \\
\cmidrule(lr){2-4} \cmidrule(lr){6-8} \cmidrule(lr){10-12} \cmidrule(lr){14-16}
Test     & \multicolumn{3}{c}{67P} & & \multicolumn{3}{c}{Bennu} & & \multicolumn{3}{c}{Ceres} & & \multicolumn{3}{c}{Vesta} & MFLOPs \\
\midrule 
\multicolumn{15}{l}{\textbf{Full precision descriptors}} \\
SIFT    & \multicolumn{3}{c}{78.41} & & \multicolumn{3}{c}{74.91} & & \multicolumn{3}{c}{72.43} & & \multicolumn{3}{c}{71.40} & --- \\
HyNet                &  7.60 &  6.71 &  5.45     & &  3.92 &  4.06 &  3.98    & & 0.02 & 0.05 & 0.01          & & 0.17 & 0.34 & 0.01 & 39.67 \\
DidymosNet           & 12.66 & 10.25 &  9.66     & &  7.43 &  6.23 &  7.43    & & 0.23 & 0.25 & 0.01          & & 0.72 & 1.15 & 0.06 &  2.84 \\
\midrule 
\rowcolor[gray]{0.9} 
\multicolumn{17}{l}{\textbf{Binary descriptors}} \\
\rowcolor[gray]{0.9} 
DidymosNet$^{\pm}$   & 22.26 & 18.12 & 17.13     & & 20.36 & 13.73 & 14.76    & & 3.29 & 1.61 & 0.12          & & 7.33 & 4.21 & 0.53 & 2.91 \\
\rowcolor[gray]{0.9} 
DidymosNet$^{\pm'}$  & 32.73 & 25.94 & 26.26      & & 35.26 & 26.49 & 27.67    & & 16.34 & 10.48 & 2.24          & & 21.76 & 17.62 & 5.32 & 1.49 \\
\bottomrule
\end{tabular}

%% file: Figures/stereo-benchmarks.tex
\begin{tabular}{lrlrrrrrrr}
\toprule
                                &             &               &            &        &       &        & \multicolumn{3}{c}{AUC}                    \\
\cmidrule(lr){8-10}
 Dataset                                    & \# Images   & Feature               & \# Matches &    P &   R &   A & \multicolumn{1}{r}{@$5^{\circ}$} & @$10^{\circ}$ & @$20^{\circ}$ \\
\midrule 
\multicolumn{2}{l}{\textbf{Full-precision descriptors}} & & & & & & \\
 Dawn @ 1 Ceres                             & 3624        & SIFT                  &     \underline{172} (\underline{346}) &  42.5 (42.9) &  78.6 (77.0) &   73.0 (72.6) &      23.0 (25.1) &       37.5 (40.1) &       49.8 (52.4) \\
                                            &             & SuperPoint            &     \textbf{191} (\textbf{368}) &  42.0 (42.7) &  78.5 (76.4) &   72.8 (71.8) &      13.2 (13.1) &       27.7 (28.1) &       42.2 (43.3) \\
                                            &             & DoG + HyNet          &     163 (325) &  \textbf{44.5} (\textbf{45.6}) &  \textbf{81.6} (\textbf{80.1}) &   \textbf{77.0} (\textbf{77.0}) &      \textbf{25.3} (\underline{28.0}) &       \textbf{41.4} (\underline{44.8}) &       \textbf{55.0} (\textbf{58.6}) \\
                                            &             & DoG + DidymosNet     &     163 (324) &  \underline{43.1} (\underline{44.1}) &  \underline{79.7} (\underline{78.4}) &   \underline{76.4} (\underline{76.3}) &      \underline{24.6} (\textbf{28.1}) &       \underline{40.4} (\textbf{45.0}) &       \underline{53.7} (\underline{58.5}) \\
\midrule 
 Dawn @ 1 Vesta                             & 2006        & SIFT                  &     \underline{176} (\underline{349}) &  36.3 (37.3) &  55.4 (54.9) &   72.8 (70.6) &      12.7 (14.6) &       21.5 (24.7) &       30.9 (34.7) \\
                                            &             & SuperPoint            &     \textbf{183} (\textbf{357}) &  33.9 (37.2) &  58.0 (55.7) &   71.6 (69.9) &      9.0 (10.2) &      17.6 (20.1) &       27.3 (31.0) \\
                                            &             & DoG + HyNet          &     174 (343) &  \textbf{46.8} (\textbf{49.6}) &  \textbf{76.1} (\textbf{75.9}) &   \textbf{79.7} (\textbf{79.0}) &      \textbf{15.0} (\textbf{18.9}) &       \textbf{26.2} (\textbf{32.2}) &       \textbf{38.4} (\textbf{45.8}) \\
                                            &             & DoG + DidymosNet     &     174 (342) &  \underline{43.5} (\underline{46.1}) &  \underline{71.0} (\underline{70.8}) &   \underline{78.0} (\underline{77.2}) &      \underline{14.3} (\underline{17.3}) &       \underline{24.6} (\underline{29.7}) &       \underline{36.1} (\underline{42.2}) \\
\midrule 
 OSIRIS-REx @ 101955 Bennu                  & 1789        & SIFT                  &     \underline{120} (\underline{264}) &  12.5 (13.2) &  17.3 (16.8) &   73.4 (68.8) &       3.5 (4.6) &        6.0 (7.5) &        8.6 (10.4) \\
                                            &             & SuperPoint            &     \textbf{148} (\textbf{287}) &  13.0 (15.6) &  22.9 (21.7) &   67.1 (63.6) &       3.4 (3.9) &        6.4 (7.3) &       9.6 (10.9) \\
                                            &             & DoG + HyNet          &     112 (236) &  \textbf{19.9} (\textbf{21.9}) &  \textbf{30.3} (\textbf{29.3}) &   \textbf{78.5} (\textbf{75.7}) &       \textbf{6.4} (\textbf{7.8}) &       \textbf{10.6} (\textbf{12.9}) &       \textbf{14.9} (\textbf{17.9}) \\
                                            &             & DoG + DidymosNet     &     103 (212) &  \underline{19.0} (\underline{21.0}) &  \underline{27.6} (\underline{26.7}) &   \underline{79.6} (\underline{77.2}) &       \underline{5.8} (\underline{7.0}) &        \underline{9.8} (\underline{11.6}) &       \underline{14.0} (\underline{16.4}) \\
\midrule 
 Rosetta @ 67P                              & 3039        & SIFT                  &     \textbf{173} (\textbf{331}) &  15.6 (15.6) &  19.2 (18.2) &   60.0 (56.1) &       1.7 (\underline{1.8}) &        3.3 (3.7) &        5.7 (6.4) \\
                                            &             & SuperPoint            &     \underline{171} (\underline{304}) &  15.7 (16.9) &  22.7 (21.5) &   58.9 (54.7) &       1.1 (1.4) &        2.7 (3.2) &        5.3 (6.0) \\
                                            &             & DoG + HyNet          &     141 (270) &  \textbf{21.3} (\textbf{21.6}) &  \textbf{26.6} (\textbf{25.0}) &   \textbf{68.3} (\textbf{65.0}) &       \textbf{2.7} (\textbf{3.0}) &        \textbf{5.2} (\textbf{5.9}) &        \textbf{9.1} (\textbf{10.0}) \\
                                            &             & DoG + DidymosNet     &     139 (266) &  \underline{19.9} (\underline{20.1}) &  \underline{24.7} (\underline{23.3}) &   \underline{68.1} (\underline{64.8}) &       \underline{2.1} (\textbf{3.0}) &        \underline{4.5} (\underline{5.7}) &        \underline{7.7} (\underline{9.1}) \\
\midrule 
\rowcolor[gray]{0.9} 
\multicolumn{2}{l}{\textbf{Binary descriptors}} & & & & & & & & \\
\rowcolor[gray]{0.9}
 Dawn @ 1 Ceres                             & 3624        & ORB                        &     \textbf{157} (\textbf{303}) &  39.9 (41.2) &  67.2 (66.2) &   73.0 (73.7) &       5.0 (\underline{6.3}) &       12.4 (\underline{15.0}) &       21.6 (\underline{25.4}) \\
\rowcolor[gray]{0.9}
                                            &             & FAST + DidymosNet$^\pm$     &     145 (288) &  \textbf{47.4} (\textbf{48.5}) &  \textbf{75.0} (\textbf{74.0}) &   \textbf{78.8} (\textbf{78.9}) &       \textbf{6.6} (\textbf{7.6}) &       \textbf{15.3} (\textbf{17.8}) &       \textbf{25.9} (\textbf{29.8}) \\
\rowcolor[gray]{0.9}
                                            &             & FAST + DidymosNet$^{\pm'}$  &     \underline{150} (\underline{296}) &  \underline{42.0} (\underline{42.6}) &  \underline{68.2} (\underline{66.5}) &   \underline{74.6} (\underline{74.6}) &       \underline{5.3} (\underline{6.3}) &       \underline{12.8} (14.8) &       \underline{22.6} (\underline{25.4}) \\
\midrule 
\rowcolor[gray]{0.9}
 Dawn @ 1 Vesta                             & 2006        & ORB                        &     \textbf{\textbf{163}} (\textbf{316}) &  30.2 (30.9) &  51.7 (49.6) &   73.0 (73.0) &       3.2 (4.4) &        7.1 (9.5) &       12.8 (16.6) \\
\rowcolor[gray]{0.9}
                                            &             & FAST + DidymosNet$^\pm$     &     151 (301) &  \textbf{40.8} (\textbf{41.7}) &  \textbf{65.2} (\textbf{63.3}) &   \textbf{80.0} (\textbf{79.7}) &       \textbf{4.4} (\textbf{5.8}) &       \textbf{10.1} (\textbf{13.2}) &       \textbf{18.2} (\textbf{23.0}) \\
\rowcolor[gray]{0.9}
                                            &             & FAST + DidymosNet$^{\pm'}$  &     \underline{152} (\underline{303}) &  \underline{34.8} (\underline{35.2}) &  \underline{55.7} (\underline{53.8}) &   \underline{76.4} (\underline{75.8}) &       \underline{3.9} (\underline{4.8}) &        \underline{8.4} (\underline{10.8}) &       \underline{15.0} (\underline{18.7}) \\
\midrule 
\rowcolor[gray]{0.9}
 OSIRIS-REx @ 101955 Bennu                  & 1789        & ORB                        &     \textbf{155} (\textbf{307}) &  10.4 (10.9) &  16.2 (14.9) &   66.6 (65.4) &       0.6 (1.0) &        1.6 (2.4) &        3.4 (4.4) \\
\rowcolor[gray]{0.9}
                                            &             & FAST + DidymosNet$^\pm$     &     133 (266) &  \textbf{16.6} (\textbf{17.7}) &  \textbf{25.3} (\textbf{23.9}) &   \textbf{74.2} (\textbf{73.1}) &       \textbf{1.8} (\textbf{2.6}) &        \textbf{3.6} (\textbf{5.3}) &        \textbf{6.6} (\textbf{8.9}) \\
\rowcolor[gray]{0.9}
                                            &             & FAST + DidymosNet$^{\pm'}$  &     \underline{136} (\underline{275}) &  \underline{13.6} (\underline{14.2}) &  \underline{20.0} (\underline{18.6}) &   \underline{71.5} (\underline{70.0}) &       \underline{1.1} (\underline{1.6}) &        \underline{2.8} (\underline{3.4}) &        \underline{5.1} (\underline{6.2}) \\
\midrule 
\rowcolor[gray]{0.9}
 Rosetta @ 67P                              & 3039        & ORB                        &     \textbf{157} (\textbf{303}) &  14.5 (14.7) &  16.8 (16.1) &   67.2 (65.8) &       0.3 (0.4) &        0.8 (1.0) &        2.0 (2.3) \\
\rowcolor[gray]{0.9}
                                            &             & FAST + DidymosNet$^\pm$     &     \underline{150} (\underline{293}) &  \textbf{18.5} (\textbf{18.9}) &  \textbf{22.4} (\textbf{21.4}) &   \textbf{71.1} (\textbf{69.4}) &       \textbf{0.6} (\textbf{0.7}) &        \textbf{1.5} (\textbf{1.7}) &        \textbf{3.2} (\textbf{3.7}) \\
\rowcolor[gray]{0.9}
                                            &             & FAST + DidymosNet$^{\pm'}$  &     148 (289) &  \underline{16.3} (\underline{16.4}) &  \underline{18.7} (\underline{17.7}) &   \underline{69.5} (\underline{67.7}) &       \underline{0.5} (\underline{0.5}) &        \underline{1.0} (\underline{1.2}) &        \underline{2.2} (\underline{2.8}) \\
\bottomrule
\end{tabular}

%% file: Figures/eval-qual-compare-fp.tex
\begin{minipage}{1.23\textwidth}
\hspace{5pt}
\begin{subfigure}[t]{0.20\linewidth}
  \centering
  \begin{subfigure}[t]{\linewidth}
    \includegraphics[width=.935\linewidth,right]{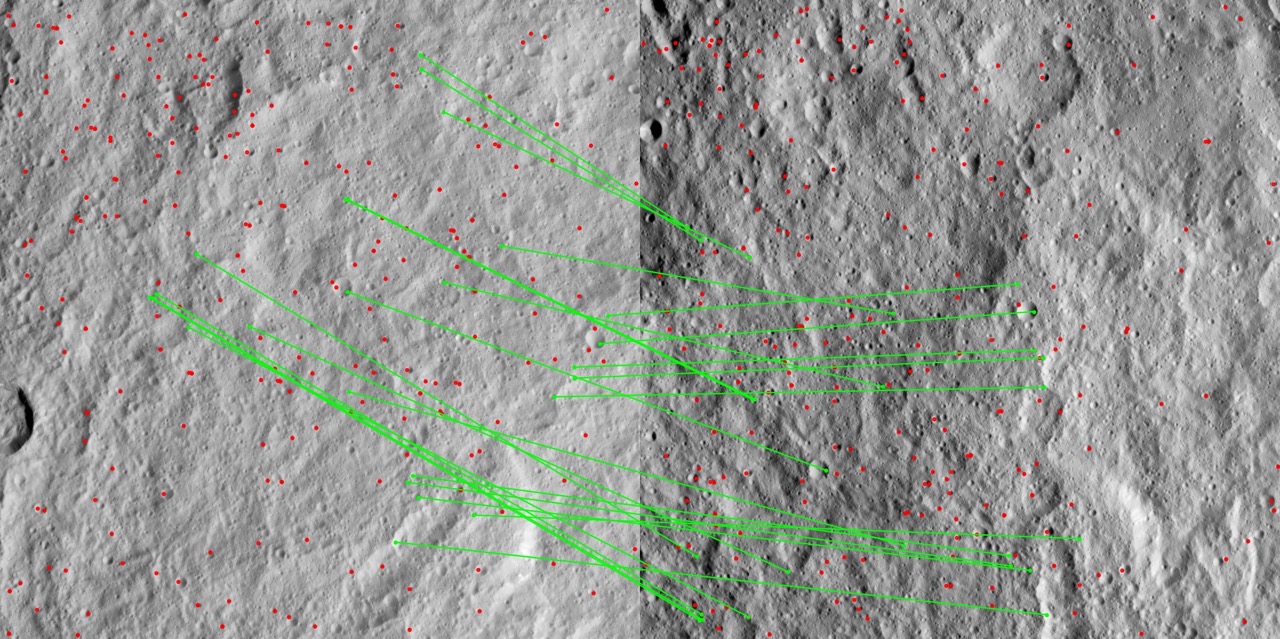}
  \end{subfigure}\\
  \vspace{0.5pt}
  \begin{subfigure}[t]{\linewidth}
    \begin{tabular}{c c}
         \hspace{-10.5pt}
         \rotatebox[origin=l]{90}{\tiny{(a) \texttt{Ceres}}}
         \hspace{-18pt}
         &
         \includegraphics[width=.935\linewidth,center]{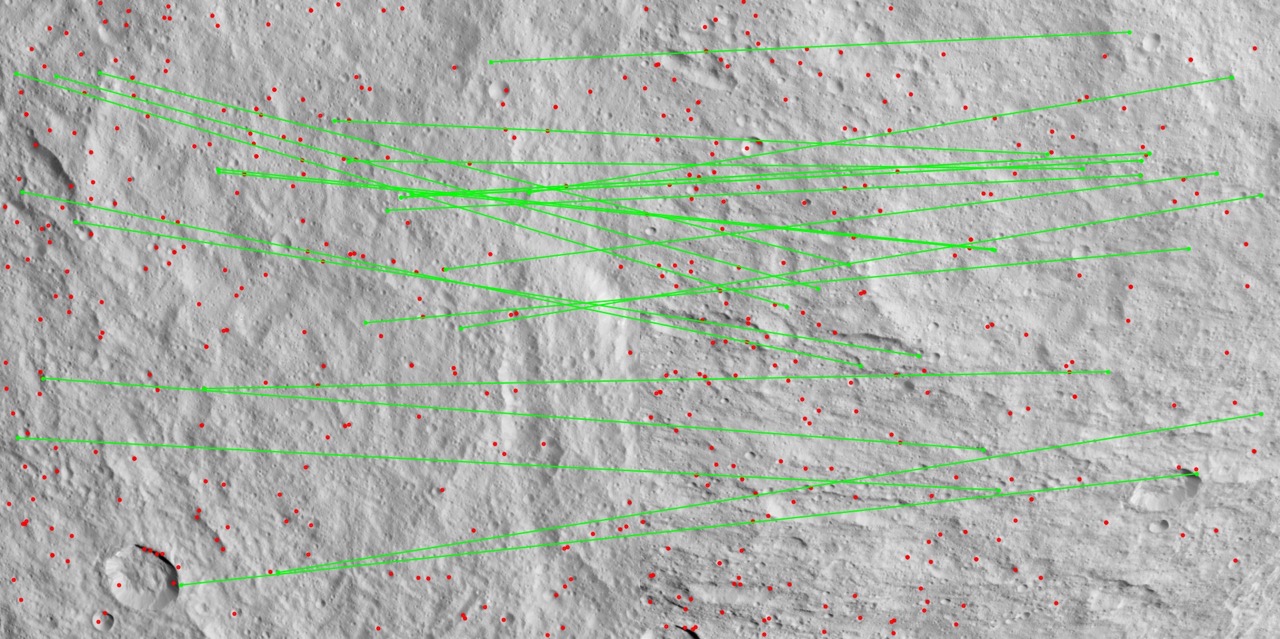}
    \end{tabular}
    \refstepcounter{subfigure}\label{fig:eval-ceres}
  \end{subfigure}\\
  \vspace{-1pt}
  \begin{subfigure}[t]{\linewidth}
    \includegraphics[width=.935\linewidth,right]{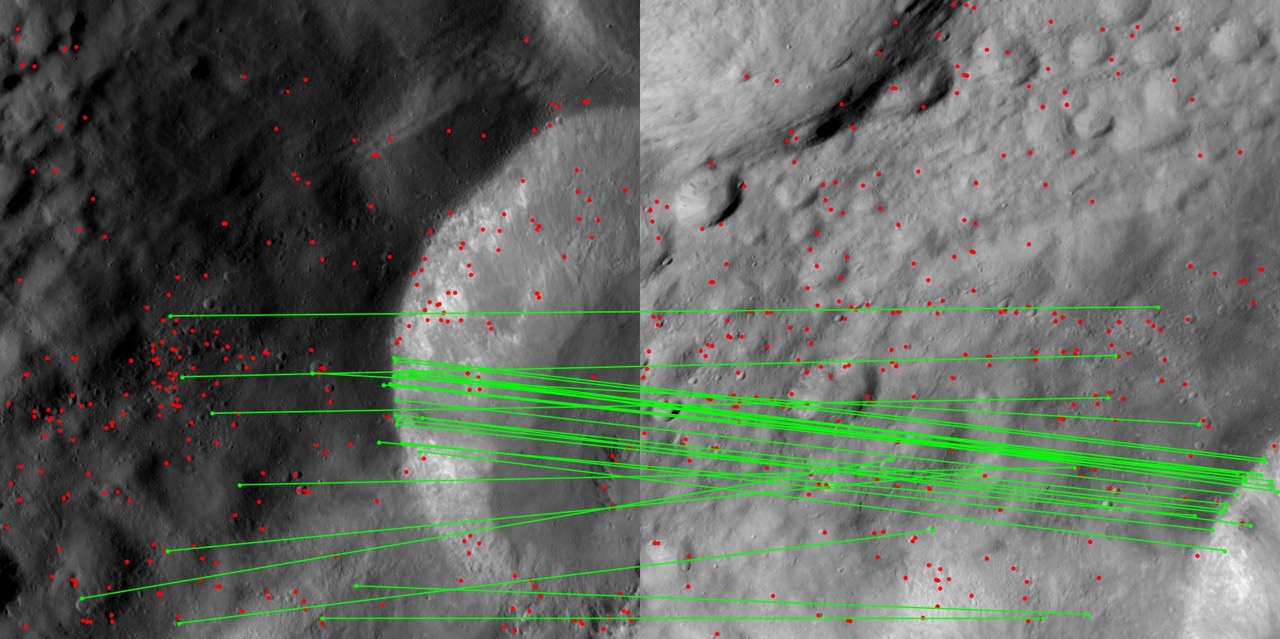}
  \end{subfigure}\\
  \begin{subfigure}[t]{\linewidth}
    \includegraphics[width=.935\linewidth,right]{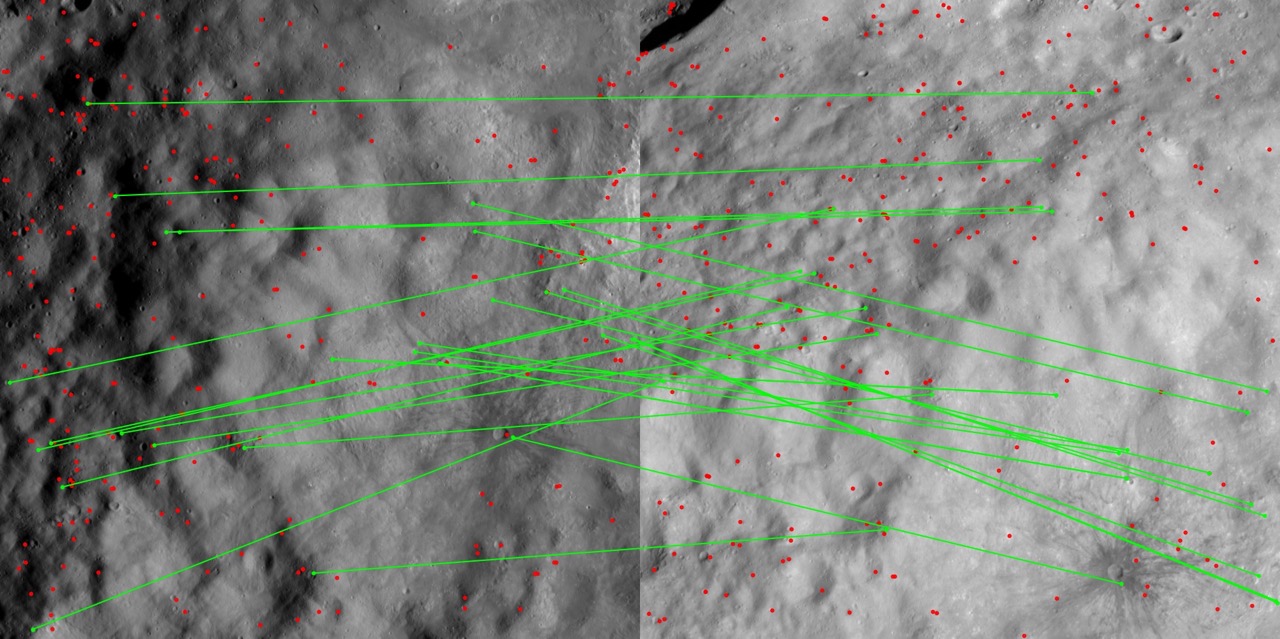}
  \end{subfigure}\\
  \vspace{0.5pt}
  \begin{subfigure}[t]{\linewidth}
    \begin{tabular}{c c}
         \hspace{-10.5pt}
         \rotatebox[origin=l]{90}{\tiny{(b) \texttt{Vesta}}}
         \hspace{-20.75pt}
         &
         \includegraphics[width=.935\linewidth,center]{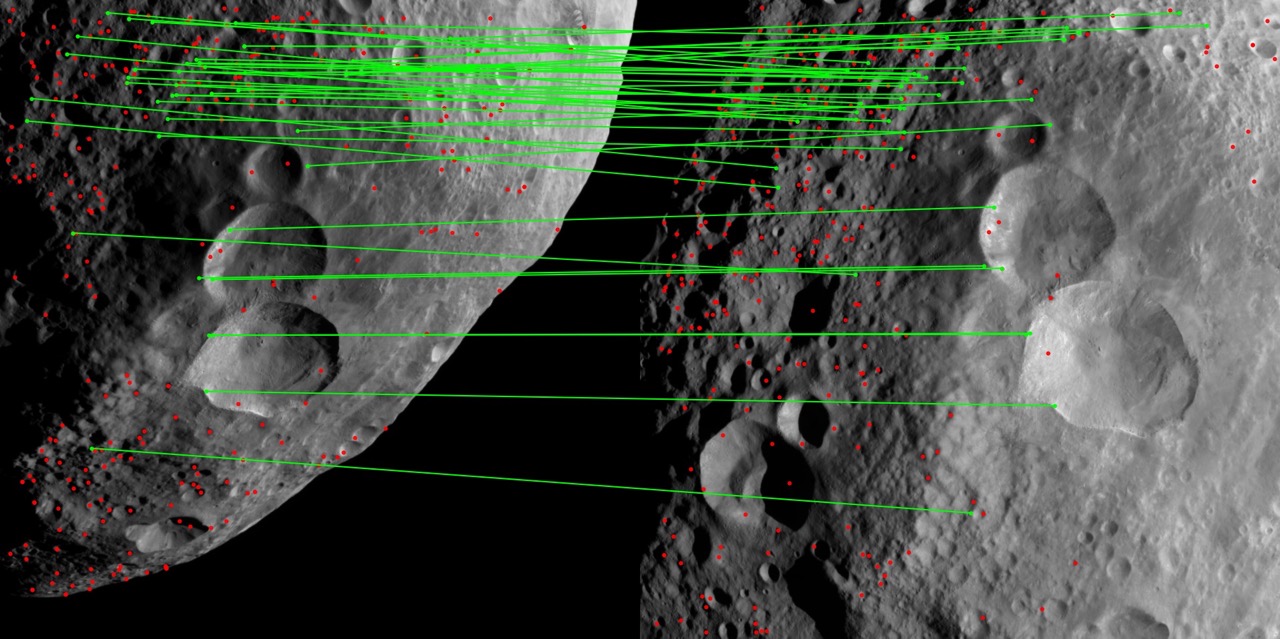}
    \end{tabular}
    \refstepcounter{subfigure}\label{fig:eval-vesta}
  \end{subfigure}\\
  \vspace{-1pt}
  \begin{subfigure}[t]{\linewidth}
    \includegraphics[width=.935\linewidth,right]{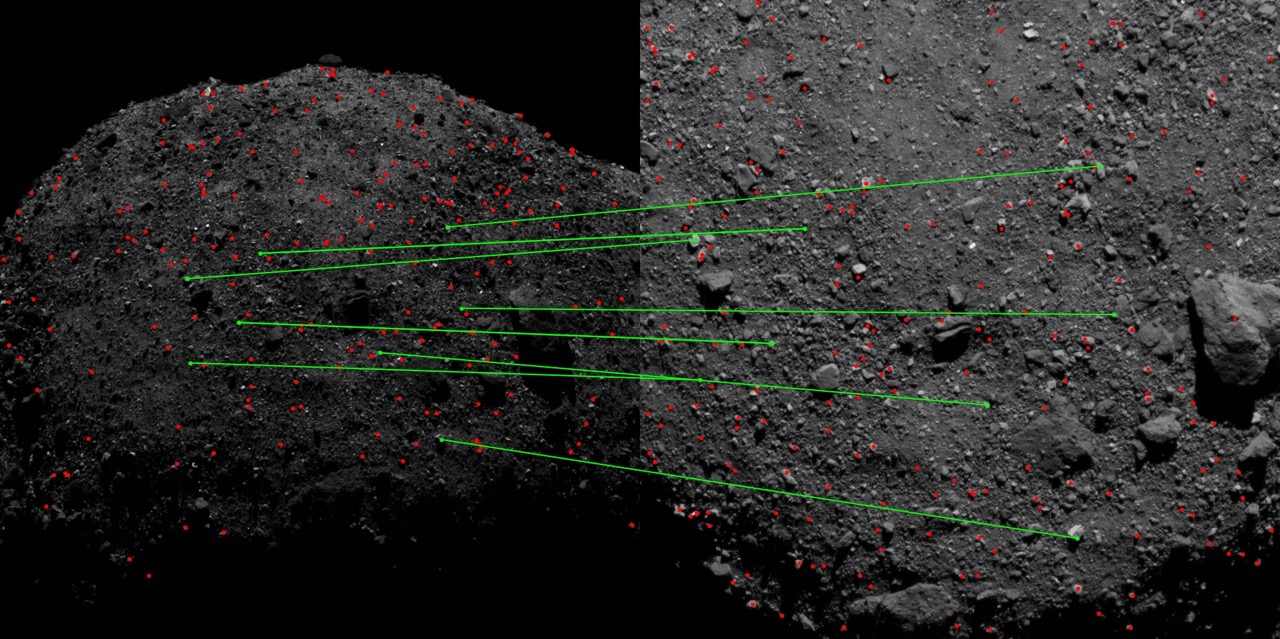}
  \end{subfigure}\\
  \begin{subfigure}[t]{\linewidth}
    \includegraphics[width=.935\linewidth,right]{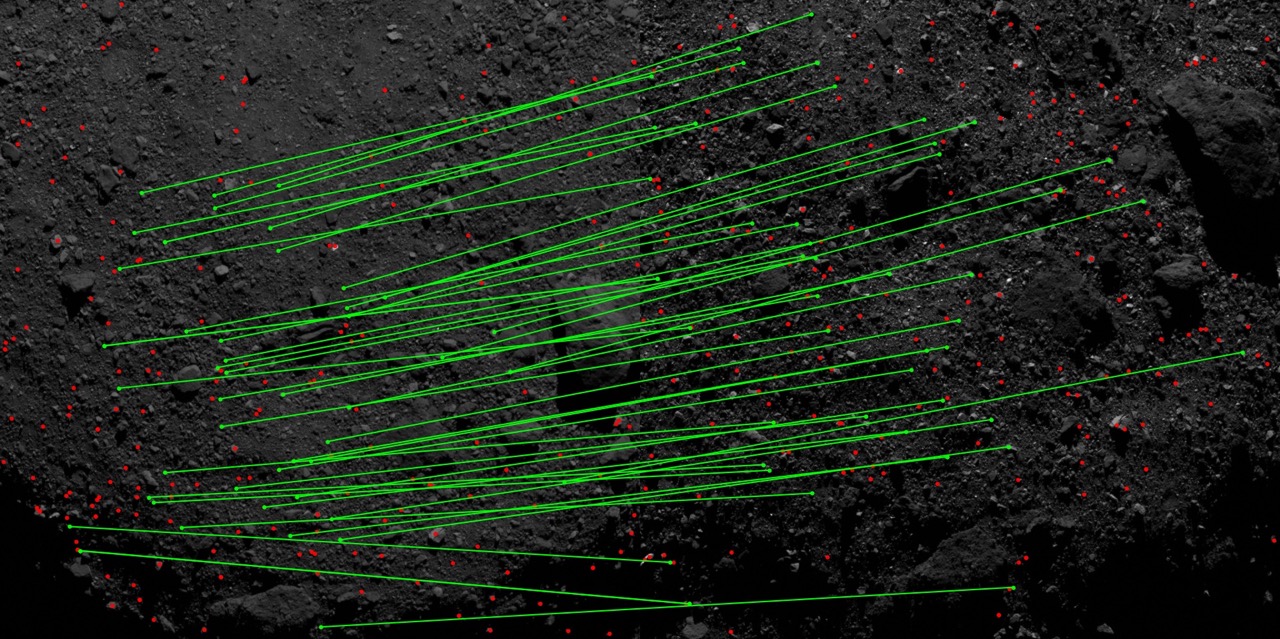}
  \end{subfigure}\\
  \vspace{0.5pt}
  \begin{subfigure}[t]{\linewidth}
    \begin{tabular}{c c}
         \hspace{-10.5pt}
         \rotatebox[origin=l]{90}{\tiny{(c) \texttt{Bennu}}}
         \hspace{-17pt}
         &
         \includegraphics[width=.935\linewidth]{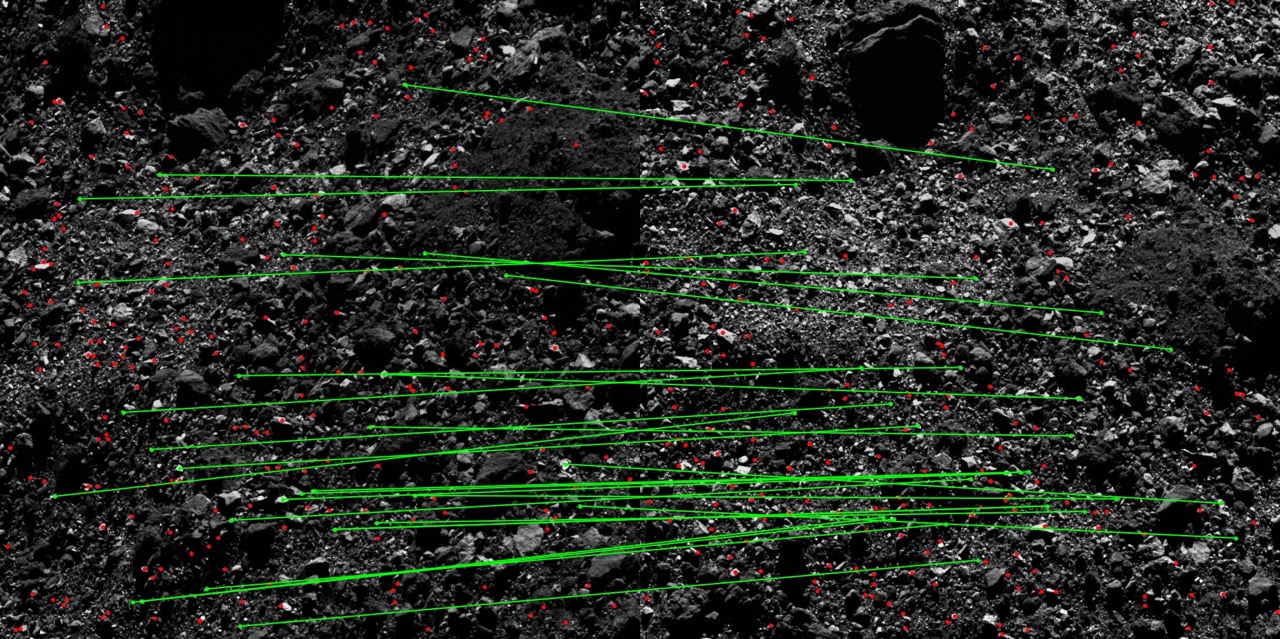}
    \end{tabular}
    \refstepcounter{subfigure}\label{fig:eval-bennu}
  \end{subfigure}\\
  \vspace{-1pt}
  \begin{subfigure}[t]{\linewidth}
    \includegraphics[width=.935\linewidth,right]{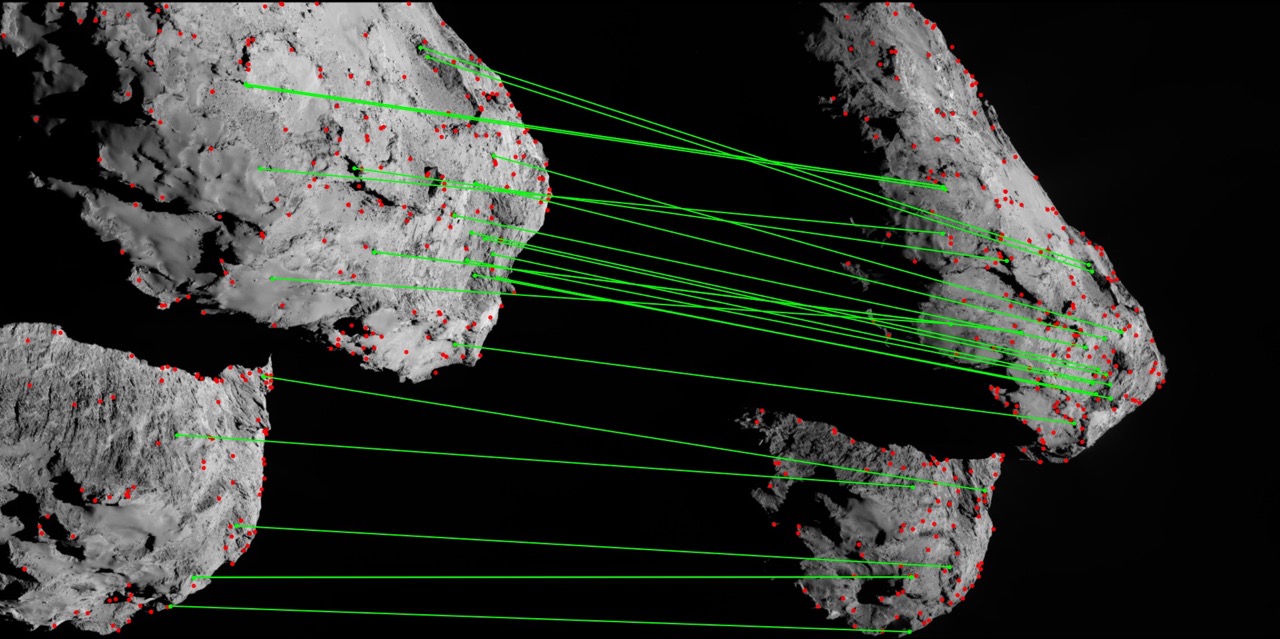}
  \end{subfigure}\\
  \begin{subfigure}[t]{\linewidth}
    \includegraphics[width=.935\linewidth,right]{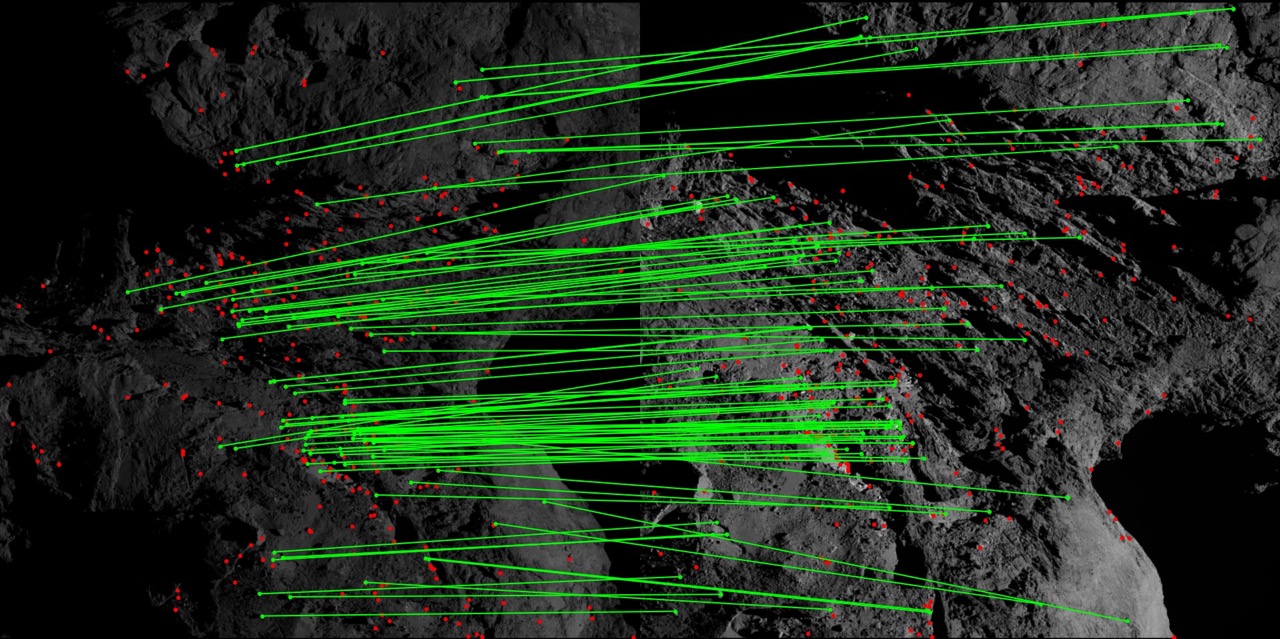}
  \end{subfigure}\\
  \vspace{0.5pt}
  \begin{subfigure}[t]{\linewidth}
    \begin{tabular}{c c}
         \hspace{-10.5pt}
         \rotatebox[origin=l]{90}{\tiny{(d) \texttt{67P}}}
         \hspace{-17pt}
         &
         \includegraphics[width=.935\linewidth]{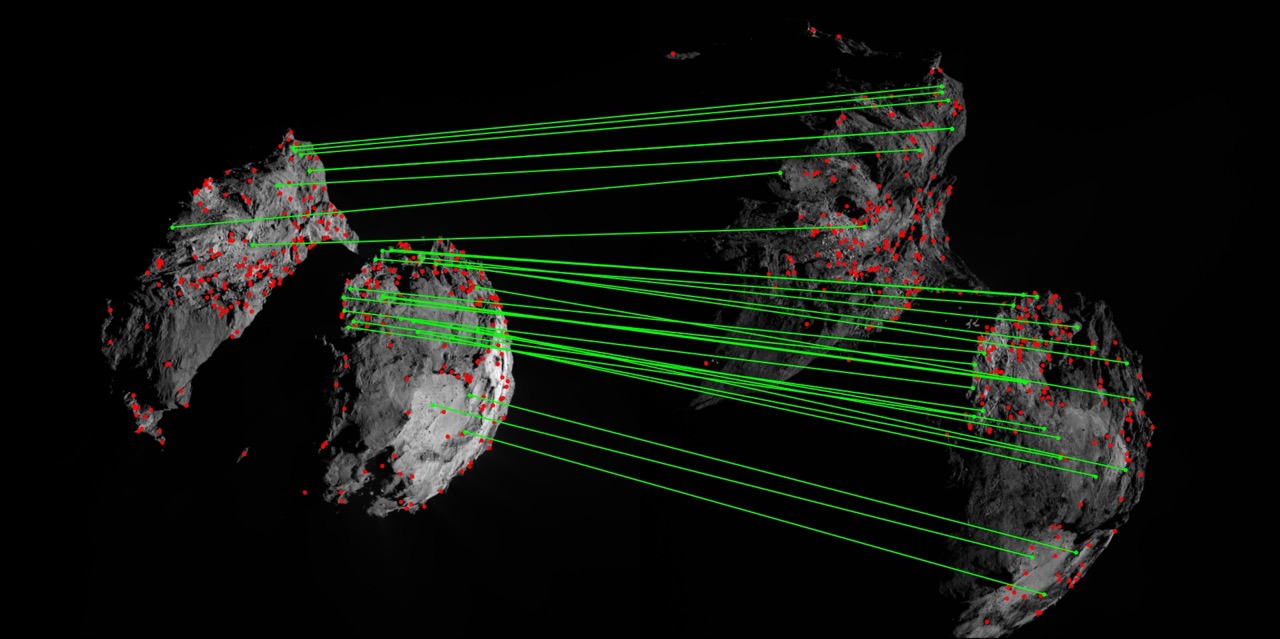}
    \end{tabular}
    \refstepcounter{subfigure}\label{fig:eval-chury}
  \end{subfigure}%
  \vspace{-5.5pt}
  \caption*{\footnotesize{SIFT}}
\end{subfigure}
\hspace{1pt}
\begin{subfigure}[t]{0.20\linewidth}
  \centering
  \begin{subfigure}[t]{\linewidth}
    \includegraphics[width=.935\linewidth]{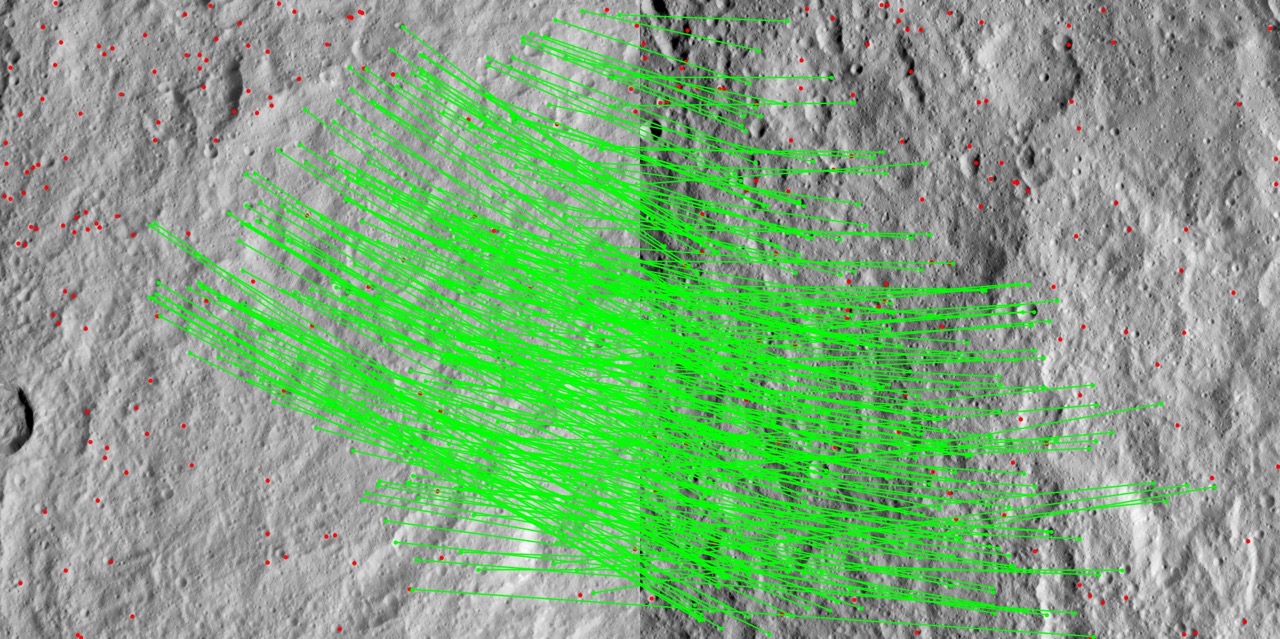}
  \end{subfigure}\\
  \vspace{0.5pt}
  \begin{subfigure}[t]{\linewidth}
    \includegraphics[width=.935\linewidth]{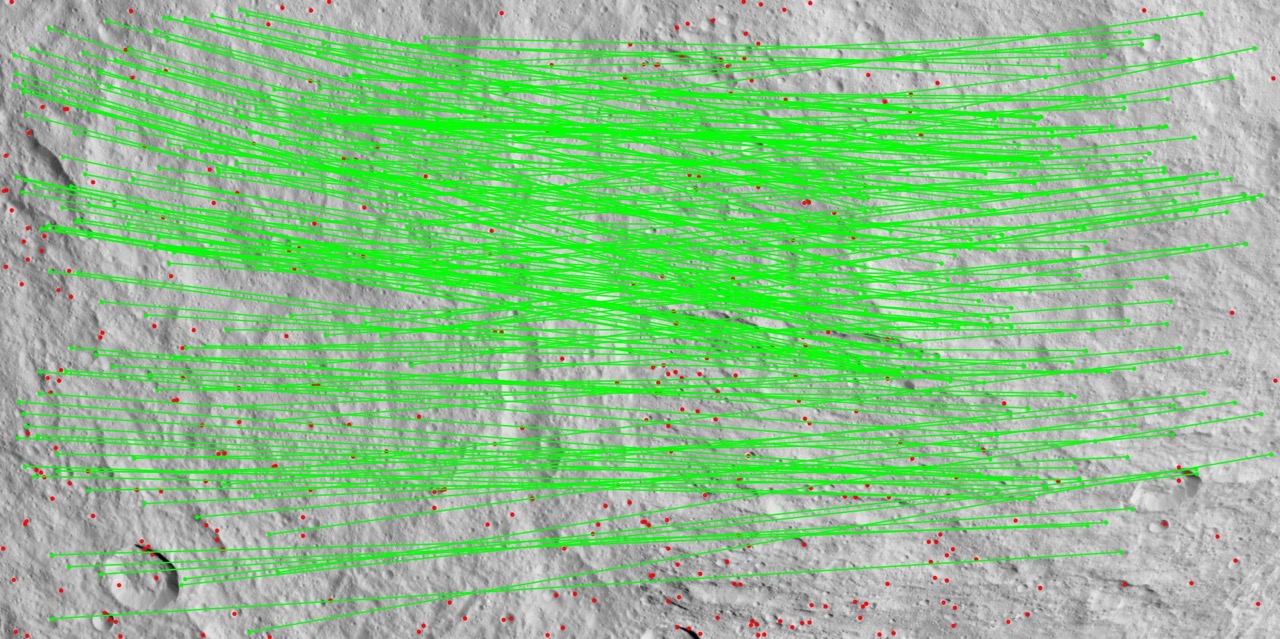}
  \end{subfigure}\\
  \vspace{4pt}
  \begin{subfigure}[t]{\linewidth}
    \includegraphics[width=.935\linewidth]{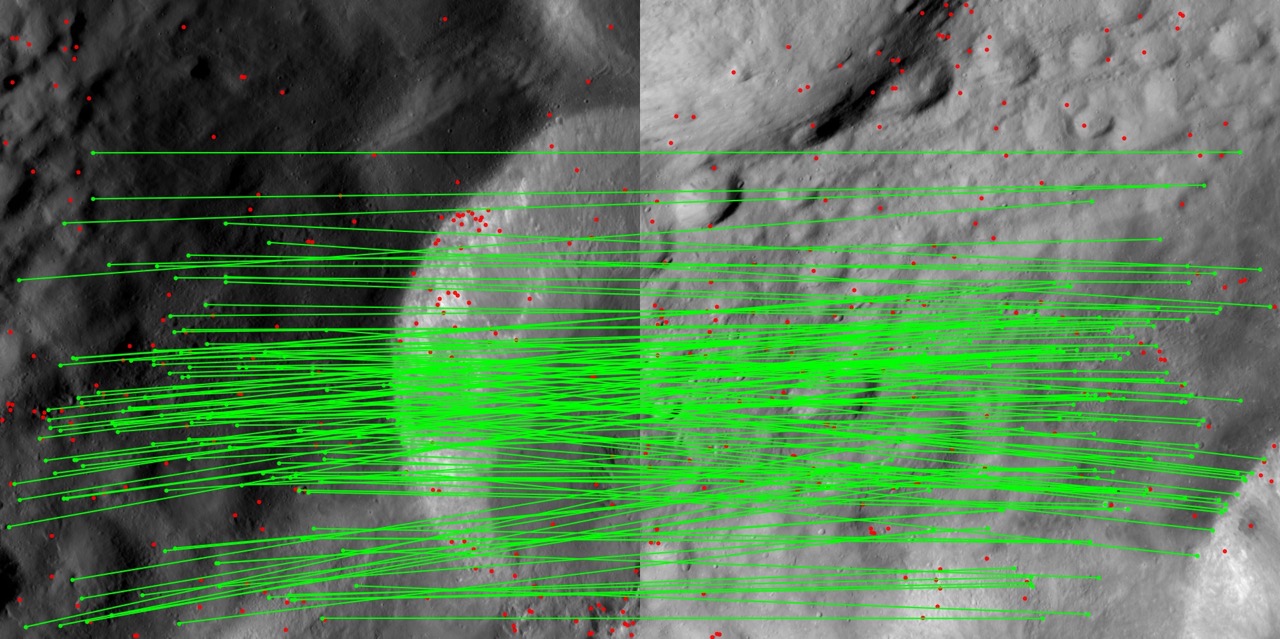}
  \end{subfigure}\\
  \begin{subfigure}[t]{\linewidth}
    \includegraphics[width=.935\linewidth]{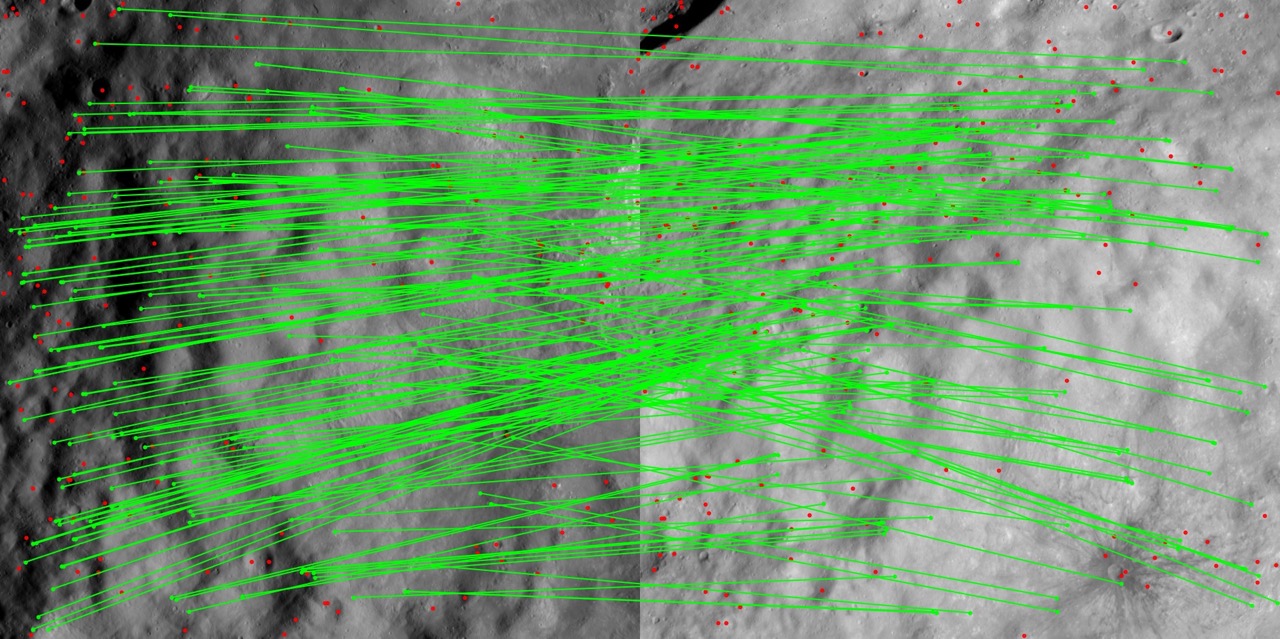}
  \end{subfigure}\\
  \vspace{0.5pt}
  \begin{subfigure}[t]{\linewidth}
    \includegraphics[width=.935\linewidth]{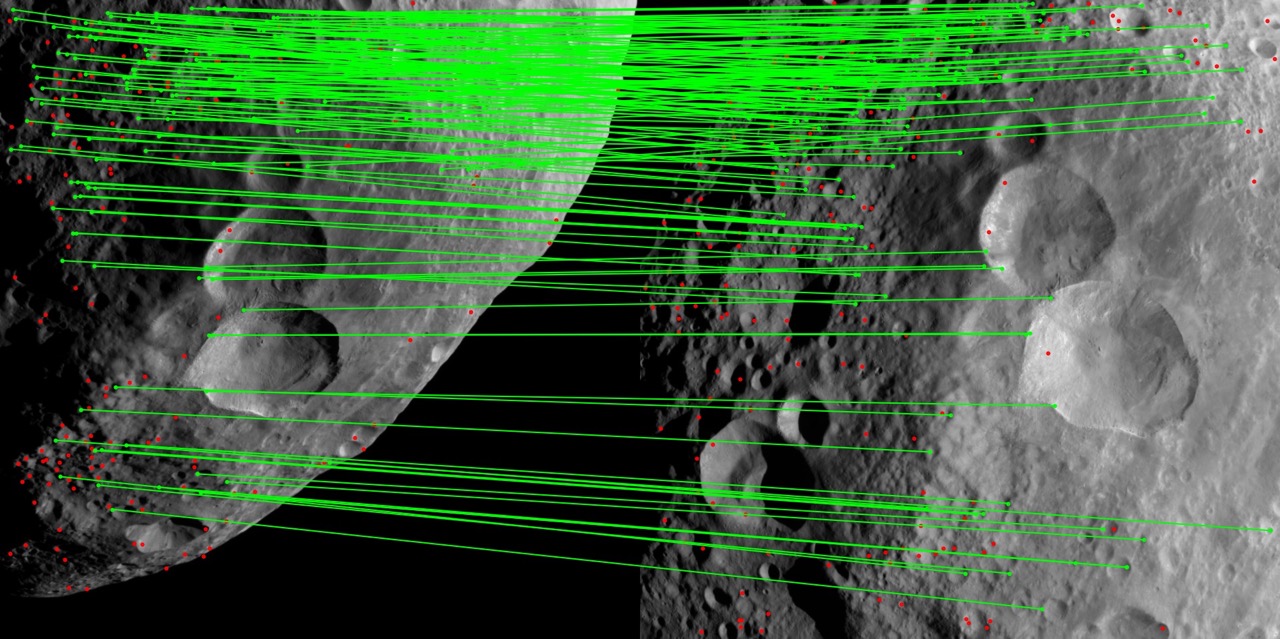}
  \end{subfigure}\\
  \vspace{4pt}
  \begin{subfigure}[t]{\linewidth}
    \includegraphics[width=.935\linewidth]{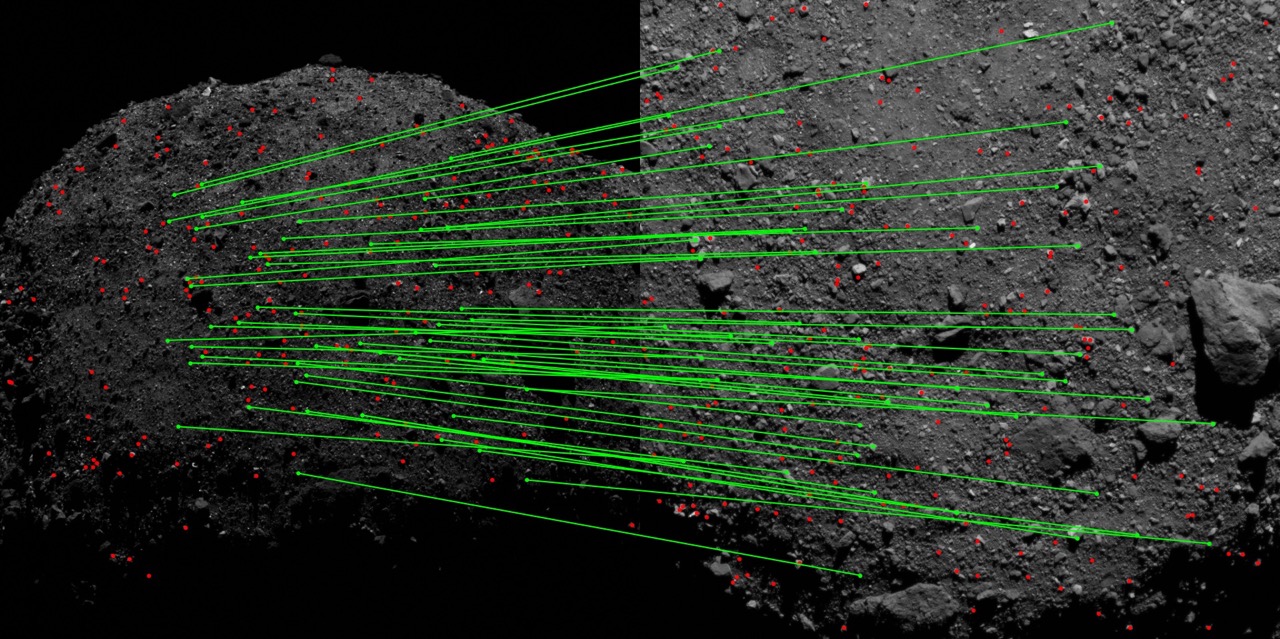}
  \end{subfigure}\\
  \begin{subfigure}[t]{\linewidth}
    \includegraphics[width=.935\linewidth]{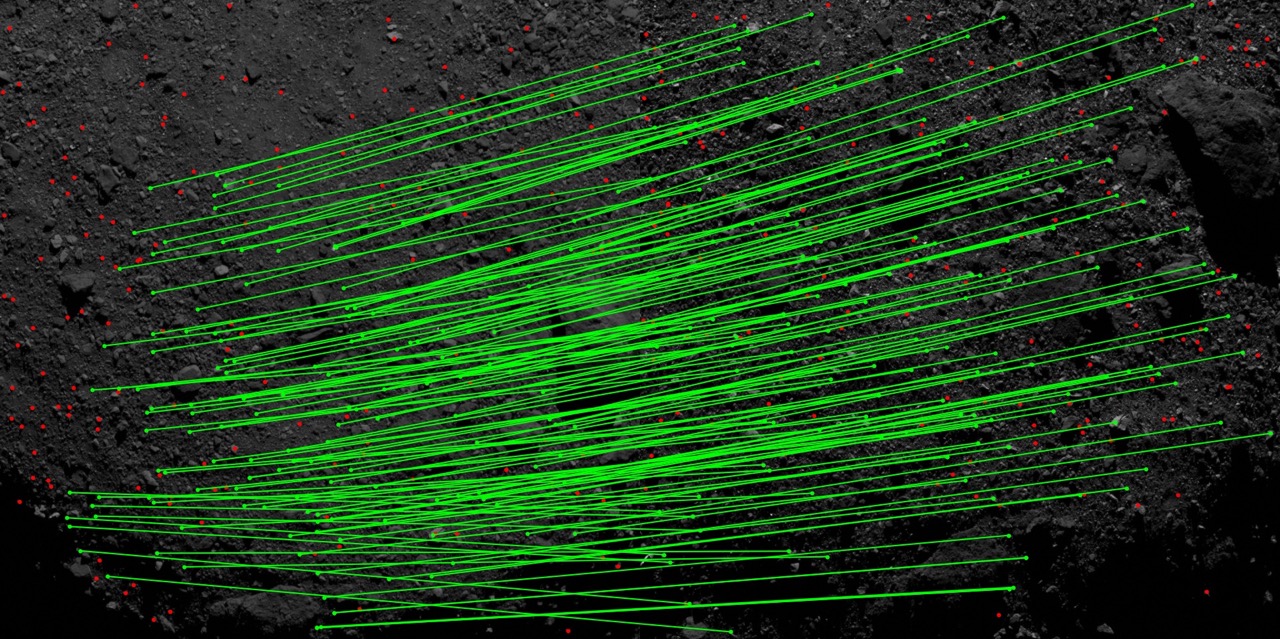}
  \end{subfigure}\\
  \vspace{0.5pt}
  \begin{subfigure}[t]{\linewidth}
    \includegraphics[width=.935\linewidth]{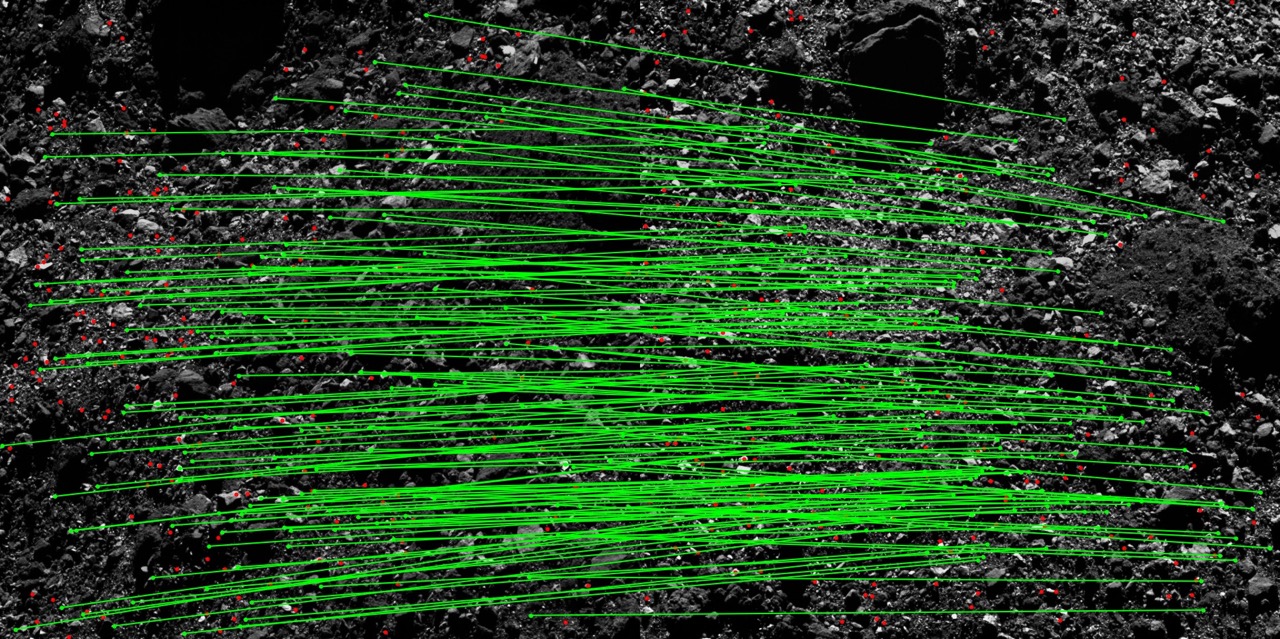}
  \end{subfigure}\\
  \vspace{4pt}
  \begin{subfigure}[t]{\linewidth}
    \includegraphics[width=.935\linewidth]{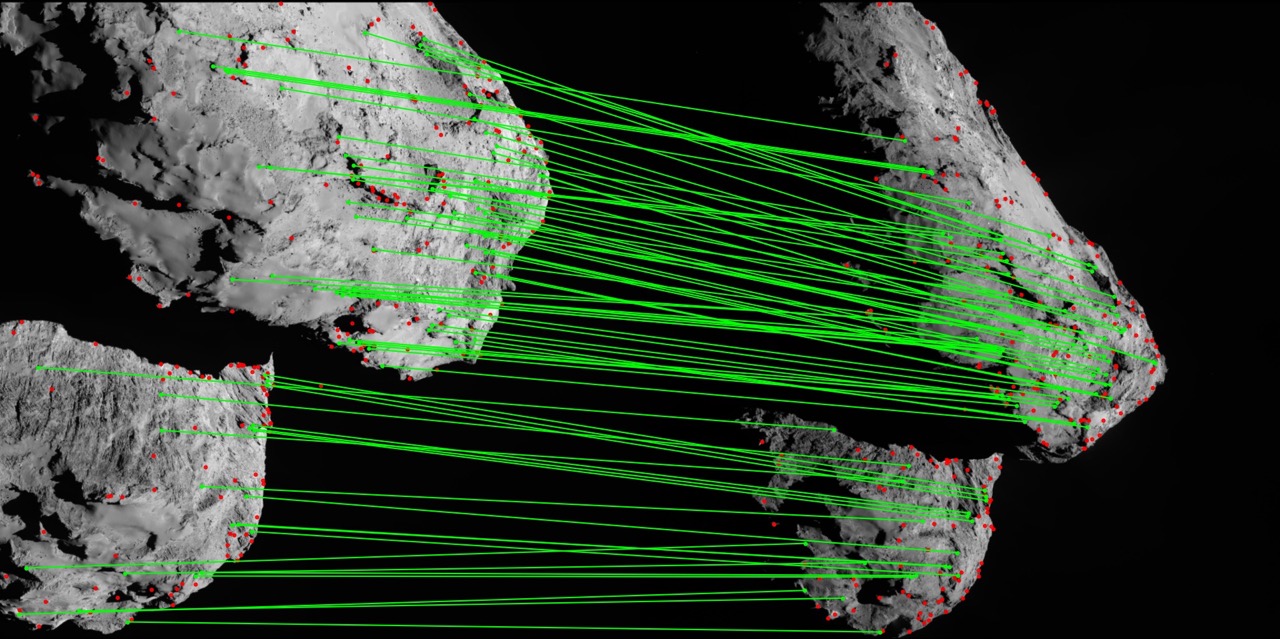}
  \end{subfigure}\\
  \begin{subfigure}[t]{\linewidth}
    \includegraphics[width=.935\linewidth]{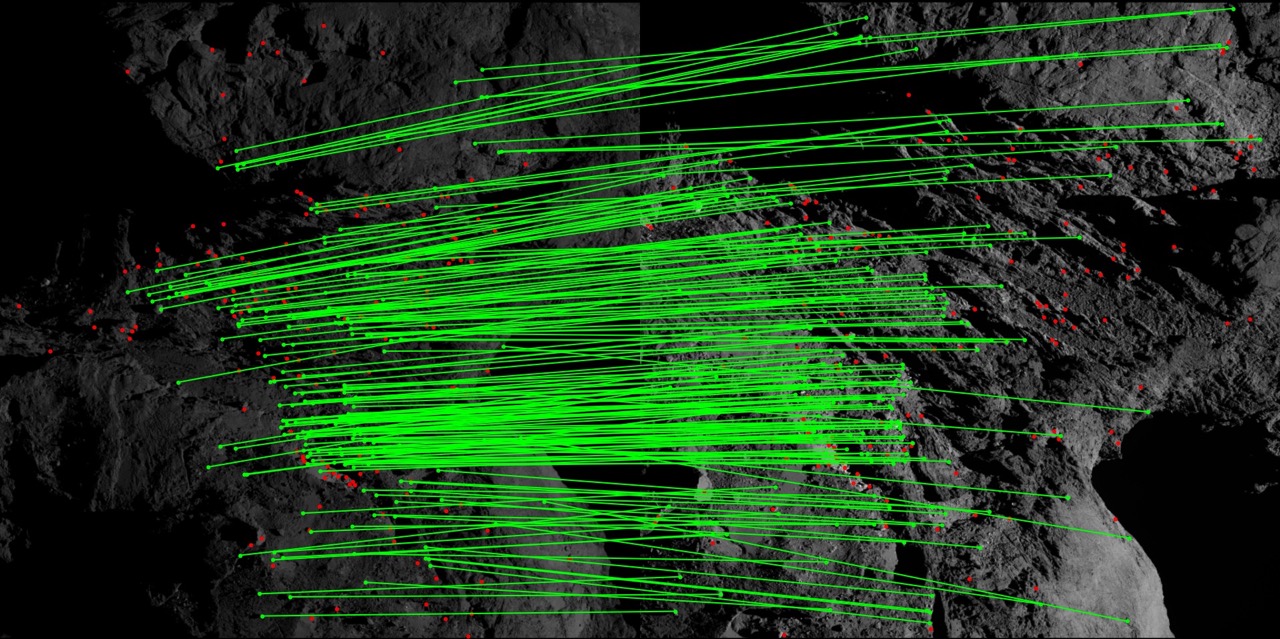}
  \end{subfigure}\\
    \vspace{0.5pt}
  \begin{subfigure}[t]{\linewidth}
    \includegraphics[width=.935\linewidth]{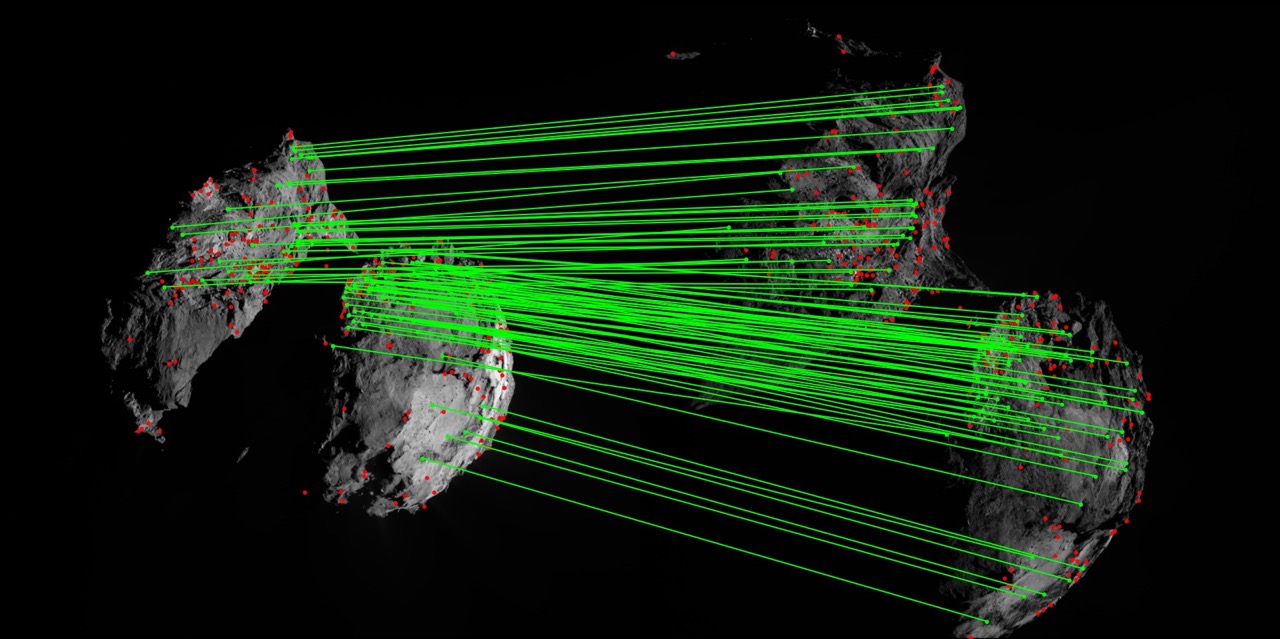}
  \end{subfigure}%
  \caption*{\footnotesize{DoG + HyNet}}
\end{subfigure}
\hspace{-5.5pt}
\begin{subfigure}[t]{0.20\linewidth}
  \centering
  \begin{subfigure}[t]{\linewidth}
    \includegraphics[width=.935\linewidth]{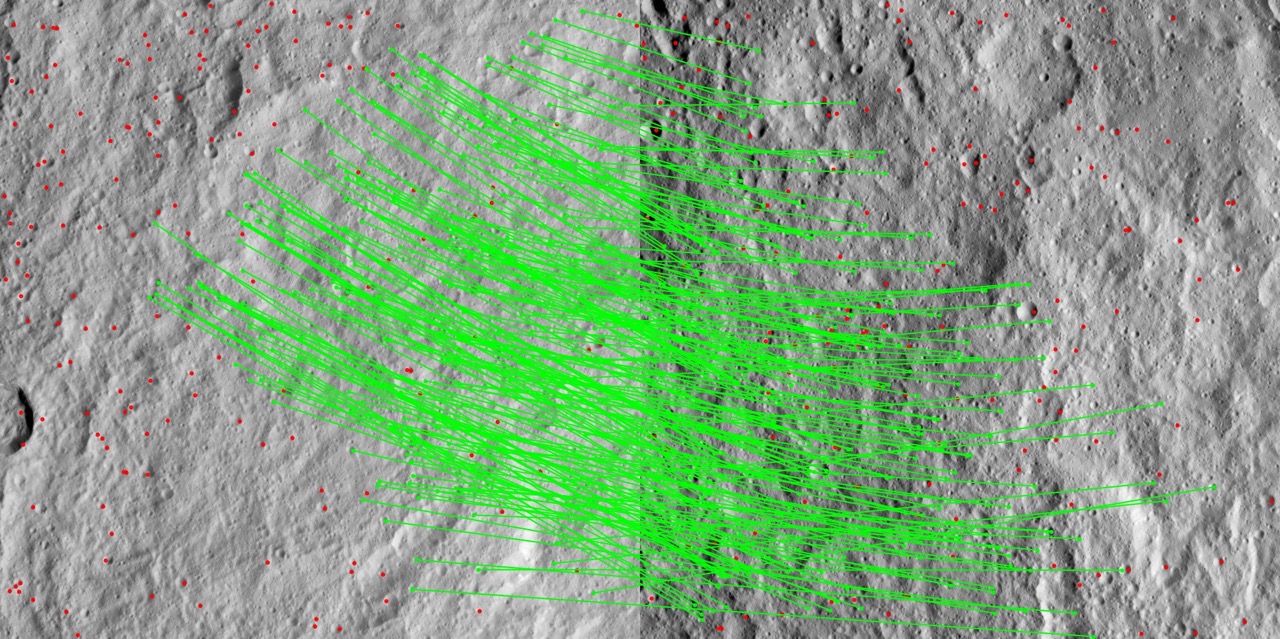}
  \end{subfigure}\\
  \vspace{0.5pt}
  \begin{subfigure}[t]{\linewidth}
    \includegraphics[width=.935\linewidth]{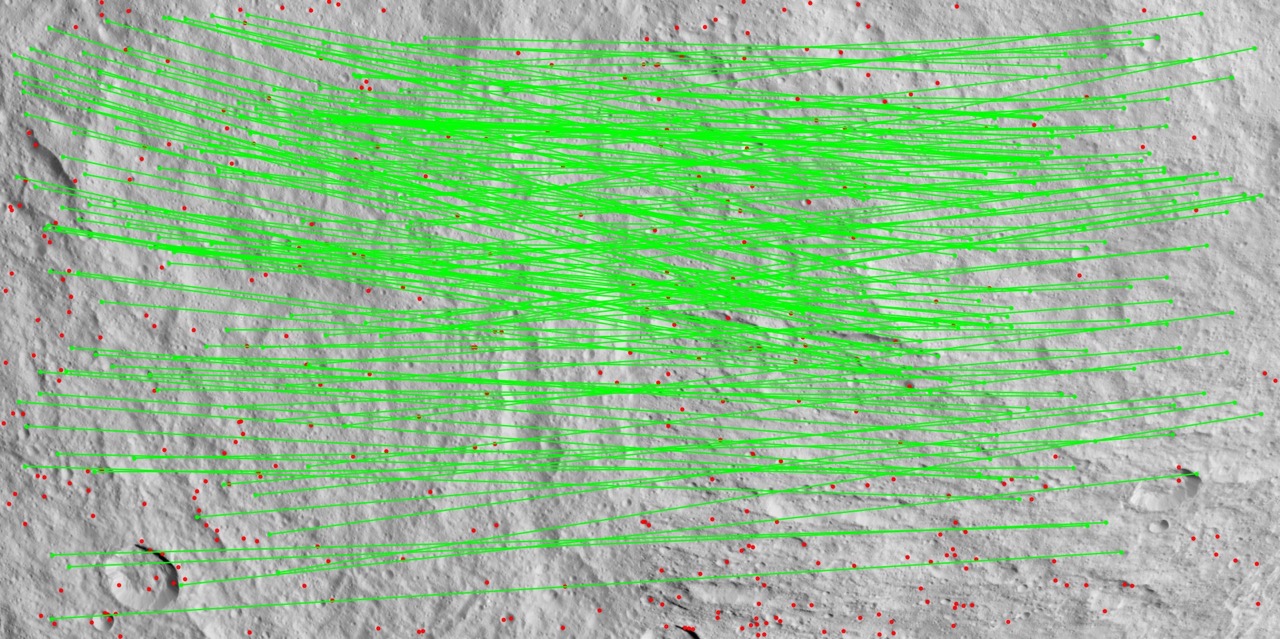}
  \end{subfigure}\\
  \vspace{4pt}
  \begin{subfigure}[t]{\linewidth}
    \includegraphics[width=.935\linewidth]{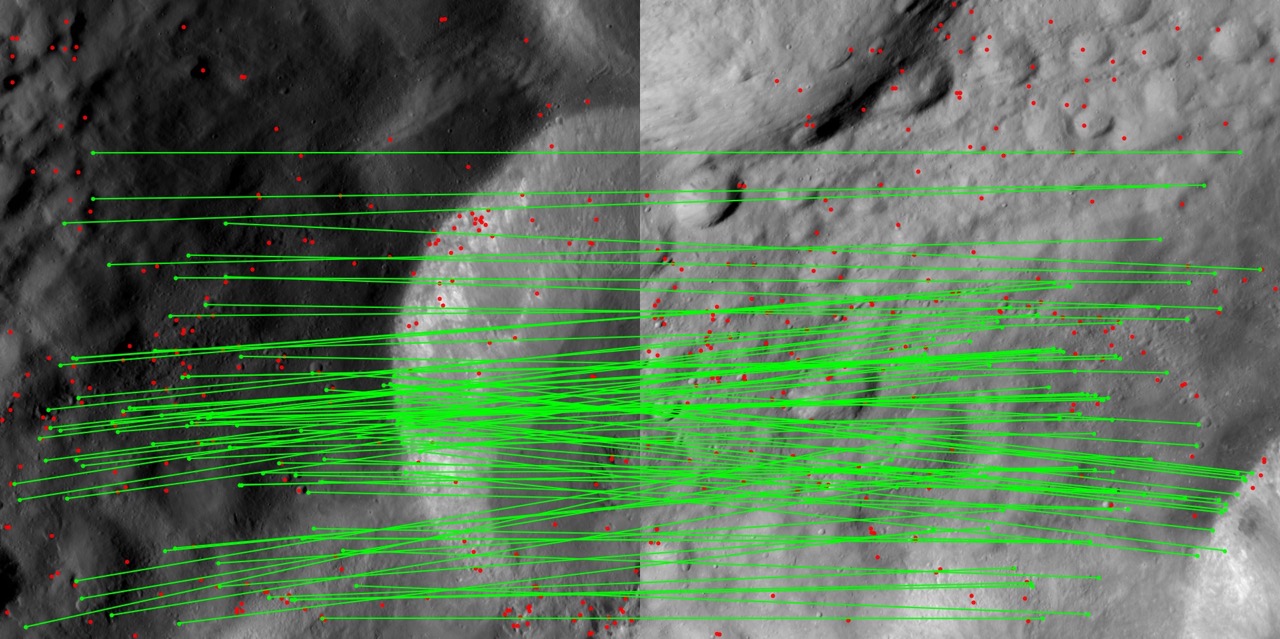}
  \end{subfigure}\\
  \begin{subfigure}[t]{\linewidth}
    \includegraphics[width=.935\linewidth]{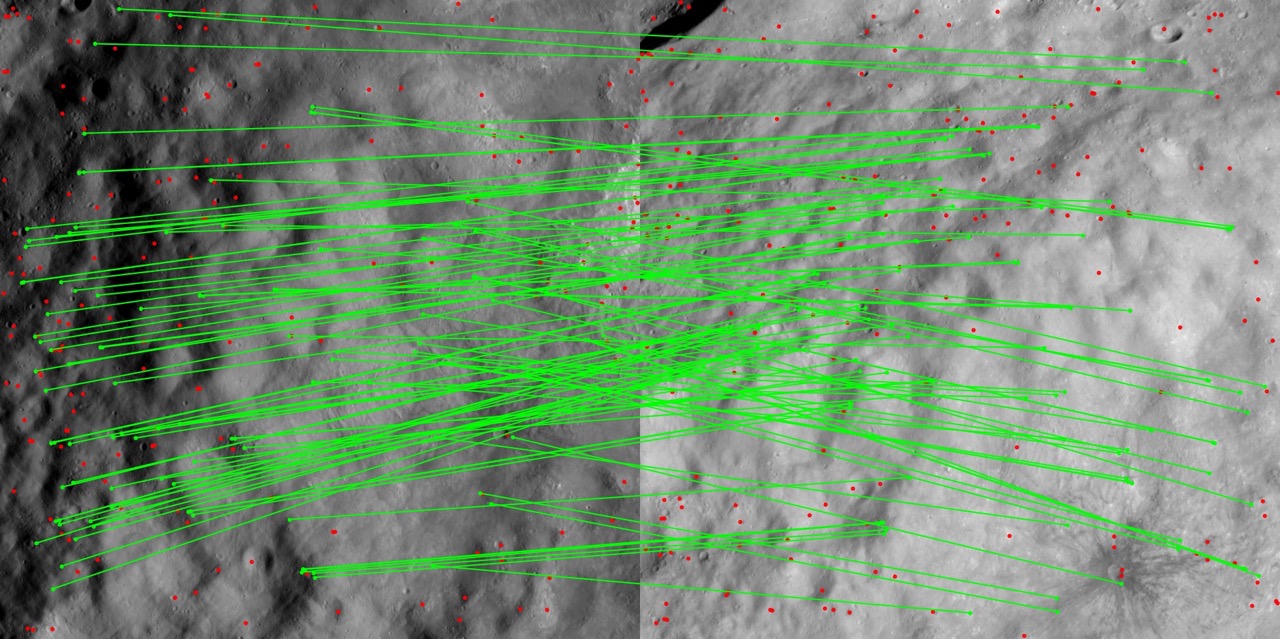}
  \end{subfigure}\\
  \vspace{0.5pt}
  \begin{subfigure}[t]{\linewidth}
    \includegraphics[width=.935\linewidth]{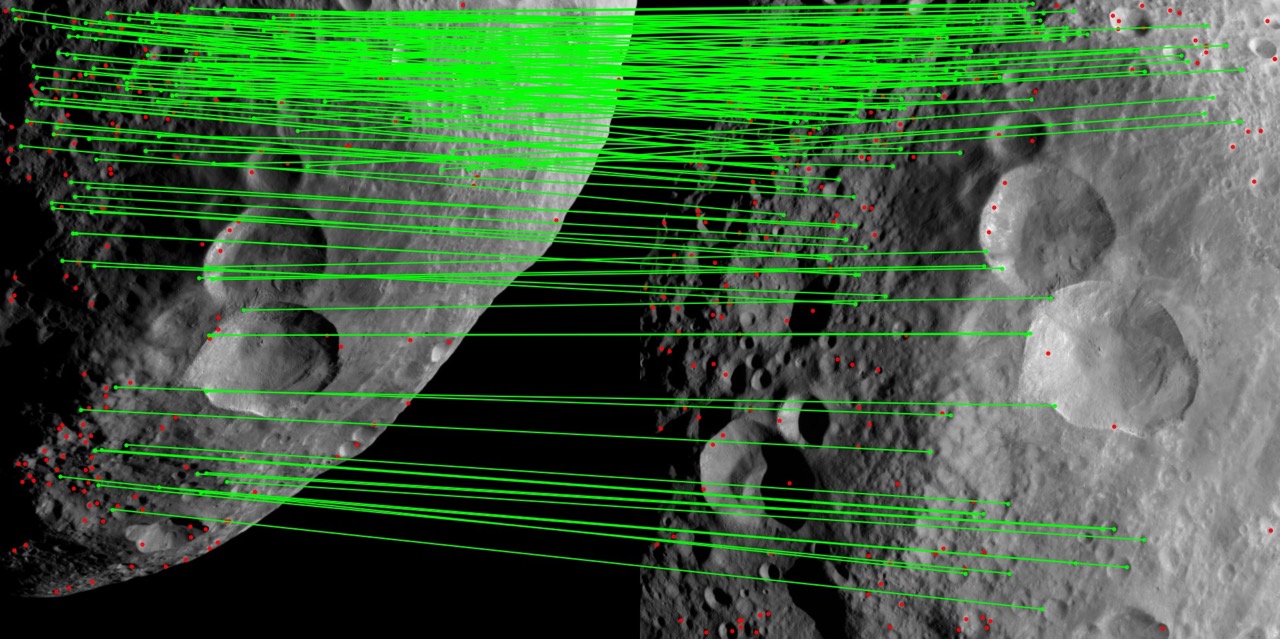}
  \end{subfigure}\\
  \vspace{4pt}
  \begin{subfigure}[t]{\linewidth}
    \includegraphics[width=.935\linewidth]{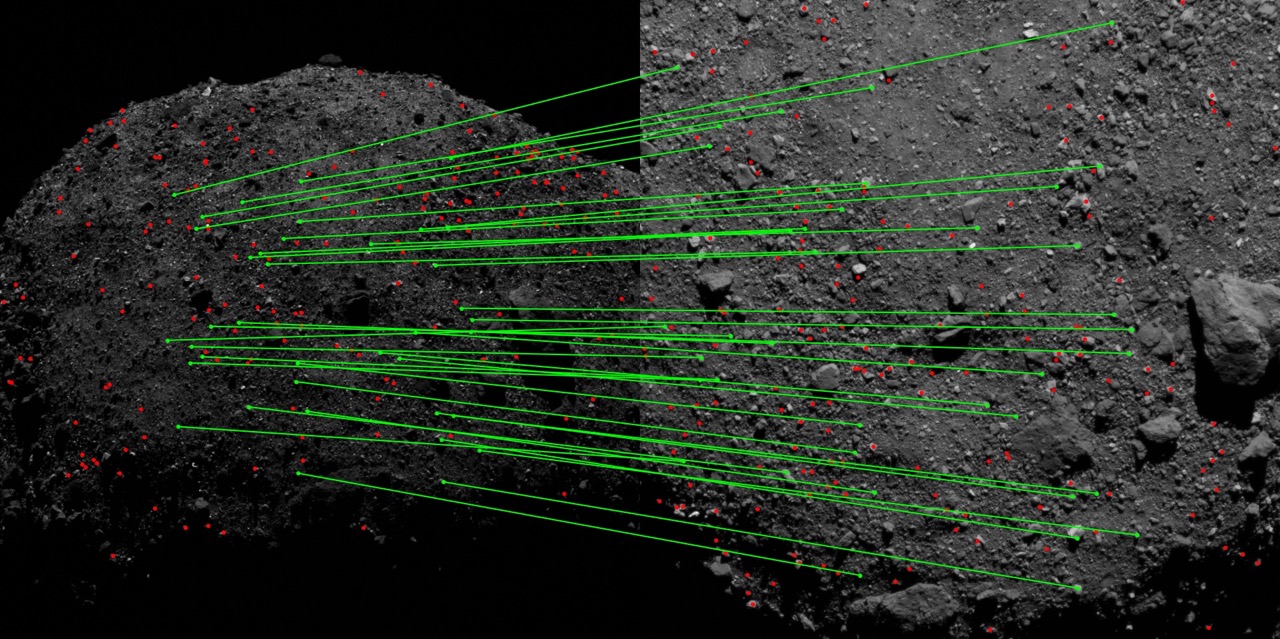}
  \end{subfigure}\\
  \begin{subfigure}[t]{\linewidth}
    \includegraphics[width=.935\linewidth]{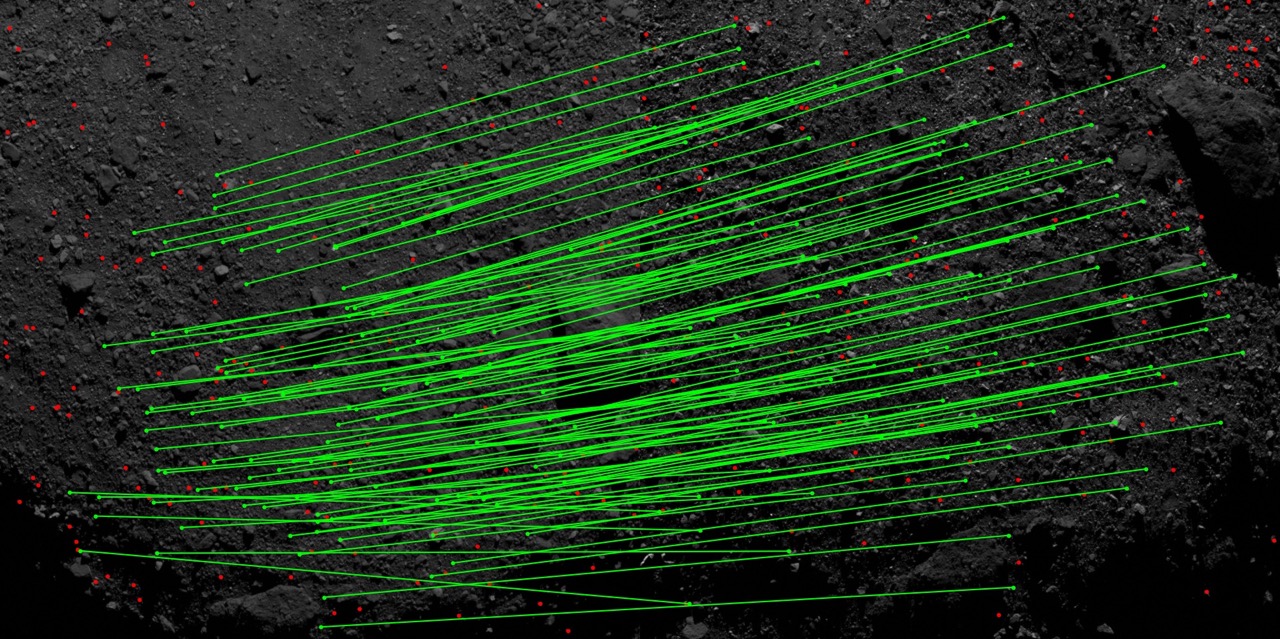}
  \end{subfigure}\\
  \vspace{0.5pt}
  \begin{subfigure}[t]{\linewidth}
    \includegraphics[width=.935\linewidth]{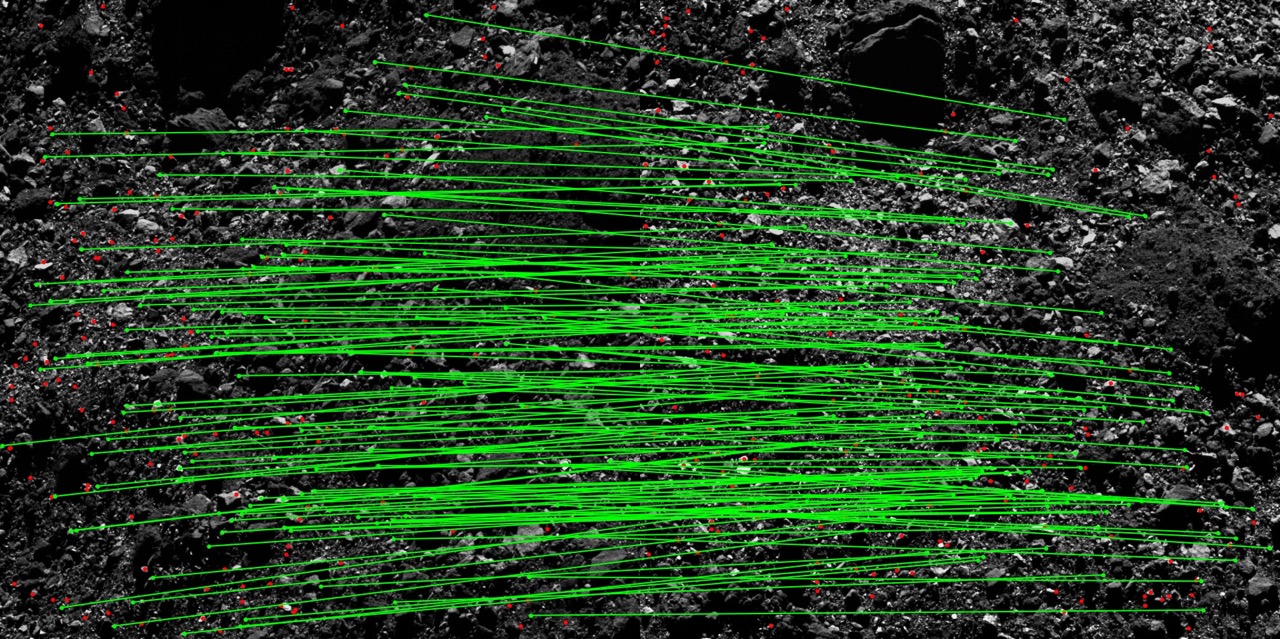}
  \end{subfigure}\\
  \vspace{4pt}
  \begin{subfigure}[t]{\linewidth}
    \includegraphics[width=.935\linewidth]{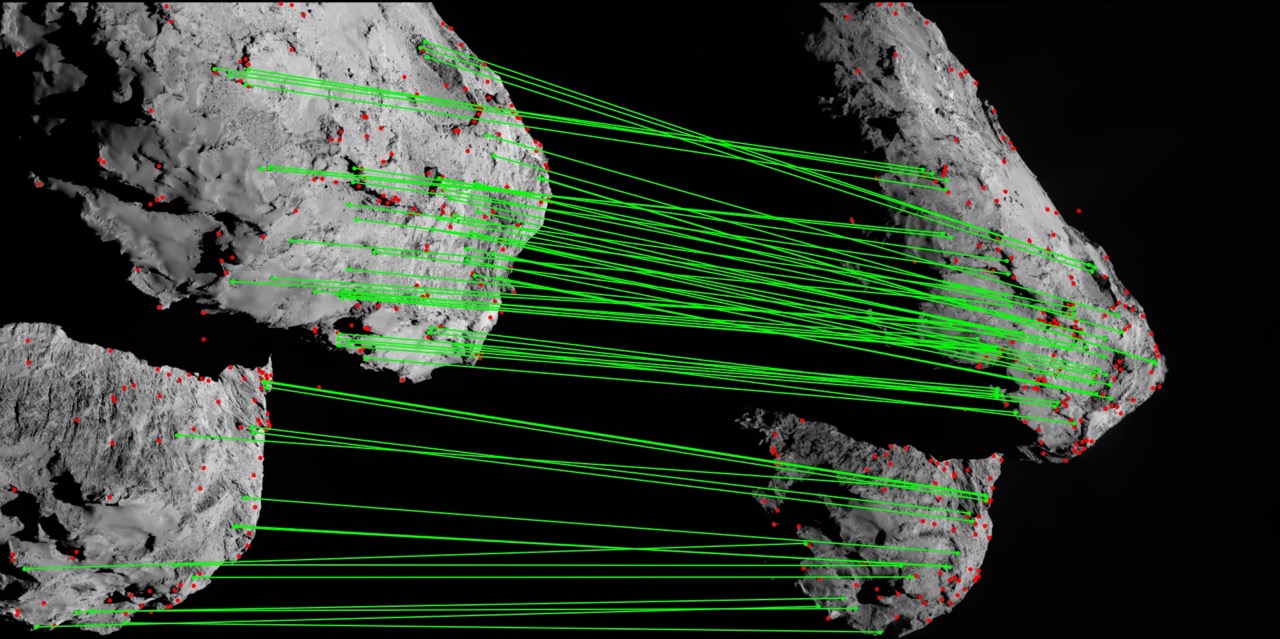}
  \end{subfigure}\\
  \begin{subfigure}[t]{\linewidth}
    \includegraphics[width=.935\linewidth]{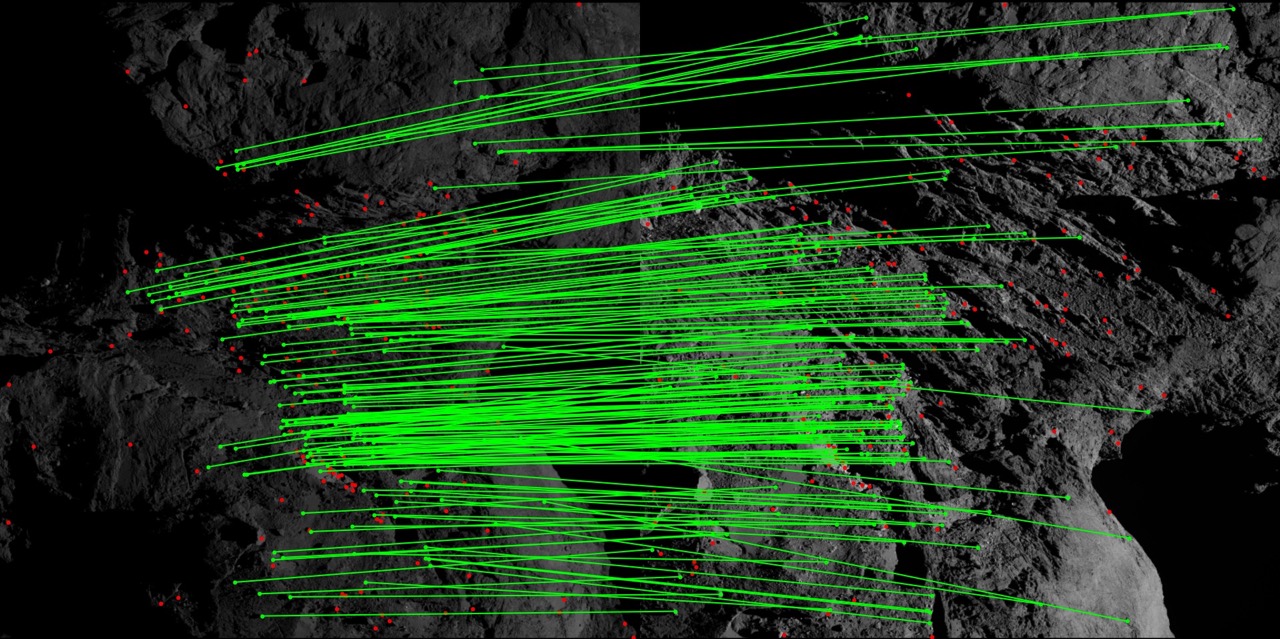}
  \end{subfigure}\\
    \vspace{0.5pt}
  \begin{subfigure}[t]{\linewidth}
    \includegraphics[width=.935\linewidth]{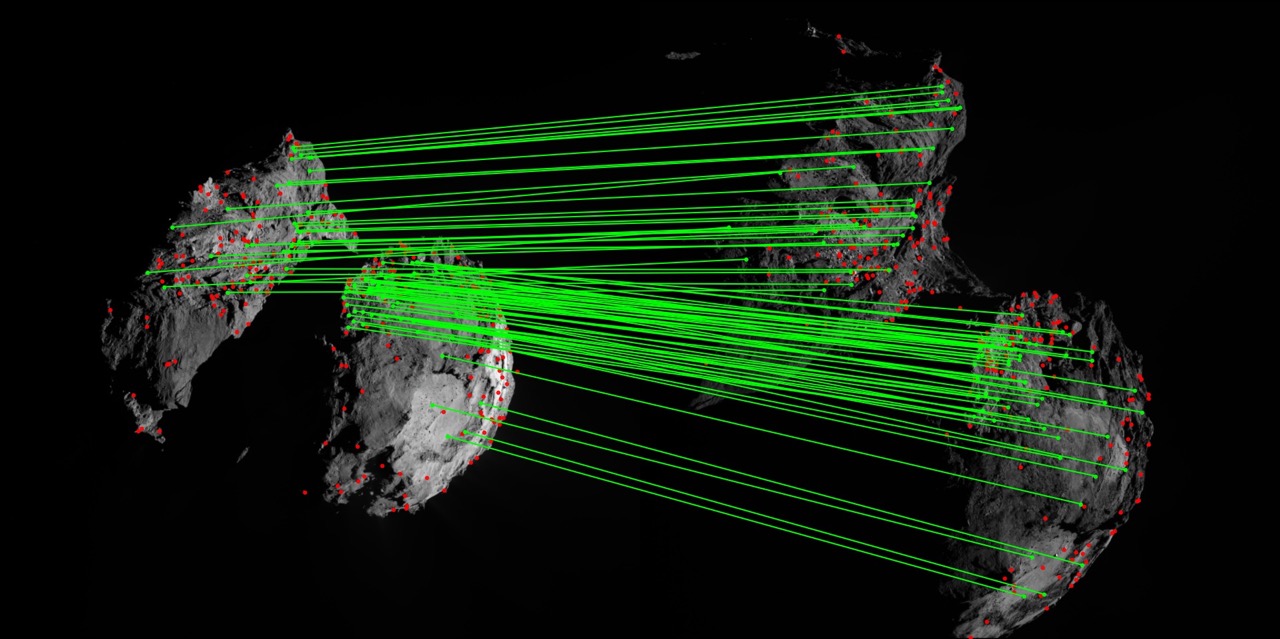}
  \end{subfigure}%
  \caption*{\footnotesize{DoG + DidymosNet}}
\end{subfigure}
\hspace{-5.5pt}
\begin{subfigure}[t]{0.20\linewidth}
  \centering
  \begin{subfigure}[t]{\linewidth}
    \includegraphics[width=.935\linewidth]{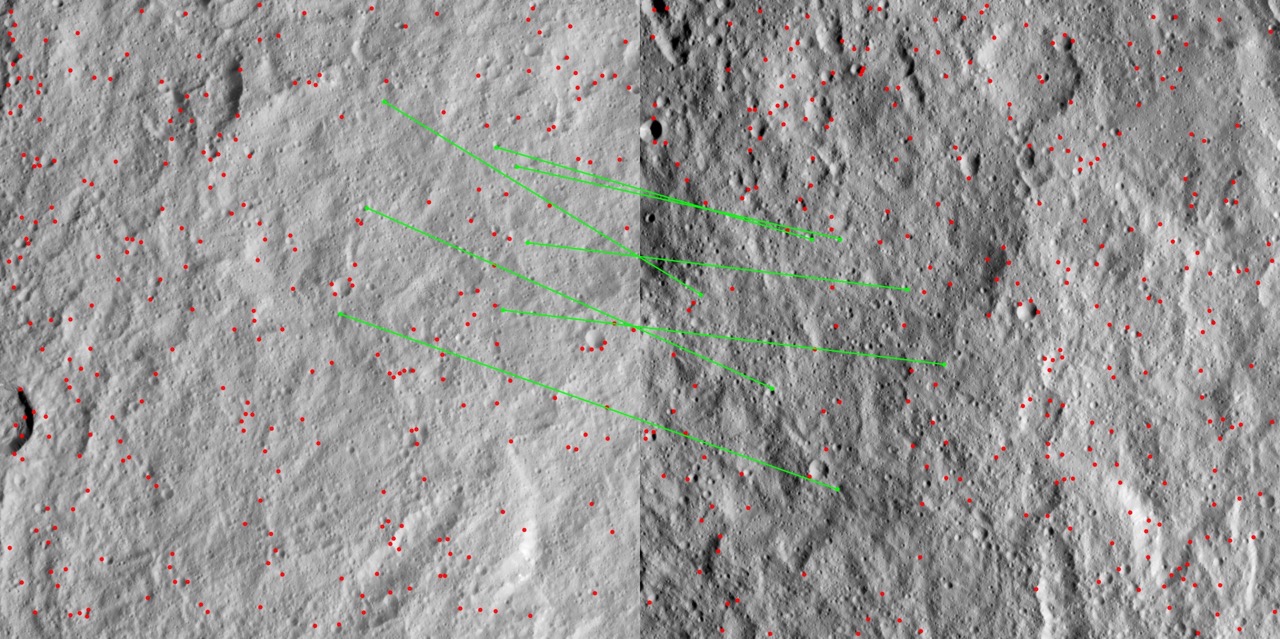}
  \end{subfigure}\\
  \vspace{0.5pt}
  \begin{subfigure}[t]{\linewidth}
    \includegraphics[width=.935\linewidth]{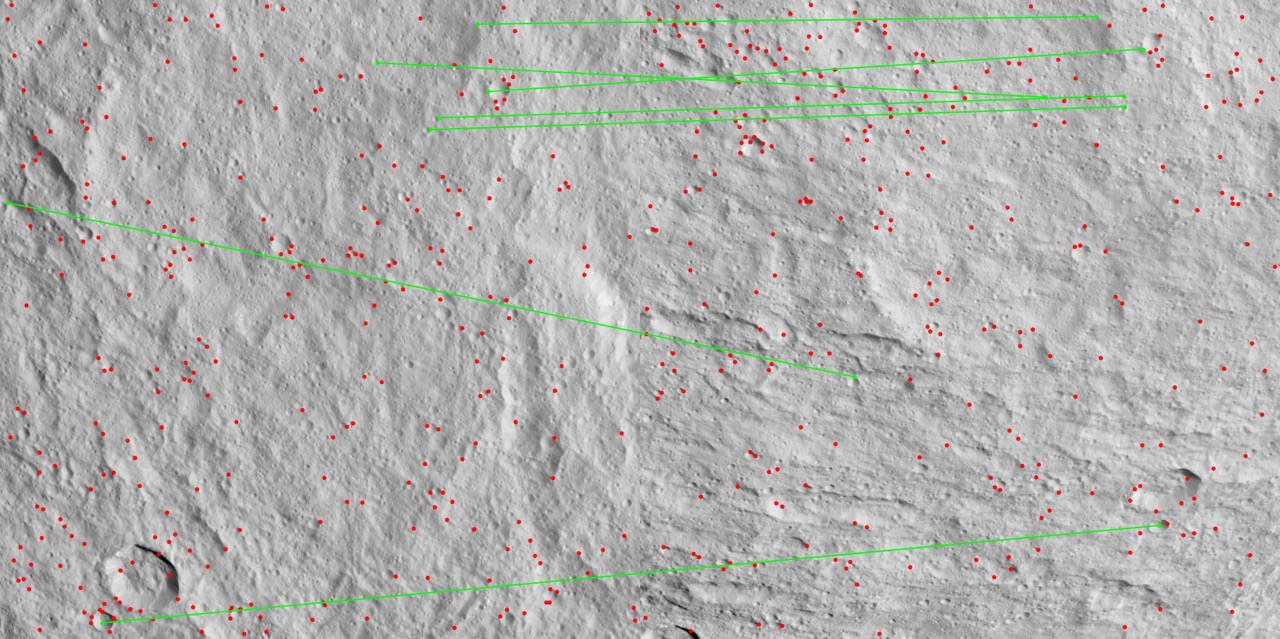}
  \end{subfigure}\\
  \vspace{4pt}
  \begin{subfigure}[t]{\linewidth}
    \includegraphics[width=.935\linewidth]{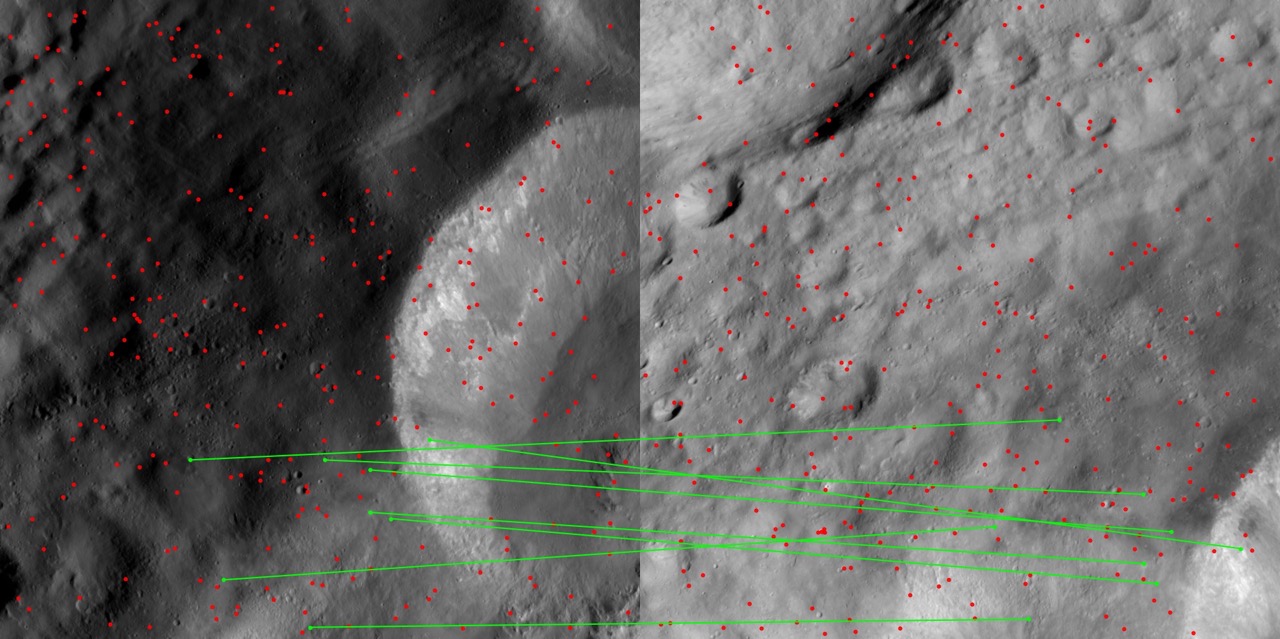}
  \end{subfigure}\\
  \begin{subfigure}[t]{\linewidth}
    \includegraphics[width=.935\linewidth]{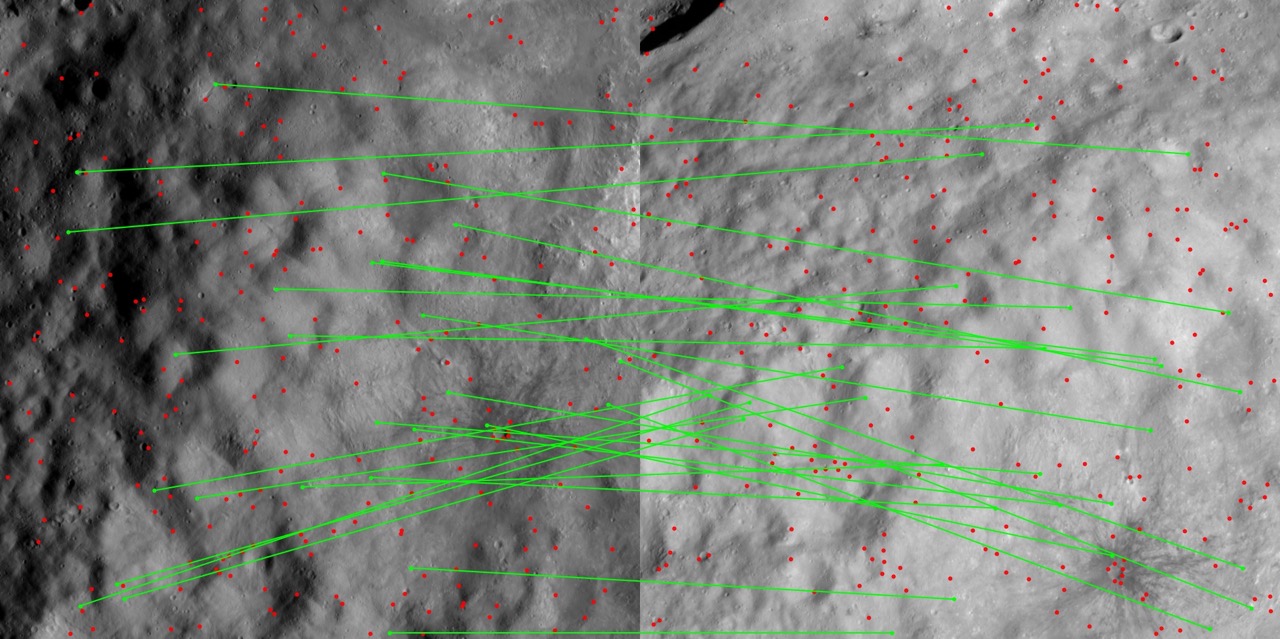}
  \end{subfigure}\\
  \vspace{0.5pt}
  \begin{subfigure}[t]{\linewidth}
    \includegraphics[width=.935\linewidth]{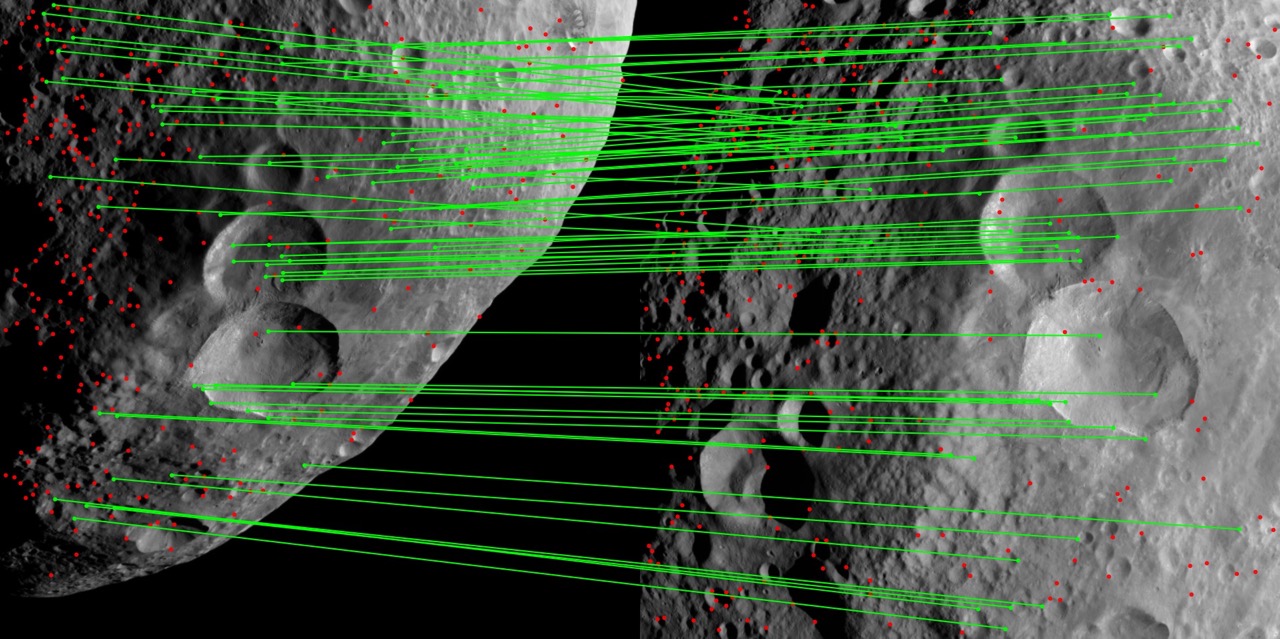}
  \end{subfigure}\\
  \vspace{4pt}
  \begin{subfigure}[t]{\linewidth}
    \includegraphics[width=.935\linewidth]{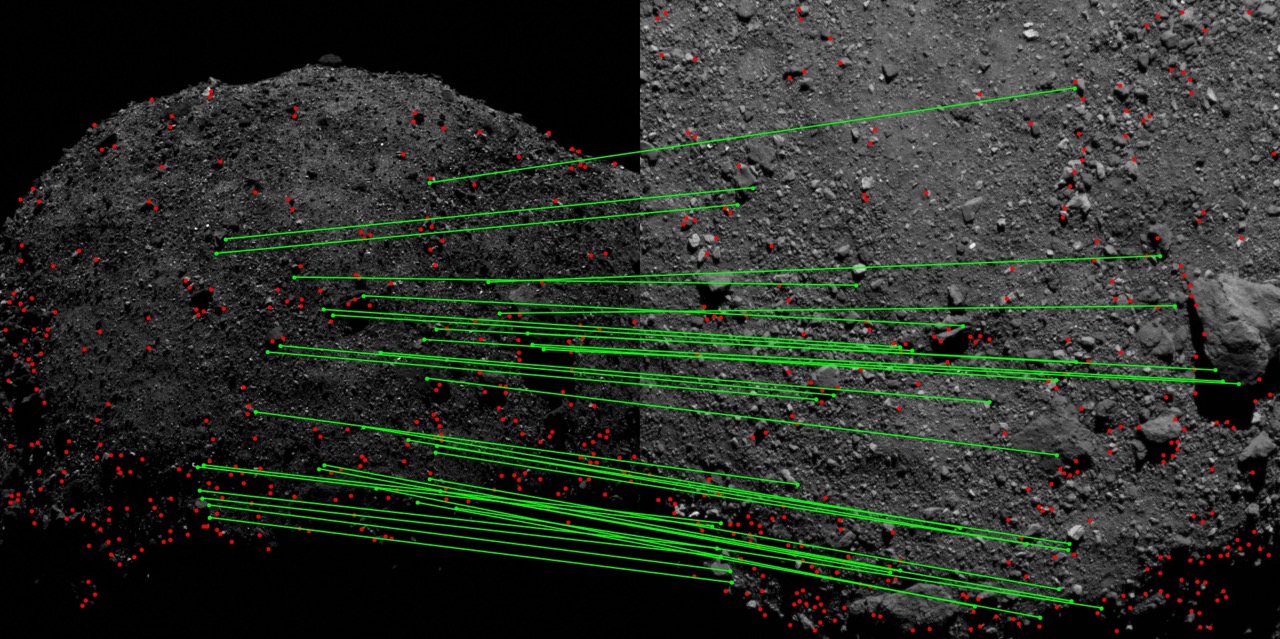}
  \end{subfigure}\\
  \begin{subfigure}[t]{\linewidth}
    \includegraphics[width=.935\linewidth]{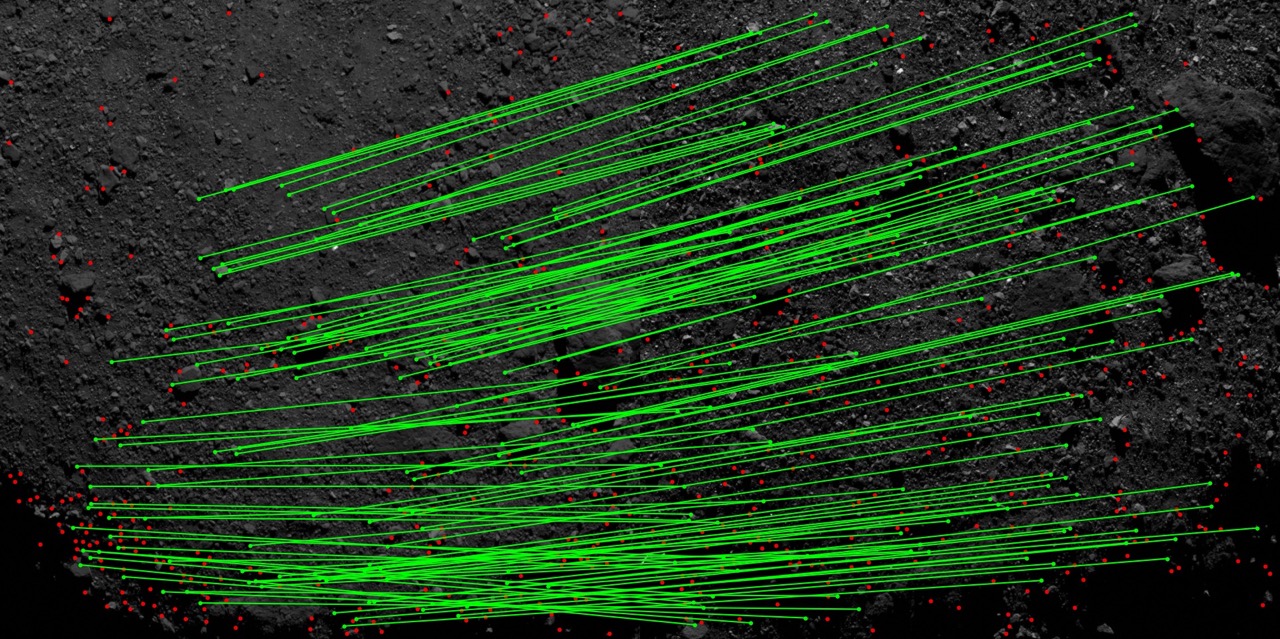}
  \end{subfigure}\\
  \vspace{0.5pt}
  \begin{subfigure}[t]{\linewidth}
    \includegraphics[width=.935\linewidth]{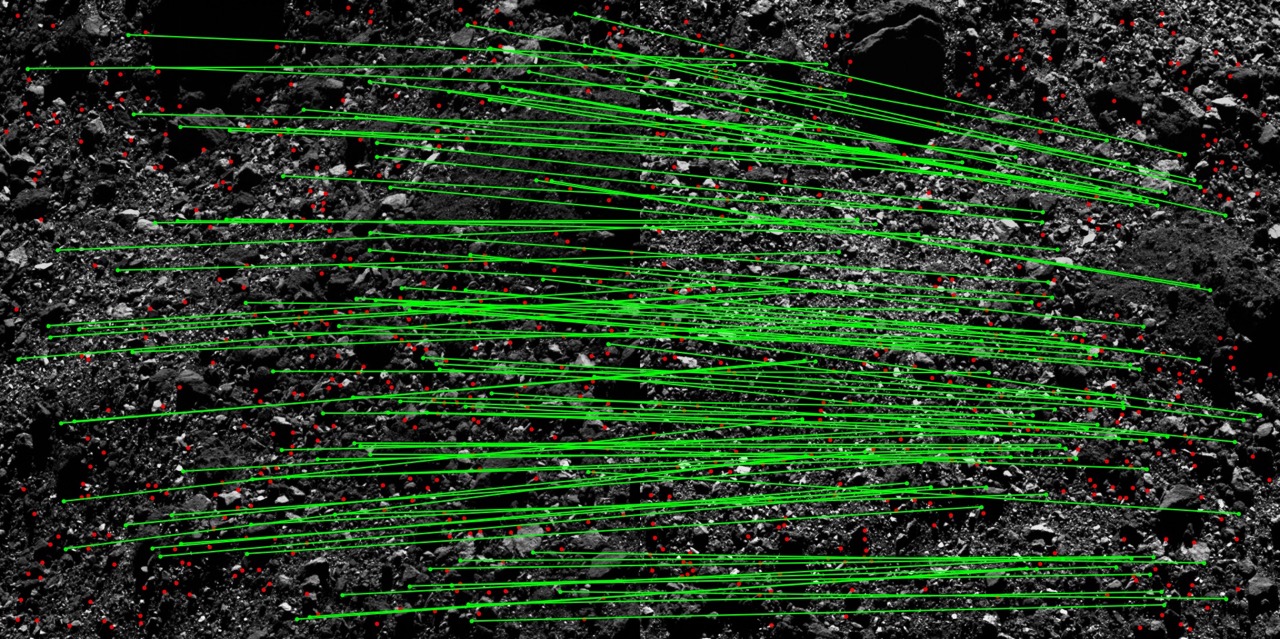}
  \end{subfigure}\\
  \vspace{4pt}
  \begin{subfigure}[t]{\linewidth}
    \includegraphics[width=.935\linewidth]{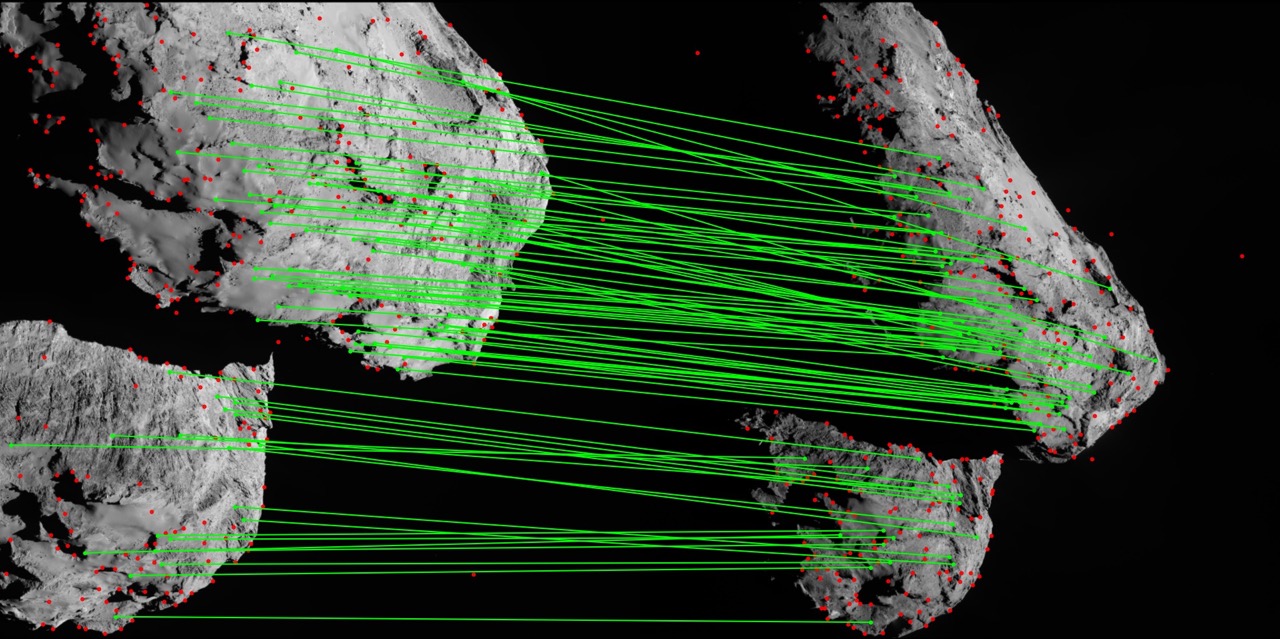}
  \end{subfigure}\\
  \begin{subfigure}[t]{\linewidth}
    \includegraphics[width=.935\linewidth]{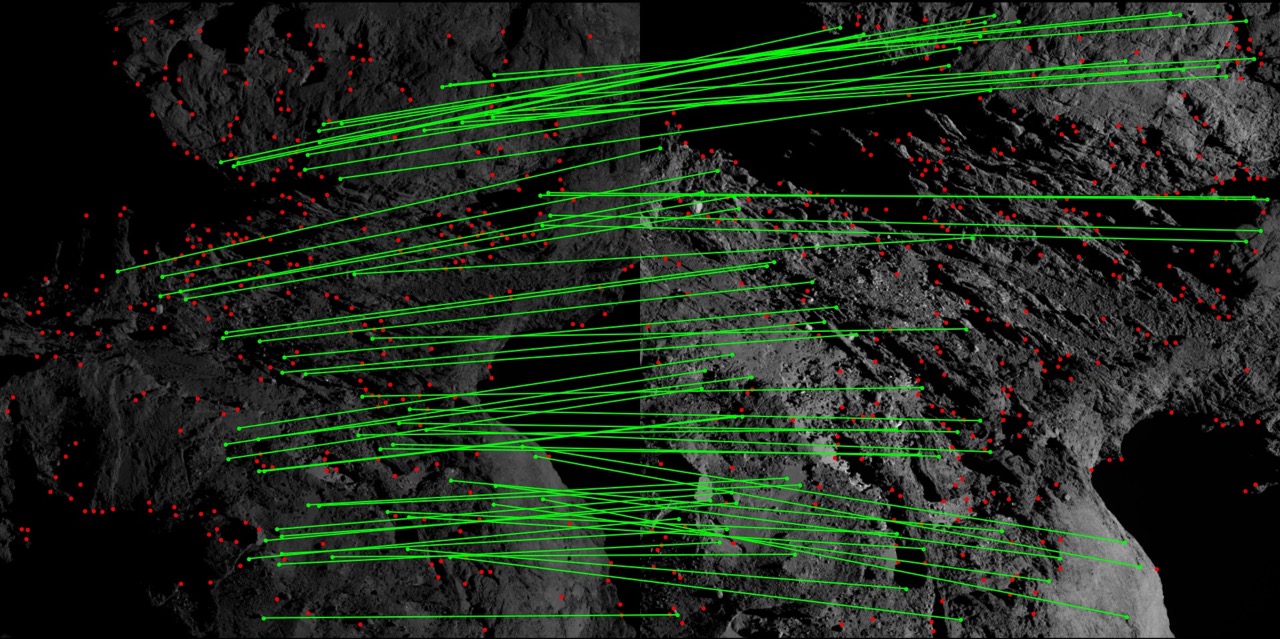}
  \end{subfigure}\\
    \vspace{0.5pt}
  \begin{subfigure}[t]{\linewidth}
    \includegraphics[width=.935\linewidth]{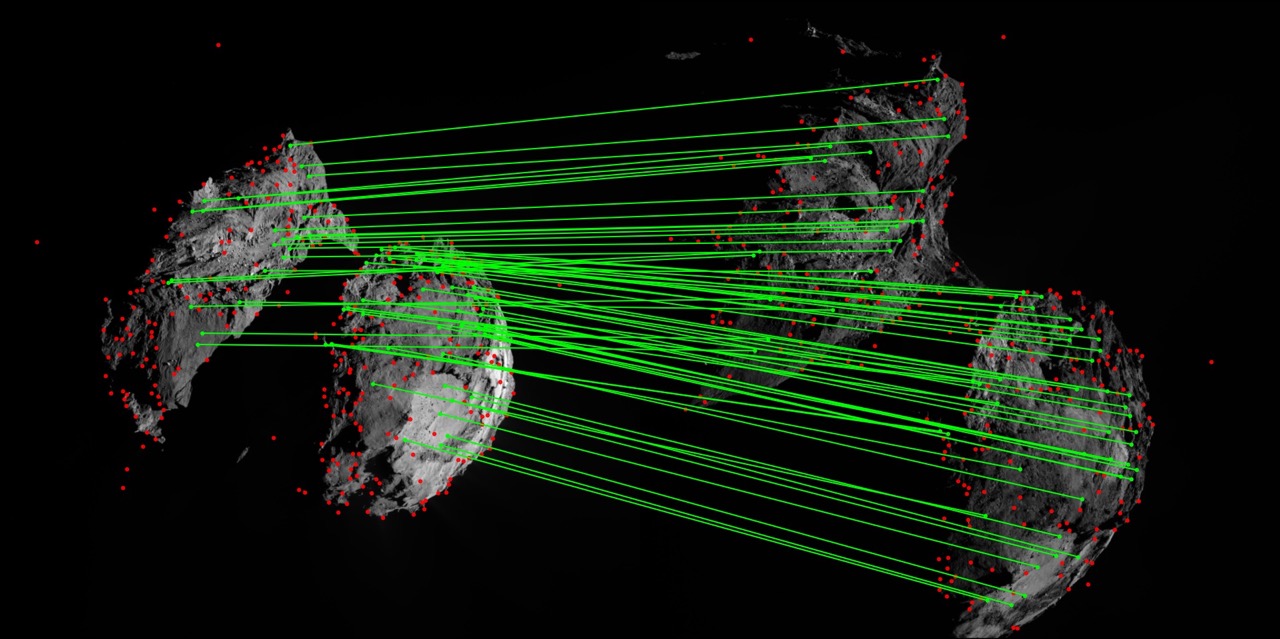}
  \end{subfigure}%
  \caption*{\footnotesize{SuperPoint}}
\end{subfigure}
\end{minipage}

%% file: Figures/eval-qual-compare-bin.tex
\begin{minipage}{1.23\textwidth}
\hspace{55pt}
\begin{subfigure}[t]{0.20\linewidth}
  \centering
  \begin{subfigure}[t]{\linewidth}
    \includegraphics[width=.935\linewidth,right]{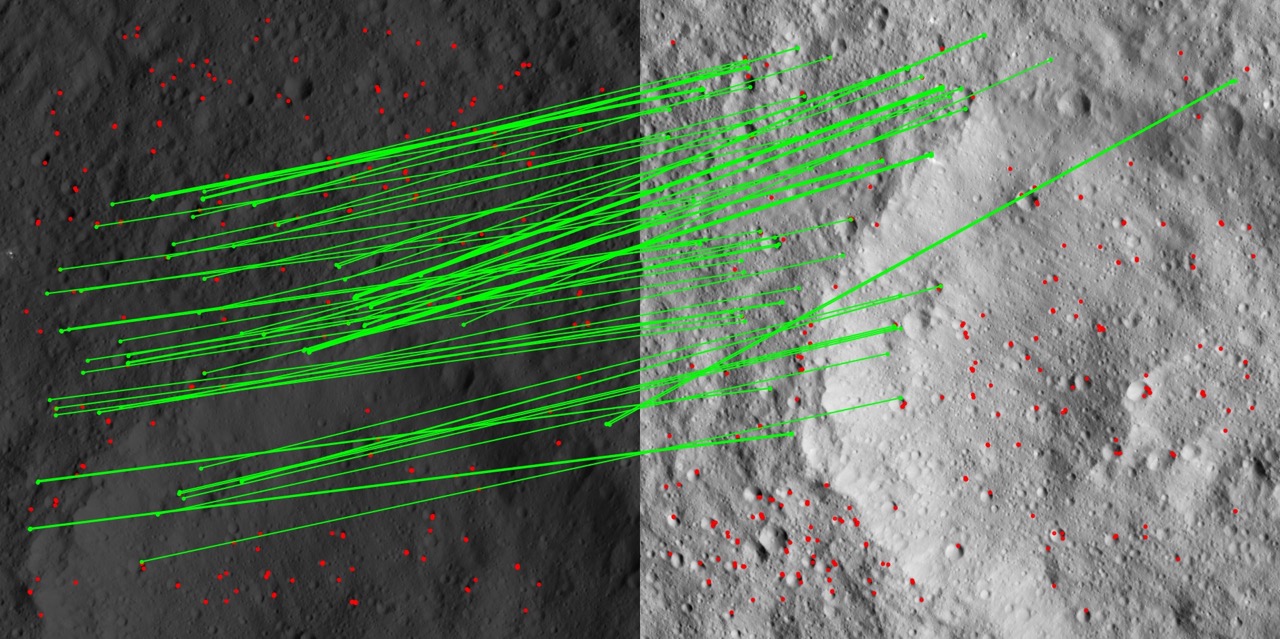}
  \end{subfigure}\\
  \vspace{0.5pt}
  \begin{subfigure}[t]{\linewidth}
    \begin{tabular}{c c}
         \hspace{-11pt}
         \rotatebox[origin=l]{90}{\tiny{(a) \texttt{Ceres}}}
         \hspace{-17.5pt}
         &
         \includegraphics[width=.935\linewidth,center]{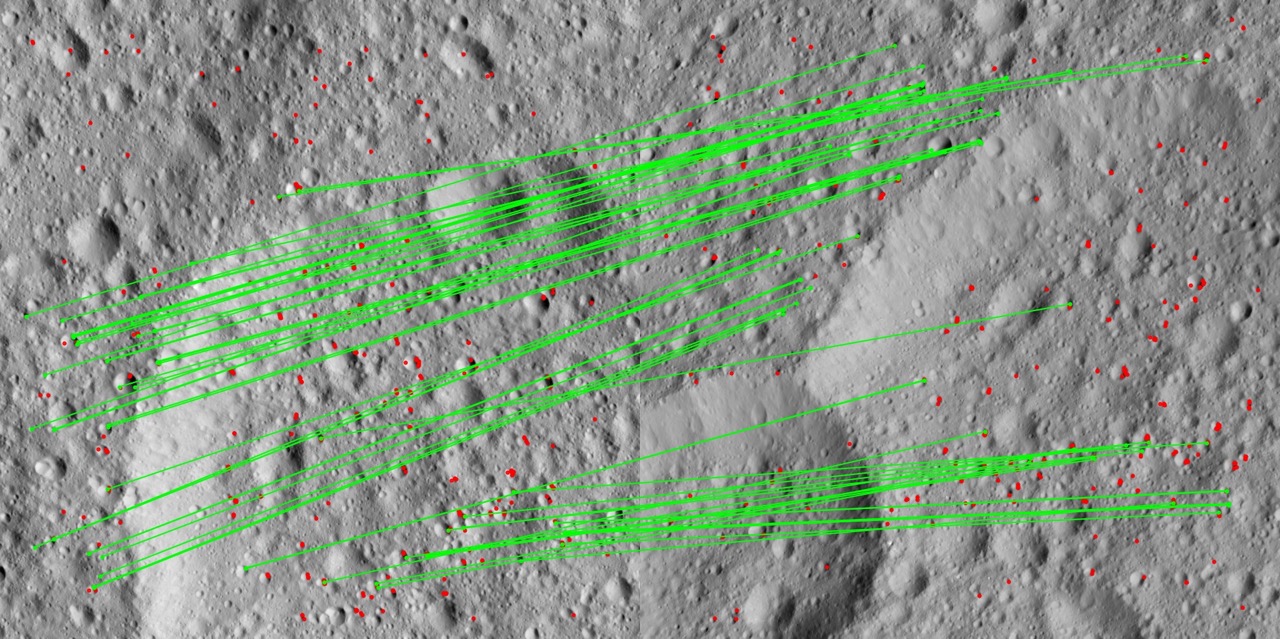}
    \end{tabular}
    \refstepcounter{subfigure}\label{fig:eval-ceres-bin}
  \end{subfigure}\\
  \vspace{-1pt}
  \begin{subfigure}[t]{\linewidth}
    \includegraphics[width=.935\linewidth,right]{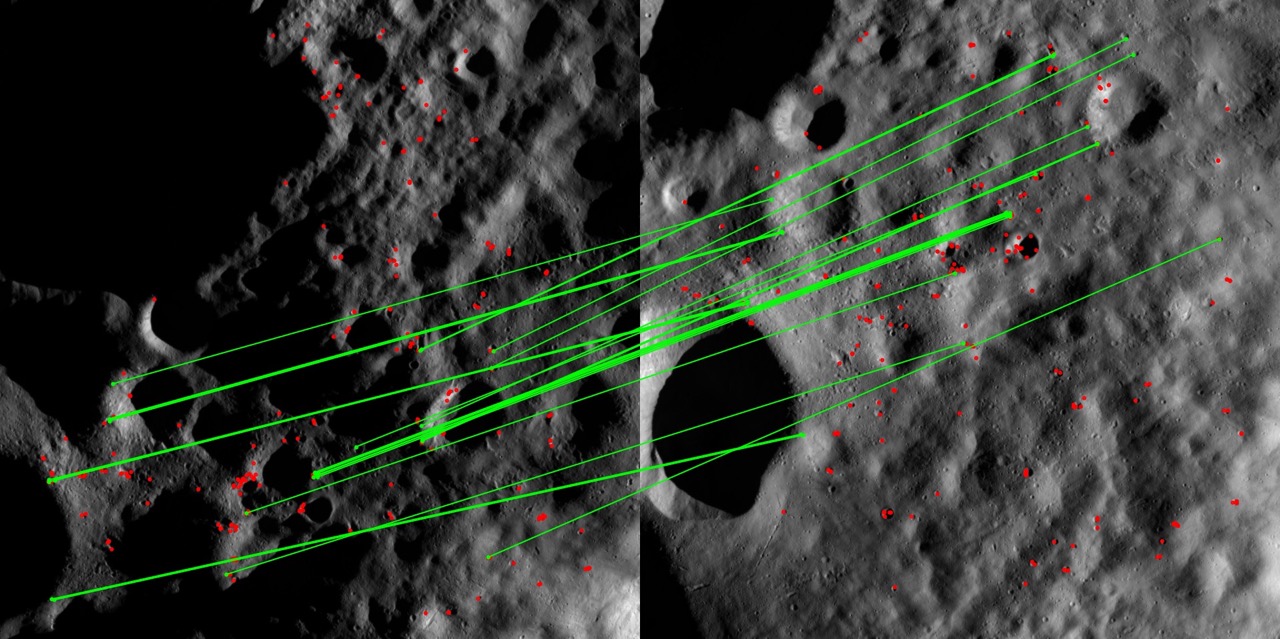}
  \end{subfigure}\\
  \begin{subfigure}[t]{\linewidth}
    \includegraphics[width=.935\linewidth,right]{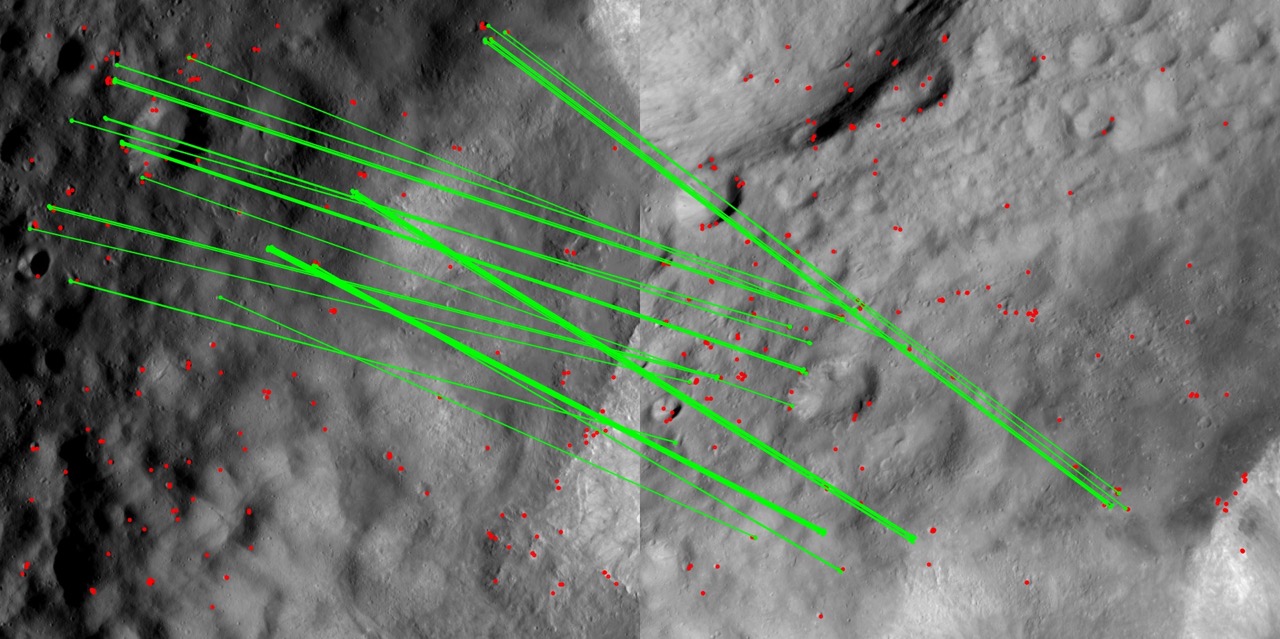}
  \end{subfigure}\\
  \vspace{0.5pt}
  \begin{subfigure}[t]{\linewidth}
    \begin{tabular}{c c}
         \hspace{-11pt}
         \rotatebox[origin=l]{90}{\tiny{(b) \texttt{Vesta}}}
         \hspace{-20.25pt}
         &
         \includegraphics[width=.935\linewidth,center]{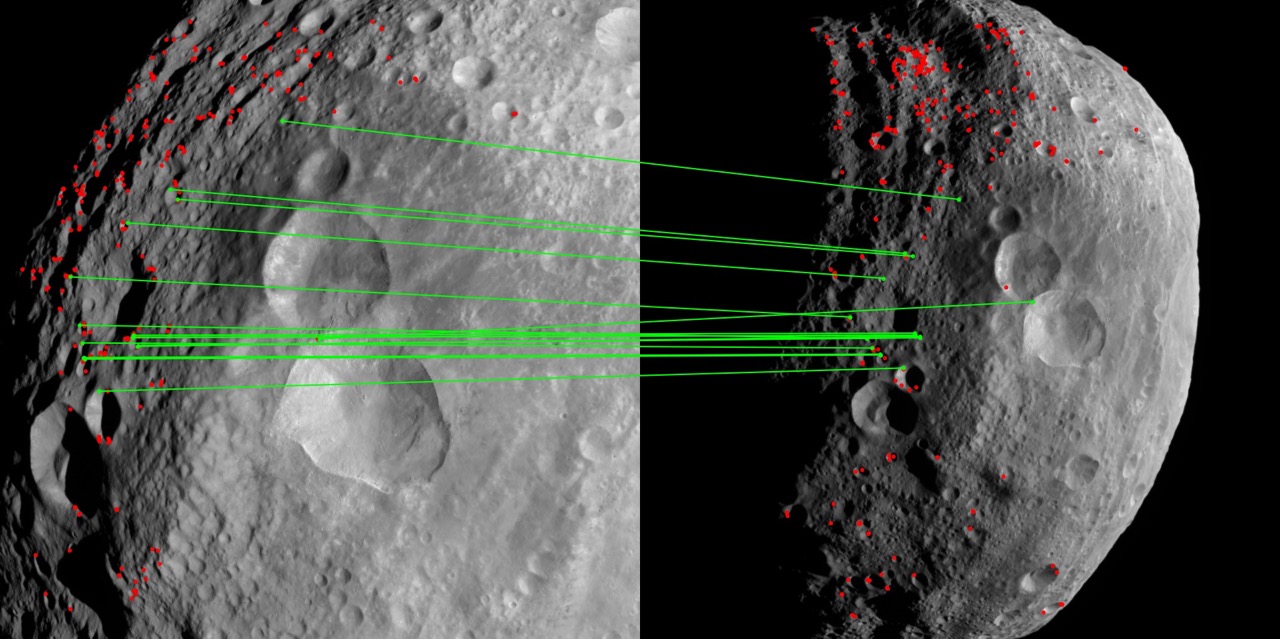}
    \end{tabular}
    \refstepcounter{subfigure}\label{fig:eval-vesta-bin}
  \end{subfigure}\\
  \vspace{-1pt}
  \begin{subfigure}[t]{\linewidth}
    \includegraphics[width=.935\linewidth,right]{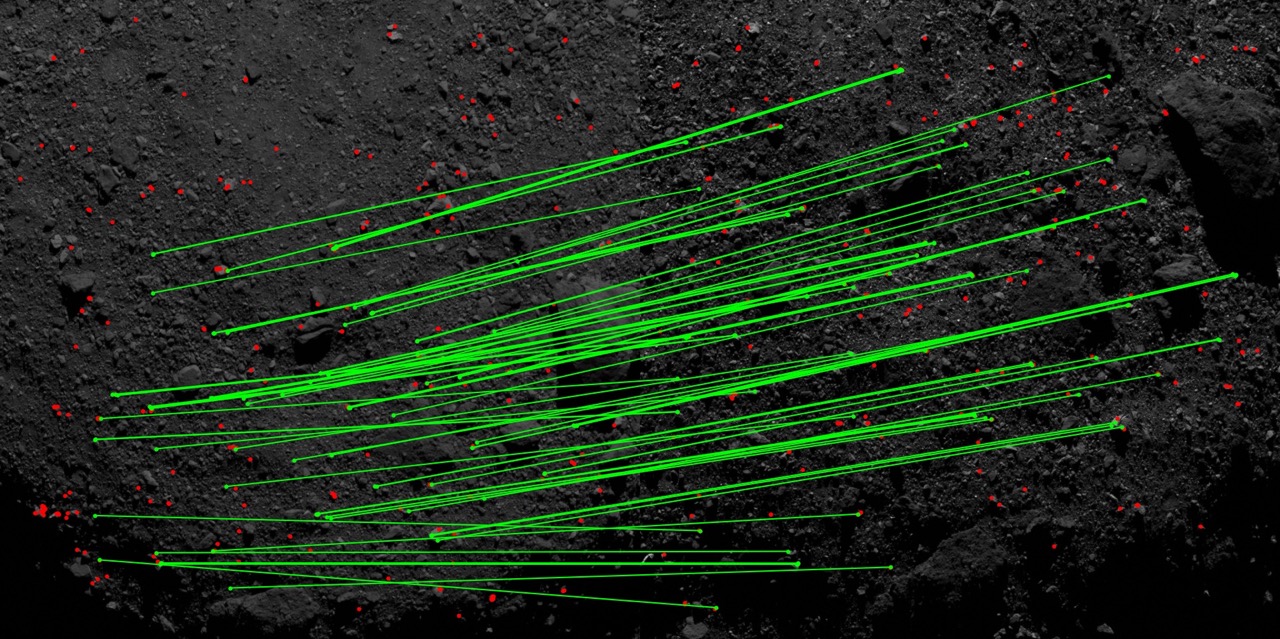}
  \end{subfigure}\\
  \begin{subfigure}[t]{\linewidth}
    \includegraphics[width=.935\linewidth,right]{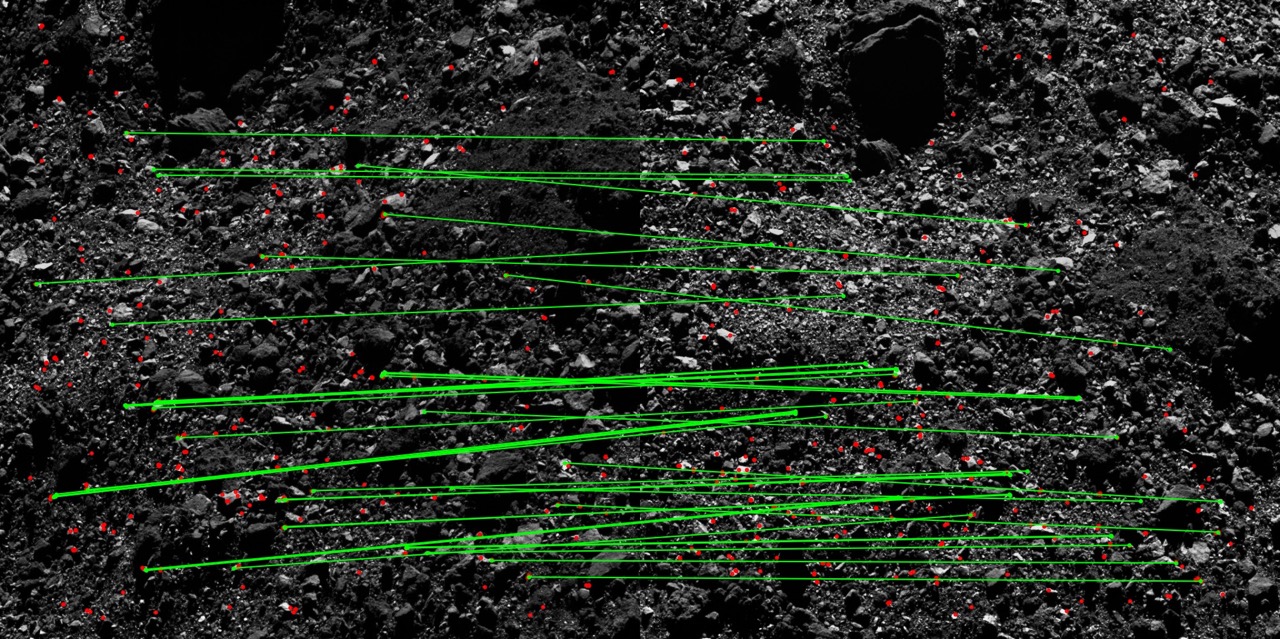}
  \end{subfigure}\\
  \vspace{0.5pt}
  \begin{subfigure}[t]{\linewidth}
    \begin{tabular}{c c}
         \hspace{-11pt}
         \rotatebox[origin=l]{90}{\tiny{(c) \texttt{Bennu}}}
         \hspace{-16.5pt}
         &
         \includegraphics[width=.935\linewidth]{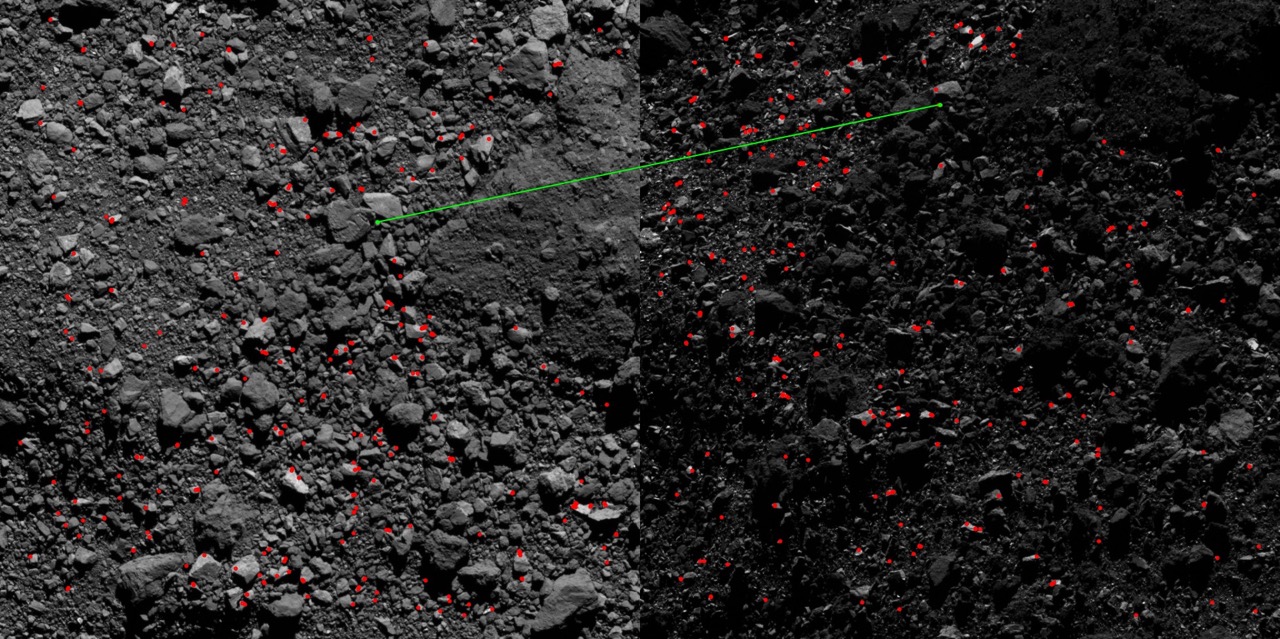}
    \end{tabular}
    \refstepcounter{subfigure}\label{fig:eval-bennu-bin}
  \end{subfigure}\\
  \vspace{-1pt}
  \begin{subfigure}[t]{\linewidth}
    \includegraphics[width=.935\linewidth,right]{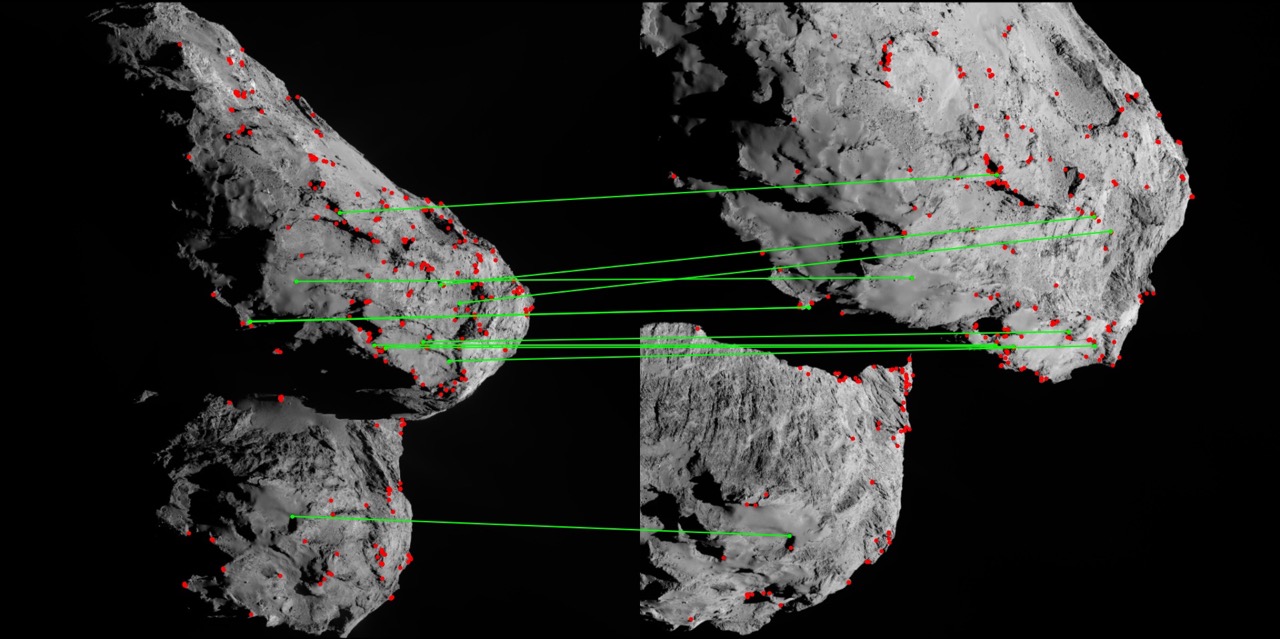}
  \end{subfigure}\\
  \begin{subfigure}[t]{\linewidth}
    \includegraphics[width=.935\linewidth,right]{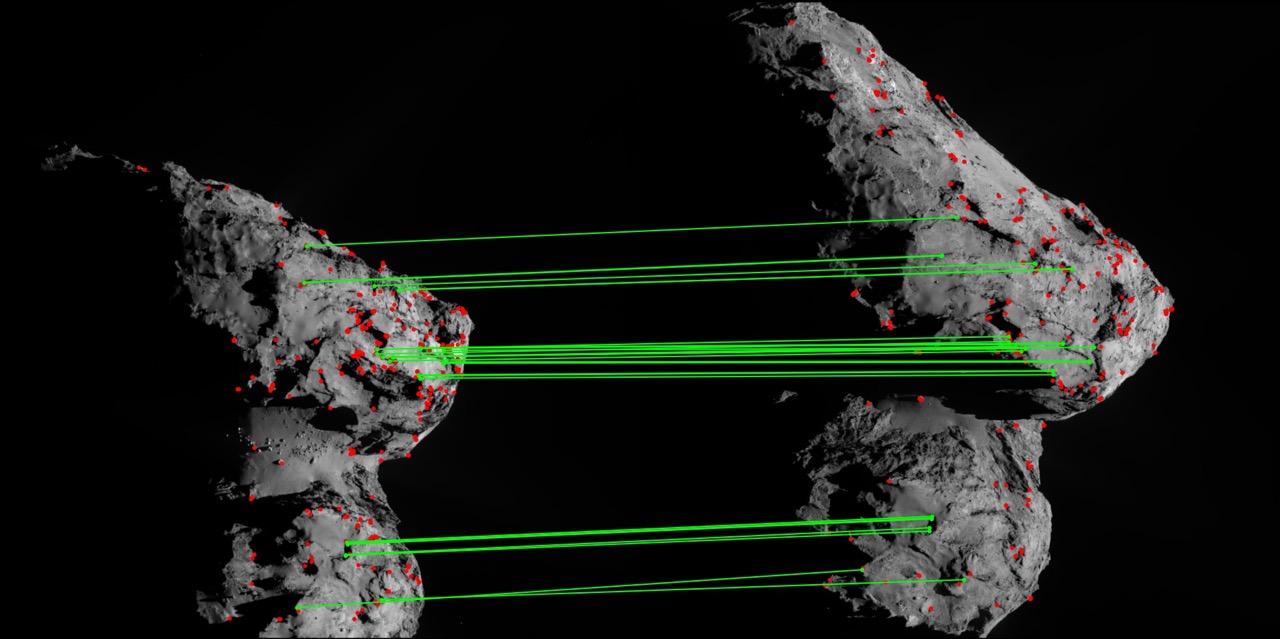}
  \end{subfigure}\\
  \vspace{0.5pt}
  \begin{subfigure}[t]{\linewidth}
    \begin{tabular}{c c}
         \hspace{-11pt}
         \rotatebox[origin=l]{90}{\tiny{(d) \texttt{67P}}}
         \hspace{-16.5pt}
         &
         \includegraphics[width=.935\linewidth]{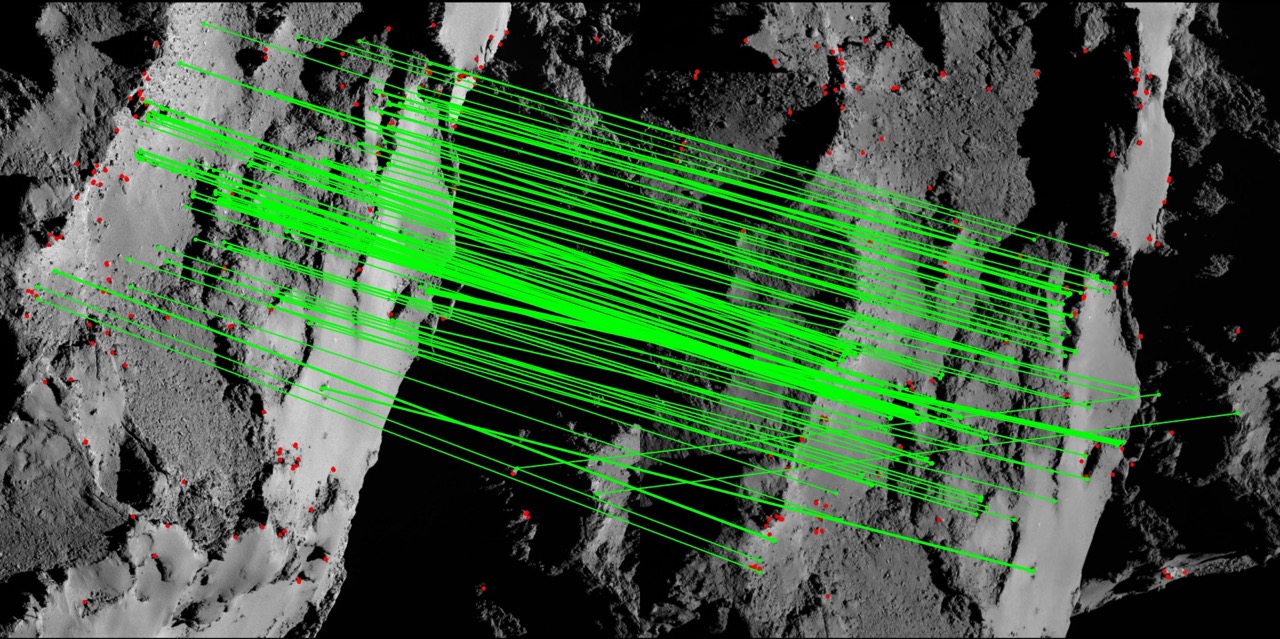}
    \end{tabular}
    \refstepcounter{subfigure}\label{fig:eval-chury-bin}
  \end{subfigure}
  \vspace{-6.5pt}
  \caption*{\footnotesize{ORB}}
\end{subfigure}
\hspace{1pt}
\begin{subfigure}[t]{0.20\linewidth}
  \centering
  \begin{subfigure}[t]{\linewidth}
    \includegraphics[width=.935\linewidth]{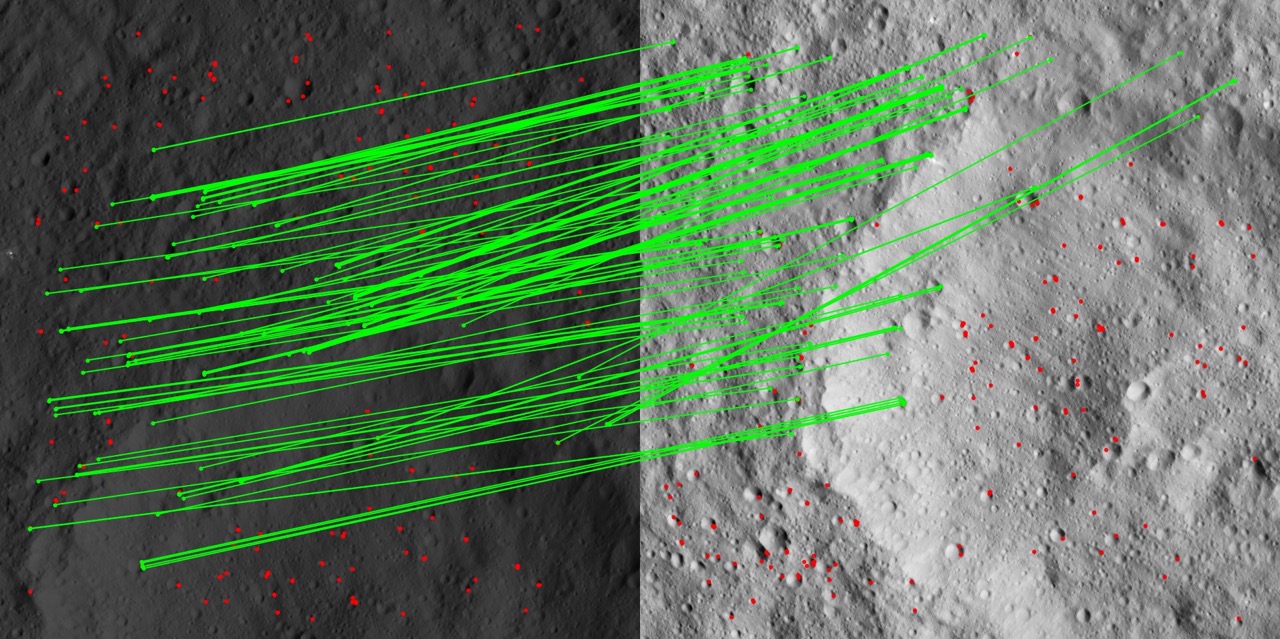}
  \end{subfigure}\\
  \vspace{0.5pt}
  \begin{subfigure}[t]{\linewidth}
    \includegraphics[width=.935\linewidth]{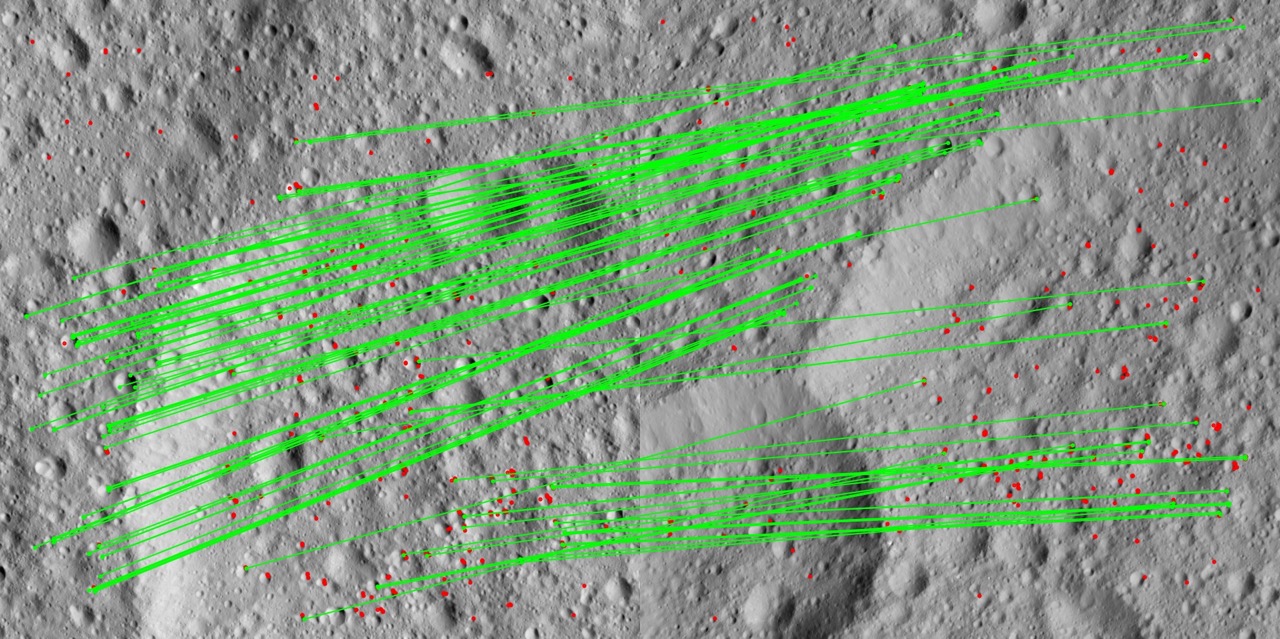}
  \end{subfigure}\\
  \vspace{4pt}
  \begin{subfigure}[t]{\linewidth}
    \includegraphics[width=.935\linewidth]{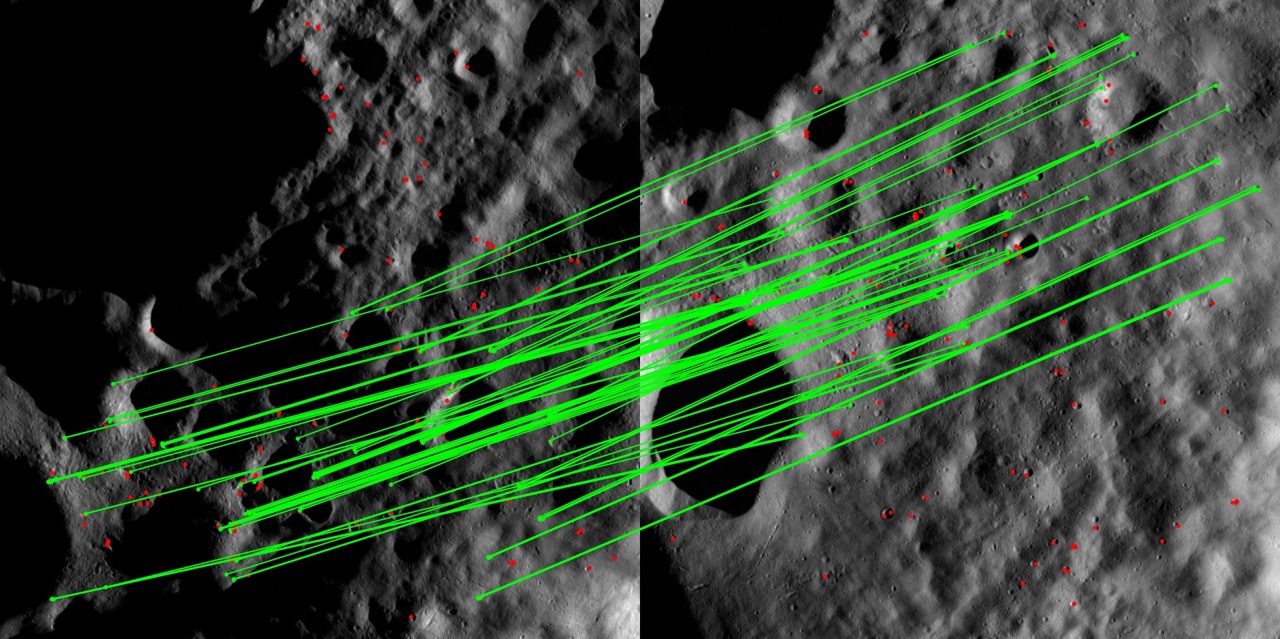}
  \end{subfigure}\\
  \begin{subfigure}[t]{\linewidth}
    \includegraphics[width=.935\linewidth]{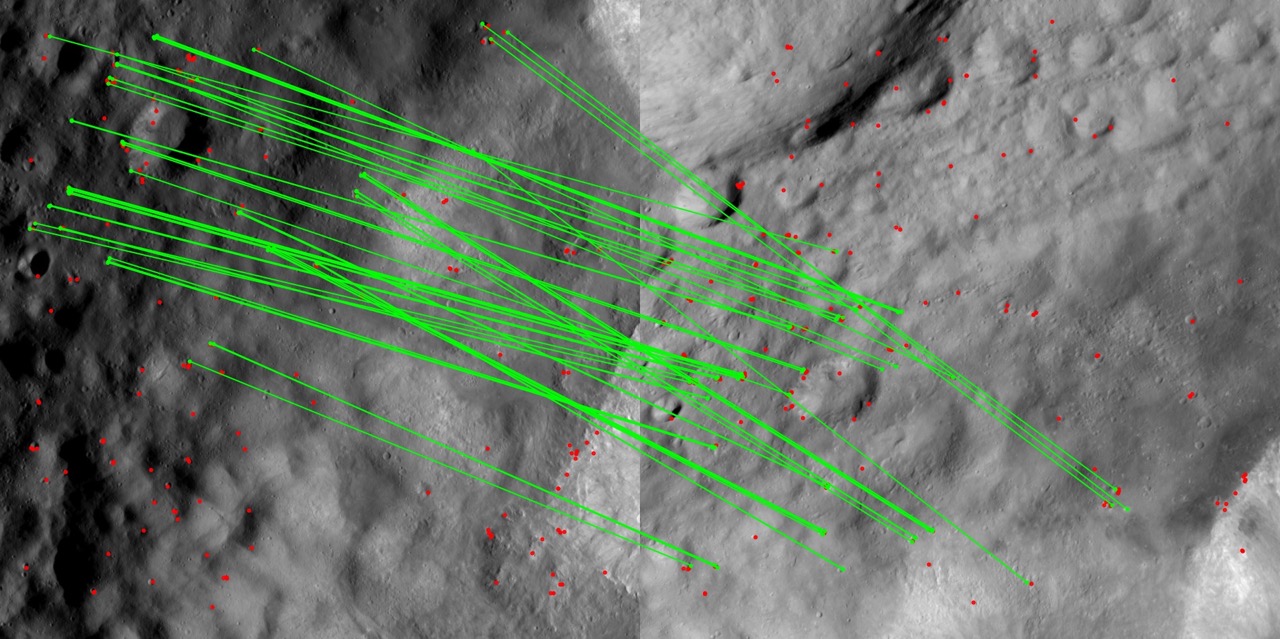}
  \end{subfigure}\\
  \vspace{0.5pt}
  \begin{subfigure}[t]{\linewidth}
    \includegraphics[width=.935\linewidth]{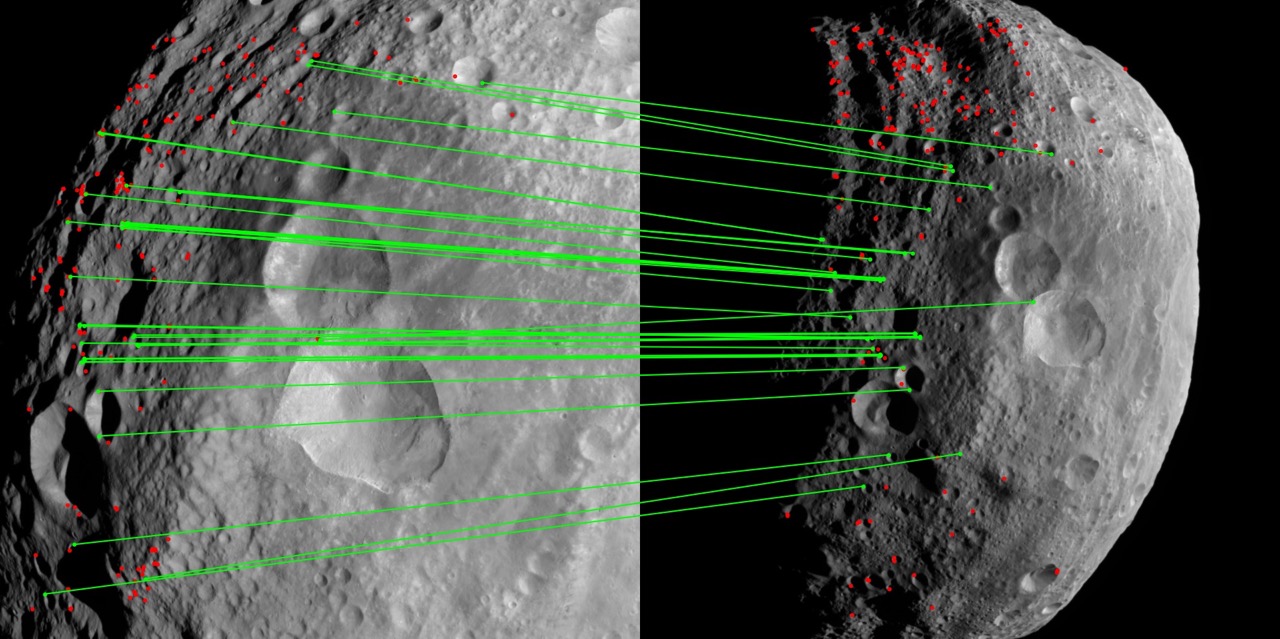}
  \end{subfigure}\\
  \vspace{4pt}
  \begin{subfigure}[t]{\linewidth}
    \includegraphics[width=.935\linewidth]{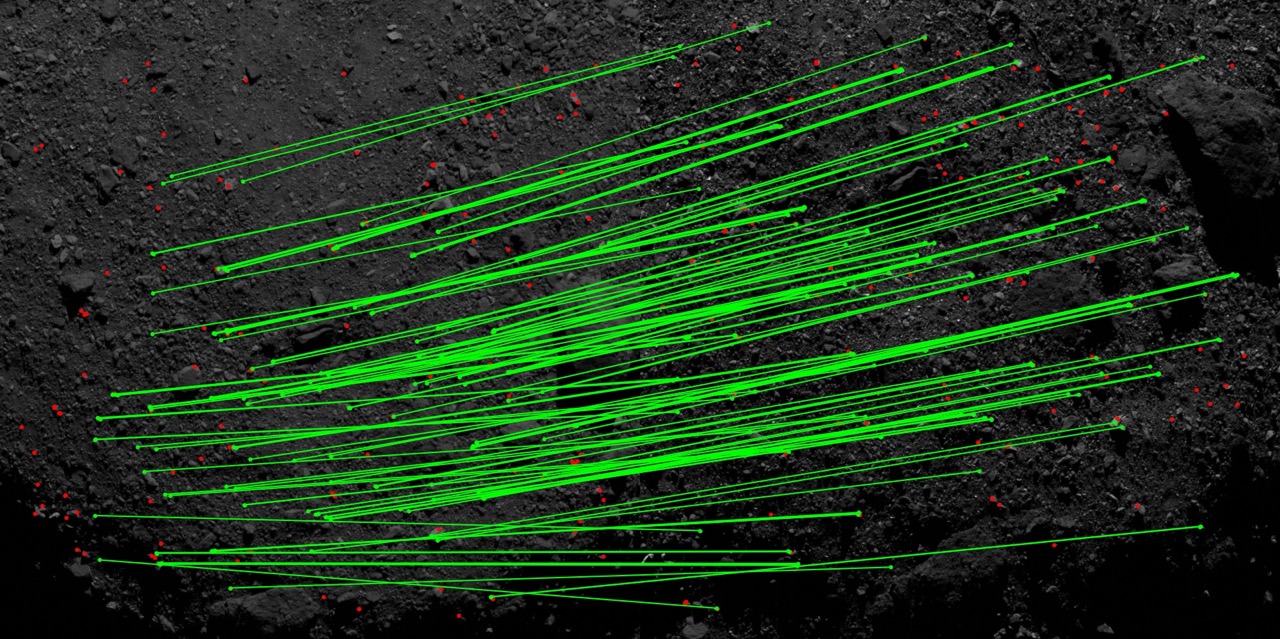}
  \end{subfigure}\\
  \begin{subfigure}[t]{\linewidth}
    \includegraphics[width=.935\linewidth]{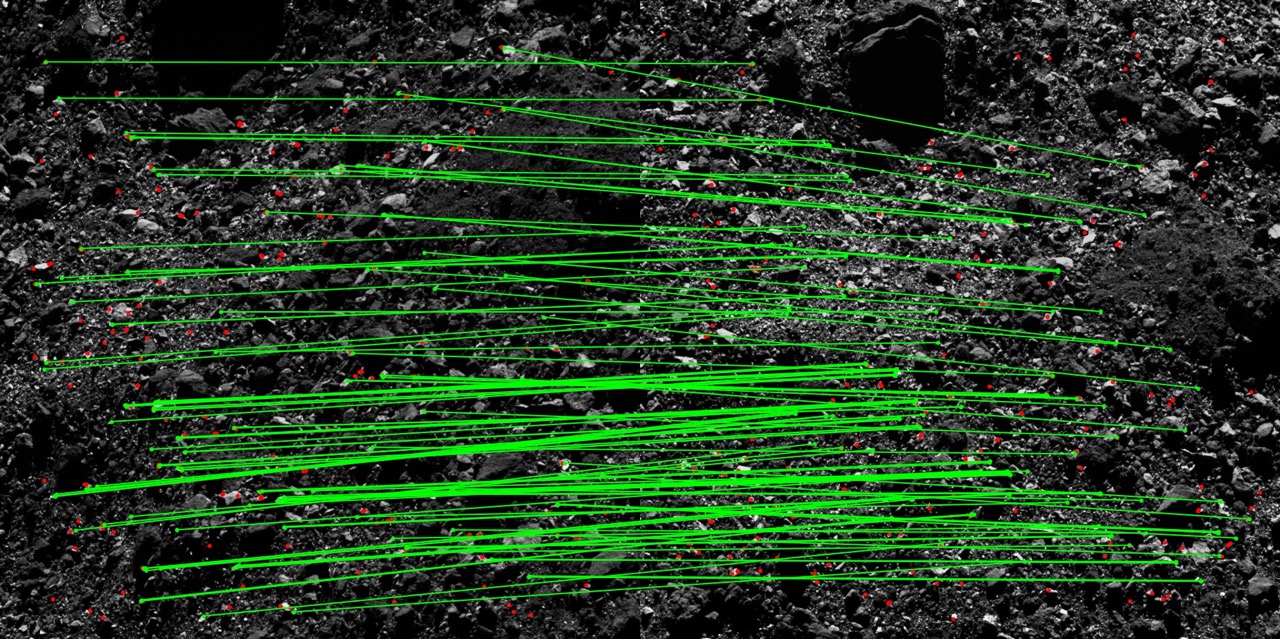}
  \end{subfigure}\\
  \vspace{0.5pt}
  \begin{subfigure}[t]{\linewidth}
    \includegraphics[width=.935\linewidth]{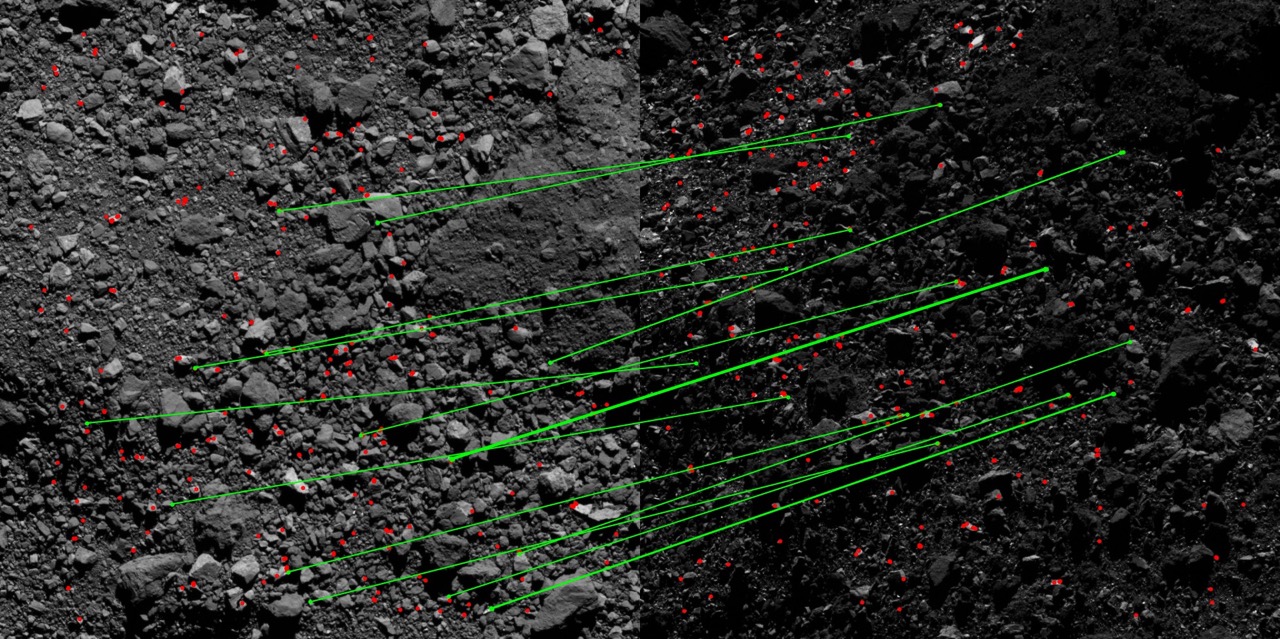}
  \end{subfigure}\\
  \vspace{4pt}
  \begin{subfigure}[t]{\linewidth}
    \includegraphics[width=.935\linewidth]{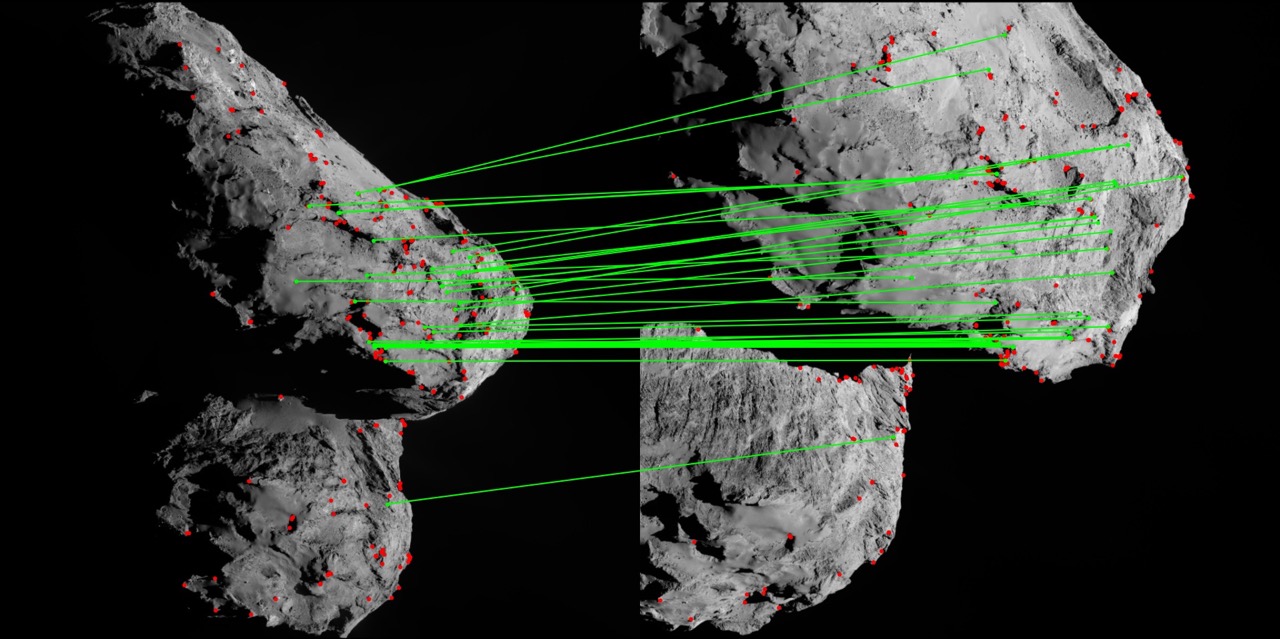}
  \end{subfigure}\\
  \begin{subfigure}[t]{\linewidth}
    \includegraphics[width=.935\linewidth]{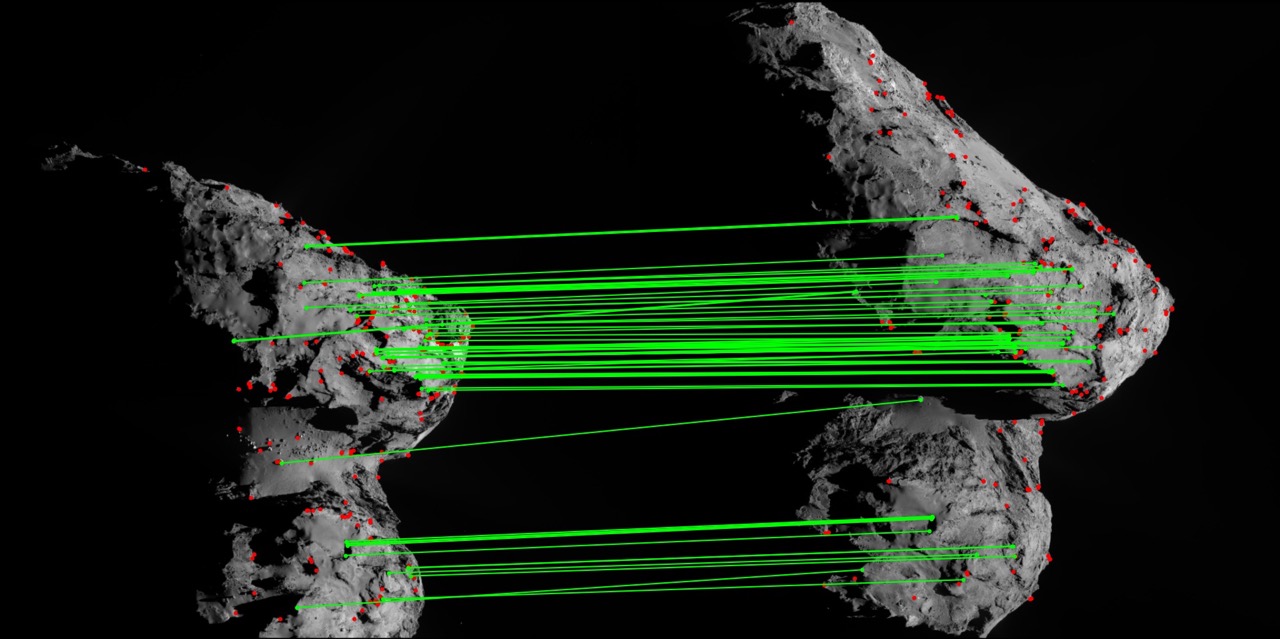}
  \end{subfigure}\\
    \vspace{0.5pt}
  \begin{subfigure}[t]{\linewidth}
    \includegraphics[width=.935\linewidth]{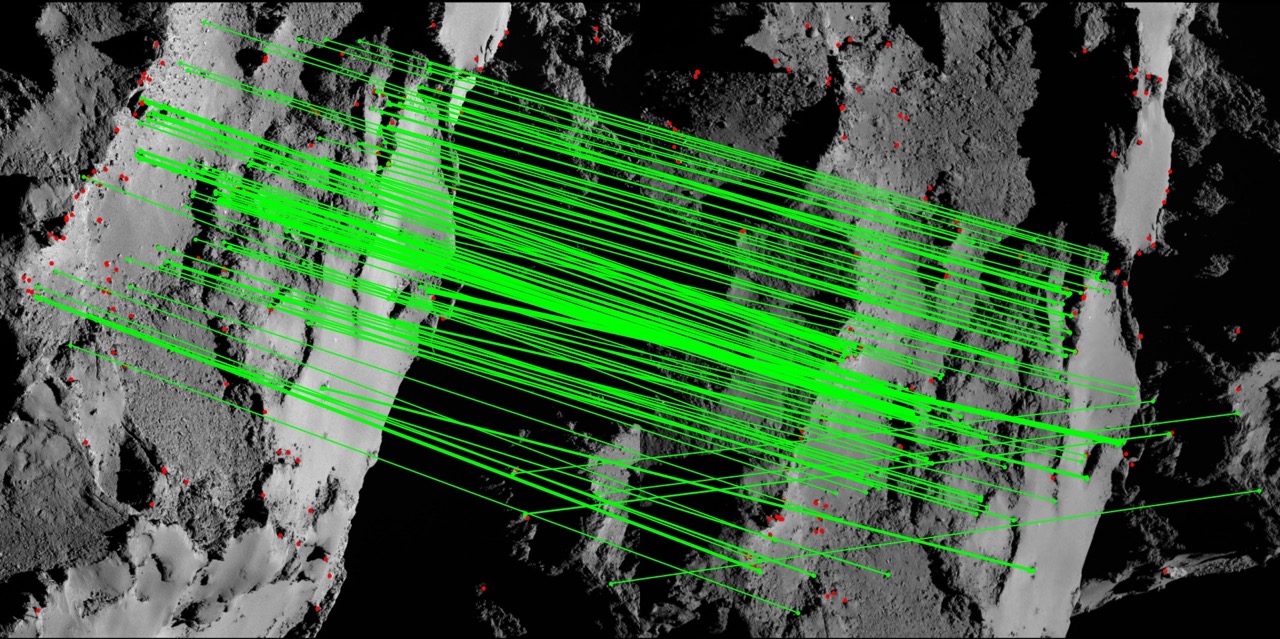}
  \end{subfigure}%
  \caption*{\footnotesize{FAST + DidymosNet$^{\pm}$}}
\end{subfigure}
\hspace{-5.5pt}
\begin{subfigure}[t]{0.20\linewidth}
  \centering
  \begin{subfigure}[t]{\linewidth}
    \includegraphics[width=.935\linewidth]{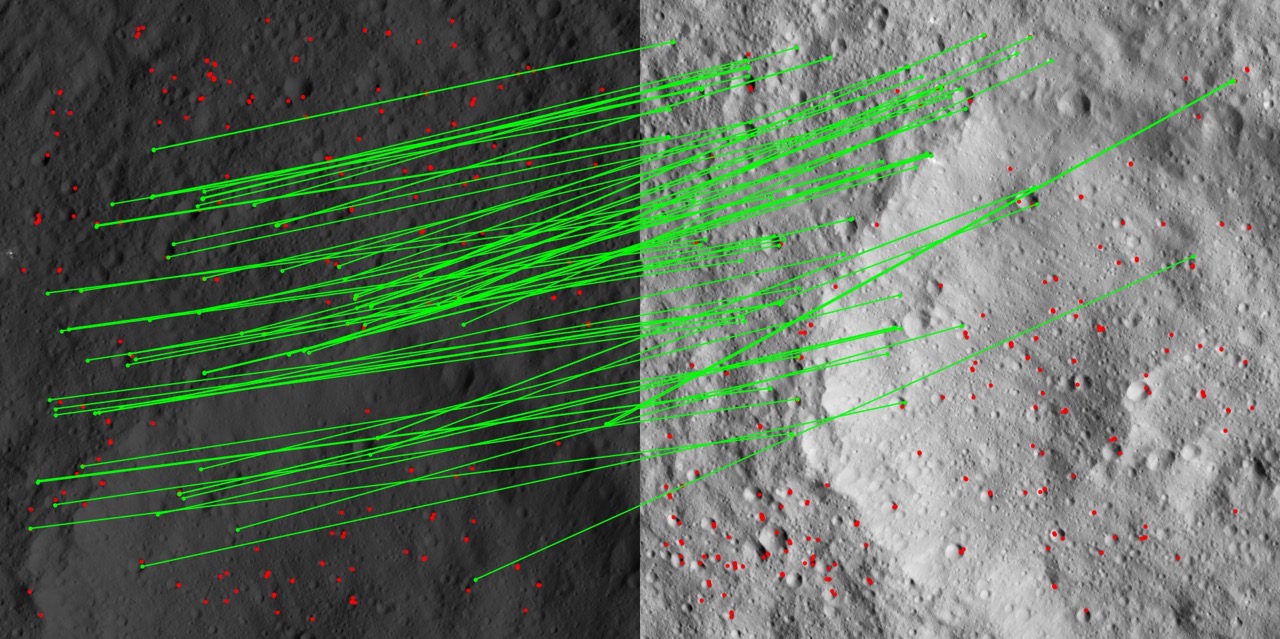}
  \end{subfigure}\\
  \vspace{0.5pt}
  \begin{subfigure}[t]{\linewidth}
    \includegraphics[width=.935\linewidth]{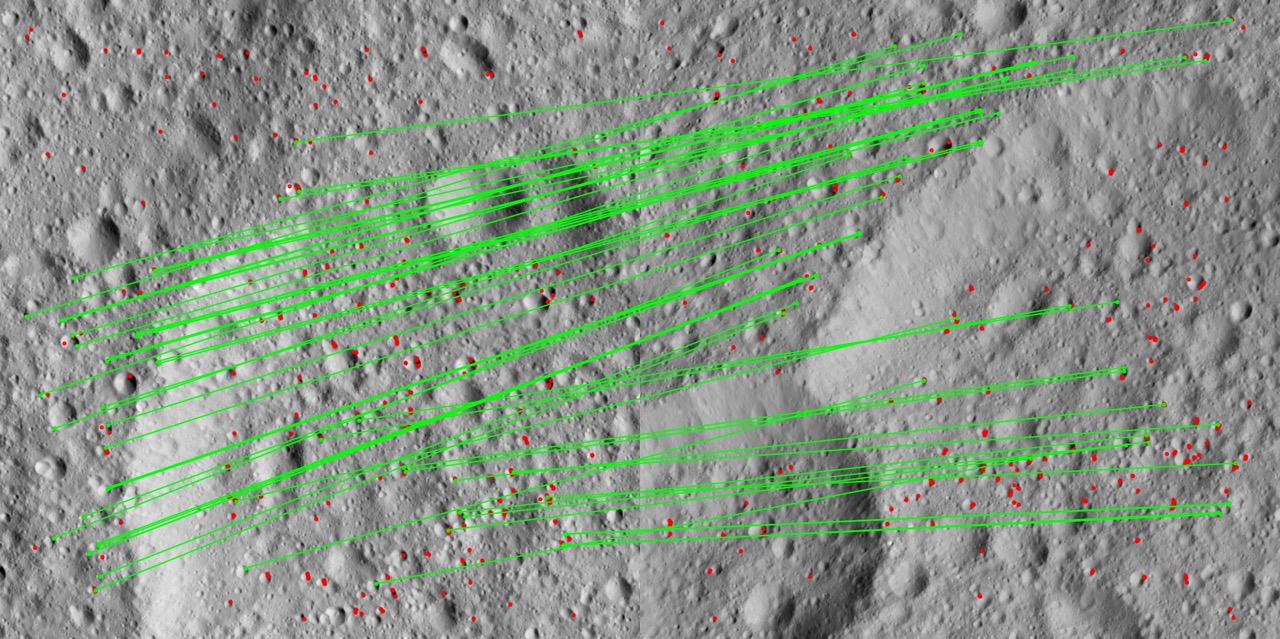}
  \end{subfigure}\\
  \vspace{4pt}
  \begin{subfigure}[t]{\linewidth}
    \includegraphics[width=.935\linewidth]{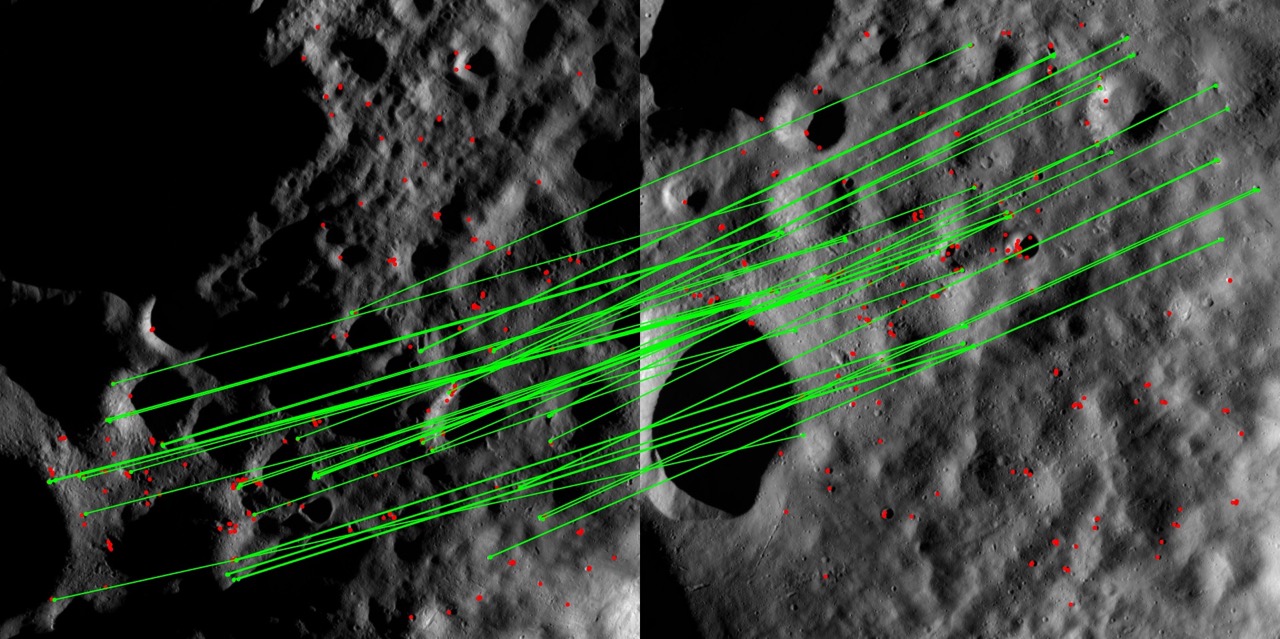}
  \end{subfigure}\\
  \begin{subfigure}[t]{\linewidth}
    \includegraphics[width=.935\linewidth]{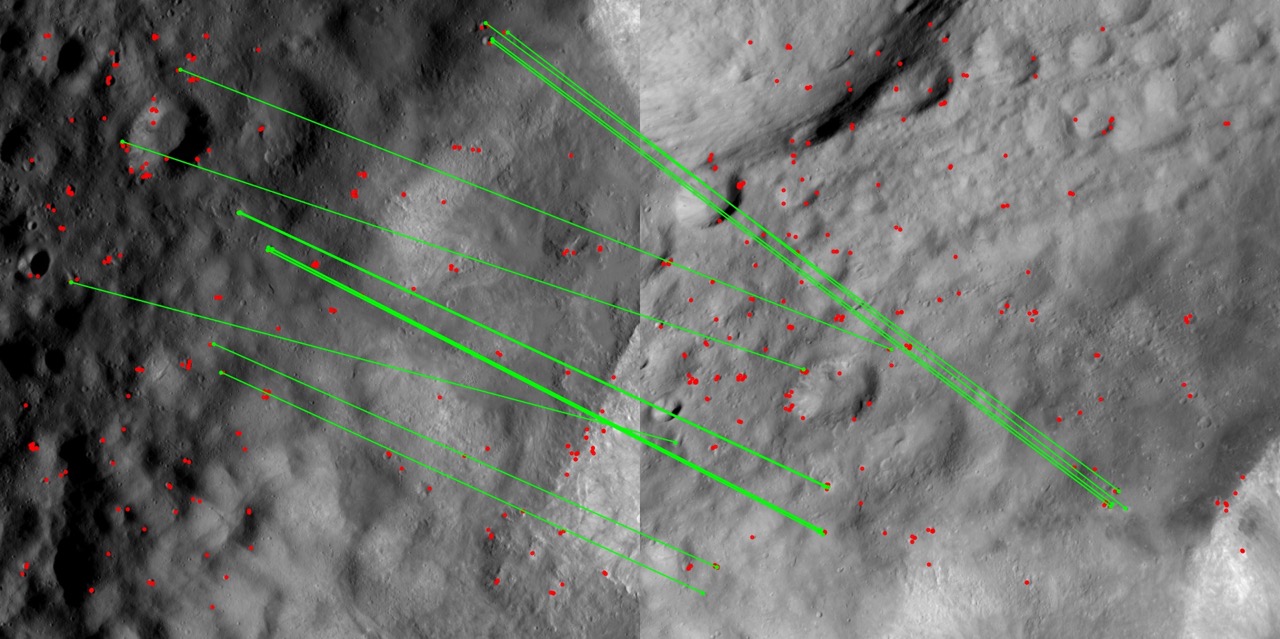}
  \end{subfigure}\\
  \vspace{0.5pt}
  \begin{subfigure}[t]{\linewidth}
    \includegraphics[width=.935\linewidth]{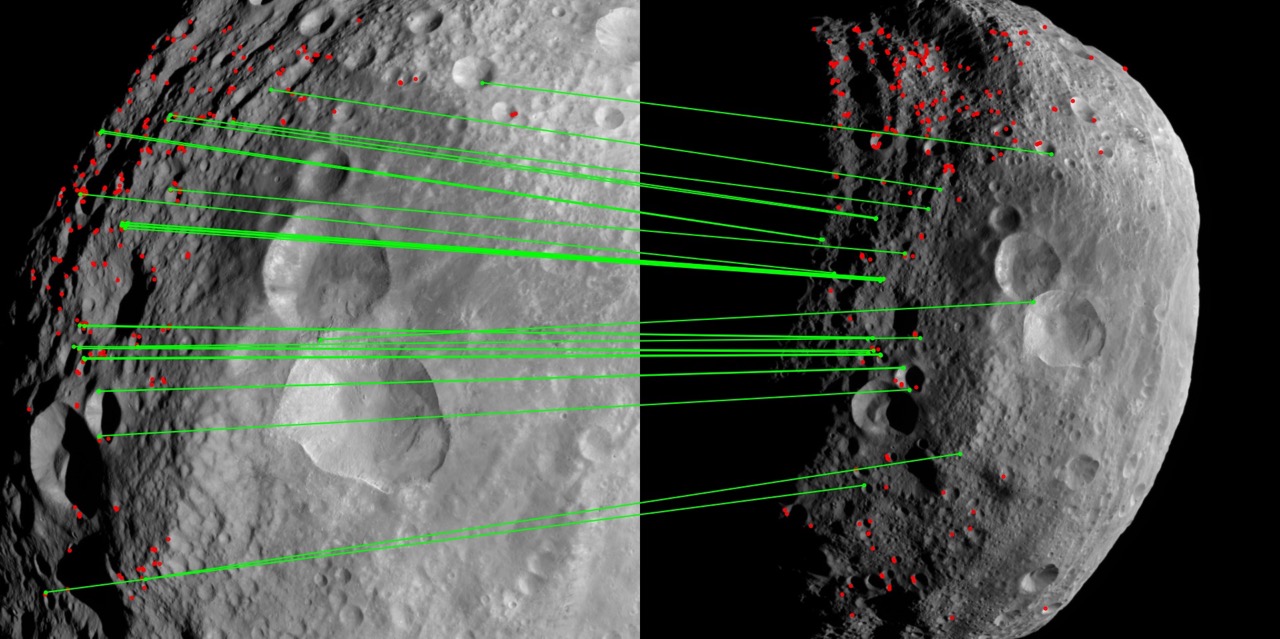}
  \end{subfigure}\\
  \vspace{4pt}
  \begin{subfigure}[t]{\linewidth}
    \includegraphics[width=.935\linewidth]{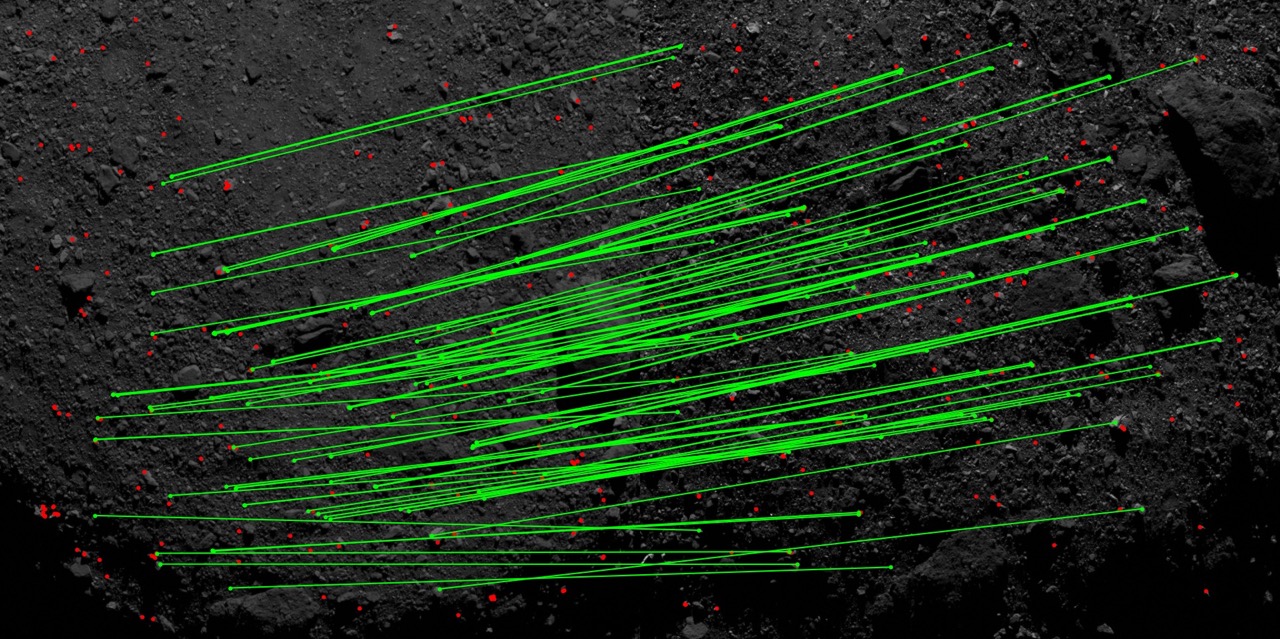}
  \end{subfigure}\\
  \begin{subfigure}[t]{\linewidth}
    \includegraphics[width=.935\linewidth]{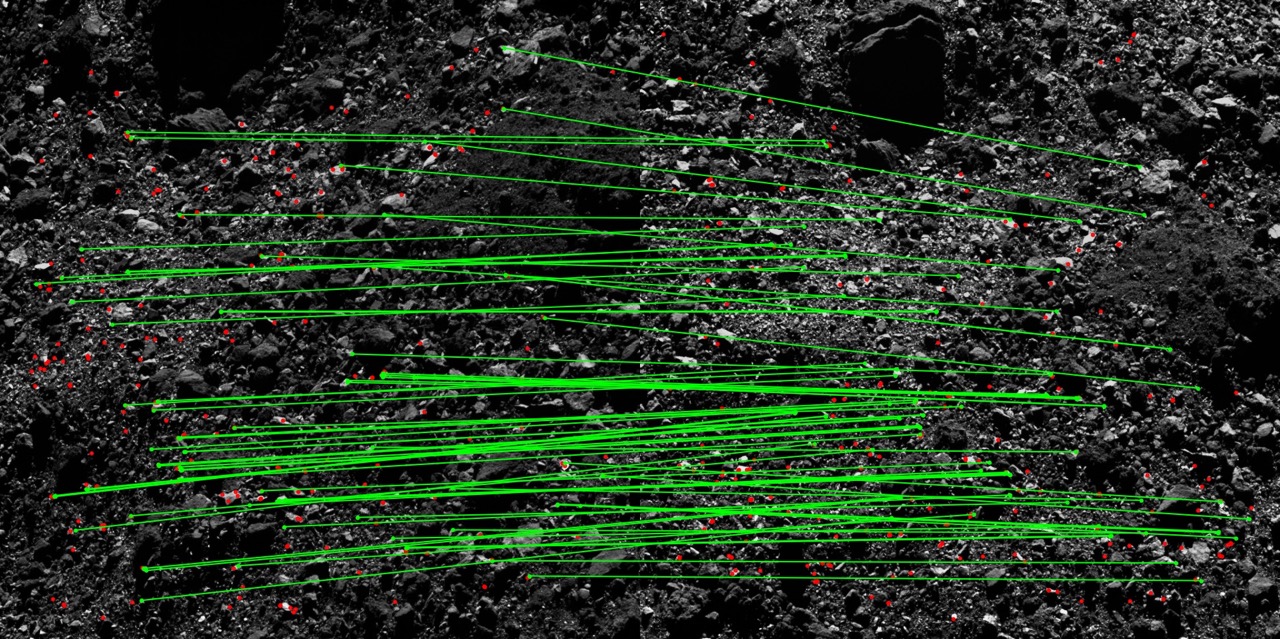}
  \end{subfigure}\\
  \vspace{0.5pt}
  \begin{subfigure}[t]{\linewidth}
    \includegraphics[width=.935\linewidth]{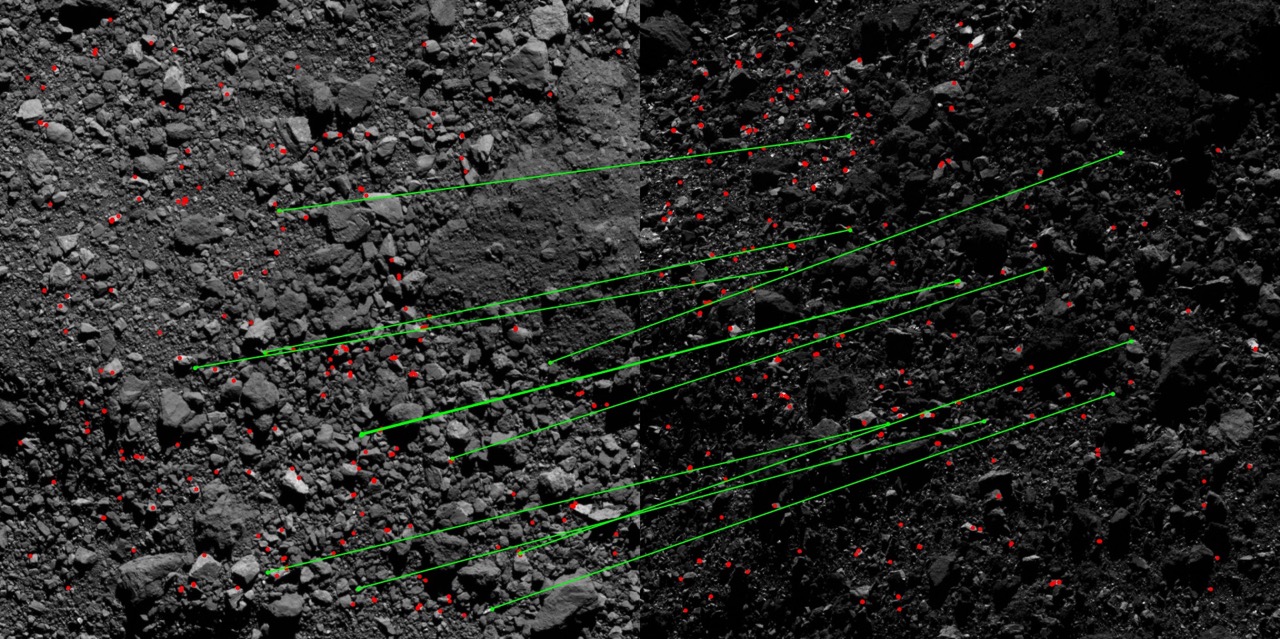}
  \end{subfigure}\\
  \vspace{4pt}
  \begin{subfigure}[t]{\linewidth}
    \includegraphics[width=.935\linewidth]{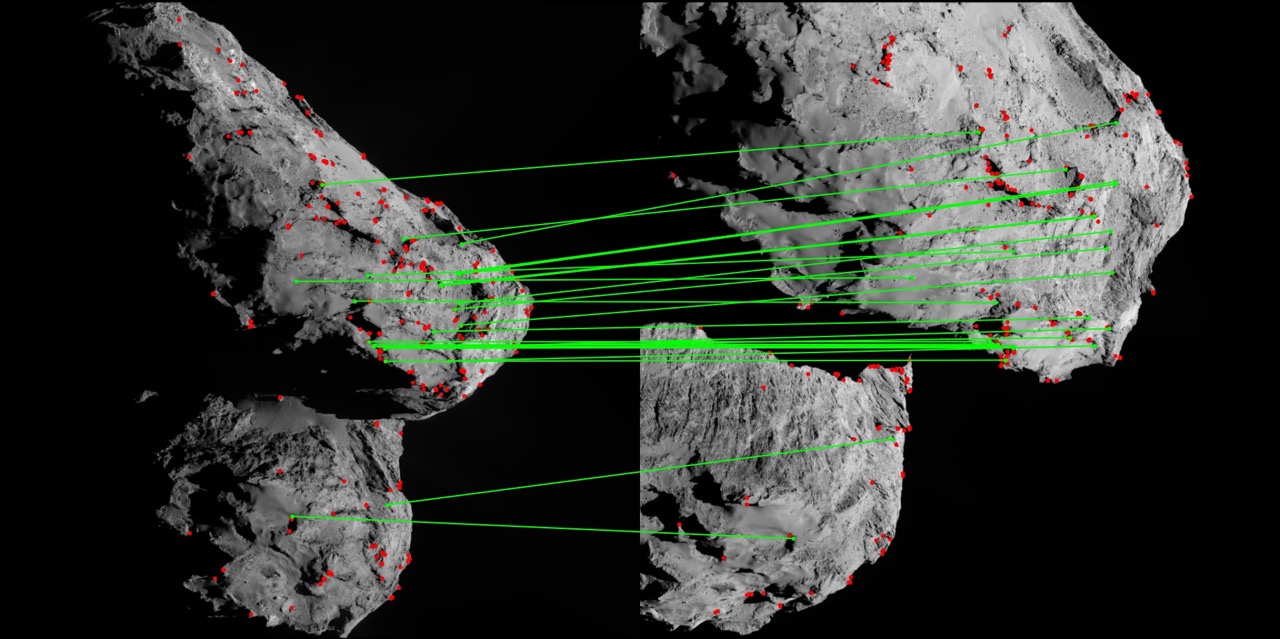}
  \end{subfigure}\\
  \begin{subfigure}[t]{\linewidth}
    \includegraphics[width=.935\linewidth]{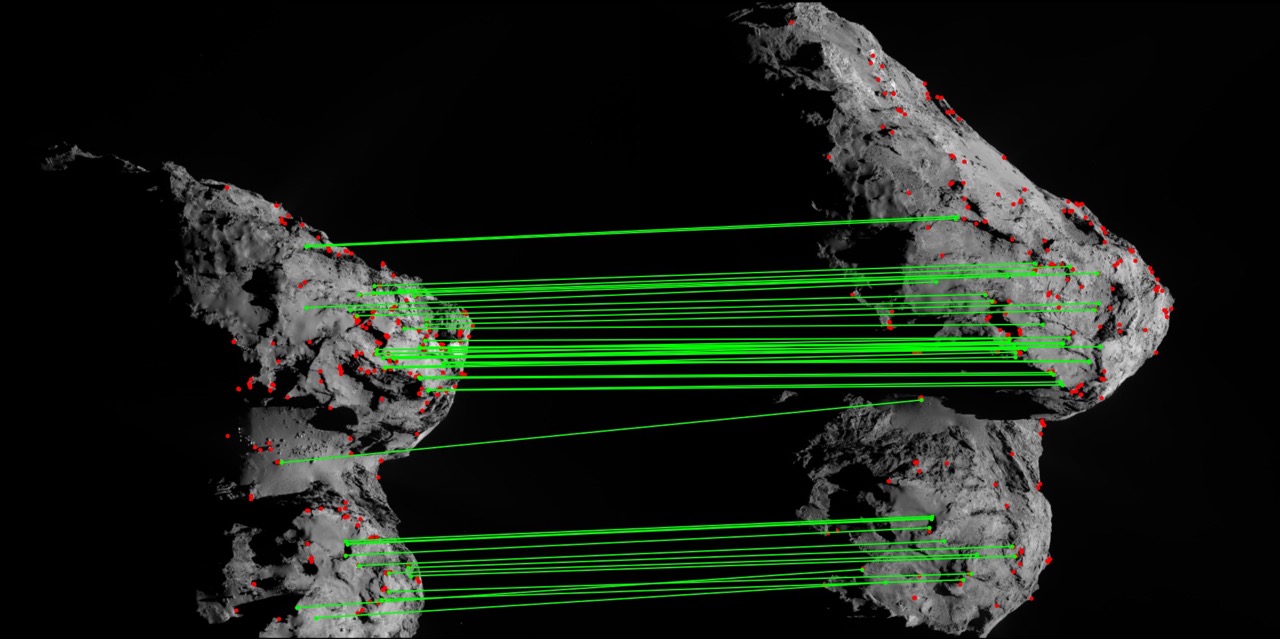}
  \end{subfigure}\\
    \vspace{0.5pt}
  \begin{subfigure}[t]{\linewidth}
    \includegraphics[width=.935\linewidth]{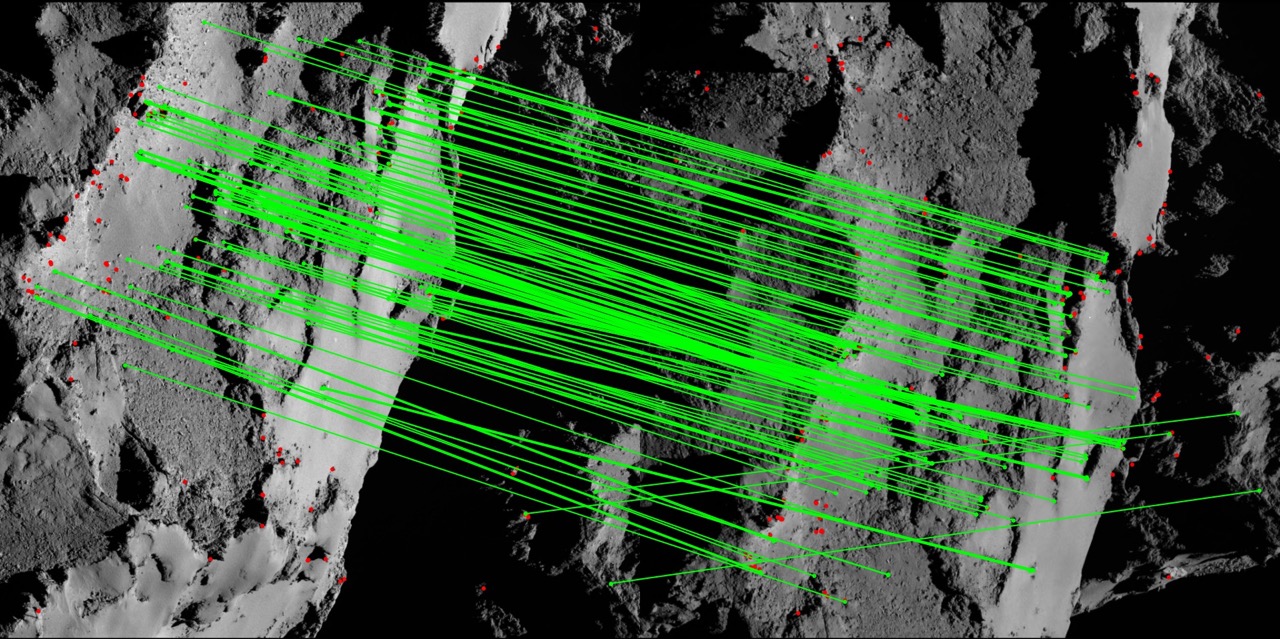}
  \end{subfigure}%
  \caption*{\footnotesize{FAST + DidymosNet$^{\pm'}$}}
\end{subfigure}
\end{minipage}

%% file: Text/appendix.tex
\section{Appendix: Architecture Details}

In this section, we present an ablation study with respect to some of the architecture elements that were not covered in the main text of the paper. 
The results in Table \ref{tab:arch-ablation} show that our baseline model, which replaces the final $8\times 8$ CNN layer of HyNet with three successive $2\times 2$ layers, offers comparable performance to HyNet, while significantly reducing the number of computations and parameters. 
Moreover, we followed Bulat and Tzimiropoulos~\cite{bulat2019bmvc} and implemented two different variants for the scaling factor $\gamma$ in Equation \eqref{eq:binary-quant}: the first variant learns one scaling factor $\lambda_k$ for each output channel $k \in \{1, \ldots, c_{\text{out}}\}$, i.e., $\gamma = \alpha_k$, and the second variant additionally learns a scaling factor $\alpha_i$ for each output vertical spatial dimension $i \in \{1, \ldots, h_{\text{out}}\}$ and a scaling factor $\beta_j$ for each output horizontal spatial dimension $j \in \{1, \ldots, w_{\text{out}}\}$, i.e., $\gamma = \alpha_i \cdot \beta_j \cdot \lambda_k$. 
These results are shown in Table \ref{tab:arch-ablation}, where ``---" corresponds to no explicit scaling in the convolutional layers. 
Contrary to previous works on network quantization for image classification~\cite{rastegari2016eccv,bulat2019bmvc}, we found that explicitly learning scaling factors within each BNN layer offered virtually no performance benefits compared to relying solely on the channel-wise learnable affine parameters of the normalization layers. 
This is likely due to the use of FRN layers~\cite{singh2020cvpr}, which forgo the mean-shift operation employed by traditional batch normalization layers and directly apply a channel-wise affine transformation to the normalized filter responses, and the lack of residual operations commonly used in image classification architectures.

\begin{table*}[htp!]
\footnotesize
\centering
\scshape
\ra{1.5}
\caption{\textbf{Architecture ablation study.} Ablation study for different components of the proposed architecture including the BNN scaling factor ($\gamma$ in Equation \eqref{eq:binary-quant}) and the structure of the last CNN layer (Conv7). The size of the weights and activations (W/A) in bits are included for clarity. Performance is measured with respect to FPR95 for the model trained on the Dawn @ 4 Vesta dataset.} 
\begin{adjustbox}{width=\linewidth}
\input{Figures/arch-ablation}
\end{adjustbox}\\
\label{tab:arch-ablation}
\end{table*}








%% file: Figures/arch-ablation.tex
\begin{tabular}{lcccrrr}
\toprule
 &     &     &        & \multicolumn{3}{c}{FPR95}                      \\
\cmidrule(lr){5-7}
    & Scaling & W/A & Conv7 & 67P & Bennu & Ceres \\
\midrule 
 HyNet               & --- & 32/32 & Conv(8,1)                                                                  &      5.45 &       3.98 &       0.01 \\
 \rowcolor[gray]{0.9}
 Baseline (ours)     & --- & 32/32 & 2$\cdot$Conv(2,2)+Conv(2,1)                                                &      6.06 &       4.06 &       0.01 \\
 \midrule 
 \rowcolor[gray]{0.9}
  DidymosNet    & --- & 1/1 & 2$\cdot$Conv(2,2)+Conv(2,1)                                                       &      9.66 &       7.43 &       0.01 \\
                & $\gamma = \lambda_k$ & 1/1 & 2$\cdot$Conv(2,2)+Conv(2,1)                                &      9.87 &       6.97 &       0.01 \\
                & $\gamma = \alpha_i \cdot \beta_j \cdot \lambda_k$ & 1/1 & 2$\cdot$Conv(2,2)+Conv(2,1)   &     10.30 &       6.87 &       0.01 \\
\bottomrule
\end{tabular}